%% file: main.tex
\documentclass{article}

\PassOptionsToPackage{numbers, compress}{natbib}

\usepackage[final]{neurips_2021}

\usepackage[utf8]{inputenc} 
\usepackage[T1]{fontenc}    
\usepackage{hyperref}       
\usepackage{url}            
\usepackage{booktabs}       
\usepackage{amsfonts}       
\usepackage{nicefrac}       
\usepackage{microtype}      
\usepackage{xcolor}         

\usepackage{graphicx}
\usepackage{subfigure}
\usepackage{amsmath}
\usepackage{amssymb}
\usepackage{algorithm}
\usepackage{algorithmic}
\usepackage{enumerate}
\usepackage{multirow}

\include{preambles/macros}
\graphicspath{{./figures/}}

\title{Brick-by-Brick: Combinatorial Construction\\with Deep Reinforcement Learning}
\input{preambles/authors}

\begin{document}

\maketitle

\begin{abstract}
Discovering a solution in a combinatorial space is prevalent 
in many real-world problems but it is also challenging due to diverse complex constraints 
and the vast number of possible combinations.
To address such a problem, we introduce a novel formulation, 
\emph{combinatorial construction}, which requires a building agent 
to assemble unit primitives (i.e., LEGO bricks) sequentially -- every connection 
between two bricks must follow a fixed rule, 
while no bricks mutually overlap.
To construct a target object, 
we provide incomplete knowledge about the desired target (i.e., 2D images) instead of exact and explicit volumetric information to the agent.
This problem requires a comprehensive understanding of partial information 
and long-term planning to append a brick sequentially, 
which leads us to employ reinforcement learning.
The approach has to consider a variable-sized action space where a large number of invalid actions, which would cause overlap between bricks, exist.
To resolve these issues, our model, dubbed \emph{Brick-by-Brick}, adopts an action validity prediction network that efficiently filters invalid actions for an actor-critic network.
We demonstrate that the proposed method successfully learns to construct an unseen object 
conditioned on a single image or multiple views of a target object.
\end{abstract}

\input{sections/01_introduction}
\input{sections/02_construction}
\input{sections/03_method}
\input{sections/04_experiments}
\input{sections/05_related_work}
\input{sections/06_conclusion}

\begin{ack}
This work was supported by the IITP
grants (No.2019-0-01906: AI Graduate School Program - POSTECH, No.2021-0-02068: AI Innovation Hub) funded by Ministry of Science and ICT, Korea and Samsung Electronics Co., Ltd (IO201208-07822-01).
JK carried out this research during a research internship at the Vector Institute, 
and JK and HC equally contributed to this work.
BK was funded by NSERC and the Ontario Graduate Scholarship.
GWT and BK also acknowledge support from CIFAR and the Canada Foundation for Innovation.
Resources used in preparing this research were provided, in part, by the Province of Ontario, the Government of Canada through CIFAR, and companies sponsoring the Vector Institute: \url{http://www.vectorinstitute.ai/\#partners}.
We also thank Hyeonwoo Noh for helpful discussions.
\end{ack}

\bibliography{kjt}
\bibliographystyle{plainnat}

\clearpage

\input{sections/07_supplementary}

\end{document}

%% file: preambles/macros.tex
\newcommand{\secref}[1]{Section~\ref{#1}}
\newcommand{\figref}[1]{Figure~\ref{#1}}
\newcommand{\tabref}[1]{Table~\ref{#1}}

\newcommand{\eqnref}[1]{Equation~(\ref{#1})}

\newcommand{\bstheta}{\boldsymbol \theta}

\newcommand{\bv}{\mathbf{v}}

\newcommand{\bx}{\mathbf{x}}

\newcommand{\bz}{\mathbf{z}}

\newcommand{\bm}{\mathbf{m}}

\newcommand{\be}{\mathbf{e}}

\newcommand{\bC}{\mathbf{C}}

\newcommand{\bT}{\mathbf{T}}

\newcommand{\calL}{\mathcal{L}}
\newcommand{\calN}{\mathcal{N}}
\newcommand{\calO}{\mathcal{O}}

\newcommand{\calT}{\mathcal{T}}

\newcommand{\bbE}{\mathbb{E}}

\newcommand{\bbR}{\mathbb{R}}

\newcommand{\bbZ}{\mathbb{Z}}

\newcommand{\twobyfour}{2\!\times\!4}

\newcommand{\mlp}{\textrm{MLP}}
\newcommand{\cnn}{\textrm{CNN}}
\newcommand{\aggregate}{\textrm{aggregate}}

\newcommand{\gn}{\textrm{GNN}}

\newcommand{\piv}{\textrm{piv}}
\newcommand{\off}{\textrm{off}}

\newcommand{\tar}{\textrm{tar}}

\newcommand{\ours}{\textrm{B}^3}

%% file: preambles/authors.tex
\author{%
Hyunsoo Chung\thanks{Equal contribution.}\enspace$^{1}$
\quad
Jungtaek Kim\footnotemark[1]\enspace$^{1}$
\quad
Boris Knyazev$^{2}$$^{3}$
\quad
Jinhwi Lee$^{1}$$^{4}$
\\
\textbf{Graham W. Taylor}$^{2}$$^{3}$
\quad
\textbf{Jaesik Park}$^{1}$
\quad
\textbf{Minsu Cho}$^{1}$
\\ \\
$^1$POSTECH \quad $^2$University of Guelph \quad $^3$Vector Institute \quad $^4$POSCO
\\
\texttt{\{hschung2,jtkim\}@postech.ac.kr}
}

%% file: sections/01_introduction.tex

\section{Introduction\label{sec:introduction}}

A combinatorial space, typically characterized by discrete variables and their combinations, 
often induces interesting yet challenging problems such as traveling salesperson and minimum spanning tree~\citep{KorteB2018book,CappartQ2021arxiv}.
The main challenges lie in the vast number of possible combinations
as well as complex constraints imposed on them.
In a similar spirit, we suggest a novel problem formulation, \emph{combinatorial construction}, that focuses on the real-world construction procedure.
Given only incomplete target information 
(i.e., 2D images or multiple views of a target object)~\citep{LiY2020eccv,HanW2020cvpr}, 
an agent sequentially assembles unit primitives (i.e., LEGO bricks).
The proposed formulation is combinatorial since it engages repetitive placement of primitives, which leads to a large number of available solutions.
Distinct property of our proposed formulation, however, is that the agent must build the solution incrementally by adding on to the partial solution.
Specifically, a brick, which is a unit primitive of the object of interest, is placed on a discrete space by connecting to one of the previously assembled bricks.
In addition, every connection between two bricks must follow a fixed rule while no bricks mutually overlap. 
Each assembly (i.e., action) executed by the agent is, thus, modeled as selecting one of the feasible connections to place a new brick.

The problem we introduce closely depicts how humans understand an object and adapt the acquired knowledge to a downstream task.
Humans naturally analyze a 3D object by picturing 
its part-by-part decomposition and 
consequently grasp a rich semantic understanding~\citep{HoffmanDD1984cog,LakeB2011cogsci}.
In various fields, they utilize an inherent ability to decompose objects to
effectively solve challenging tasks such as 
object classification~\citep{HuberD2004cvpr}, 
robot grasp planning~\citep{AleottiJ2011icra}, 
and part segmentation~\citep{QiCR2017cvpr,MoK2019cvpr}.
Likewise, humans exploit this ability to solve the inverse problem -- combinatorial construction.
Given a desired object to be constructed and no strong supervision 
(i.e., ordered step-by-step instructions), humans can often still manage 
to build a valid target object by carefully planning or, sometimes improvising, 
the sequence of actions.
Our environment, which corresponds to the proposed problem, is designed to learn and test such behavior with only partial information of the desired target available to the agent.

Successfully constructing an object in our setup requires a comprehensive understanding 
of incomplete target information with the current structured state
of assembled bricks and long-term planning to append each brick efficiently.
These requirements, along with the absence of sequence-level supervision, incentivize us to devise a reinforcement learning (RL) approach~\citep{BapstV2019icml,SimmG2020icml_b}.
In this domain, however, we must carefully handle both an indefinite action space 
and the existence of many invalid actions when applying RL~\citep{ZahavyT2018neurips}.
In particular, both defining an action space that varies by 
the number of assembled bricks and distinguishing 
an invalid action that would cause overlap with other existing bricks 
quickly become intractable as more bricks are placed.
To resolve the aforementioned issues, our model, dubbed Brick-by-Brick ($\ours$), 
adopts an action validity prediction network that filters invalid actions 
to an actor-critic network.
In addition to the novel RL formulation, we use graph-structured representation of the brick combination to interpret the assembling process as a sequential graph generation process.

Overall, we summarize our contributions as follows:
\begin{enumerate}[(i)]
    \item We propose a novel problem formulation, combinatorial construction, that closely resembles a real-world object construction process that engages repetitive placement of components;
    \item We design an RL agent for combinatorial construction, dubbed Brick-by-Brick ($\ours$), to effectively address both a growing action space and a vast set of invalid actions;
    \item We implement the corresponding environment based on OpenAI Gym and introduce new novel evaluation scenarios that vary by their incomplete partial target information.
\end{enumerate}

\input{tables/tab_comparisons}

%% file: tables/tab_comparisons.tex
\begin{table}[t]
    \centering
    \small
    \caption{Analysis of recent studies in terms of state representation, supervision, conditioning, target objects, and action validation.
    CE and IoU stand for cross-entropy and intersection over union with respect to volumetric comparisons.
    Direct forwarding (denoted as direct), sampling \& checking (denoted as sampling), and our pretrained action validity prediction network (denoted as pretrained) indicate respective strategies that filter invalid actions; see the corresponding section for their details.
    \label{tab:comparisons}}
    \setlength{\tabcolsep}{4pt}
    \begin{tabular}{lcccccc}
        \toprule
        \multirow{2}{*}{\textbf{Method}} & \multirow{2}{*}{\textbf{State}} & \multirow{2}{*}{\textbf{Supervision}} & \multirow{2}{*}{\textbf{Conditioning}} & \multirow{2}{*}{\textbf{Target}} & \textbf{Action} \\
        &&&&& \textbf{Validation} \\
        \midrule
        \citet{HamrickJB2018cogsci} & Image & Task-dependent & N/A & 2D & Direct \\
        \citet{BapstV2019icml} & Object/Image & Task-dependent & Object and/or image & 2D & Direct \\
        \citet{KimJ2020ml4eng} & Set & Overlap & Exact target volume & 3D & Sampling \\
        \citet{ThompsonR2020neuripsw} & Graph & Step-wise CE & One-hot class info. & 3D & Direct \\
        Brick-by-Brick (B$^3$, ours) & Graph/Image & IoU & Image or set of images & 3D & Pretrained \\
        \bottomrule
    \end{tabular}
\end{table}

%% file: sections/02_construction.tex
\section{Combinatorial Construction\label{sec:construction}}

To formulate the combinatorial construction problem, 
we start by defining a unit primitive that is used to construct a 3D object
and an action space that determines where to assemble the next primitive.

As a unit primitive, we utilize a $\twobyfour$ brick, 
which has eight studs and their fit cavities.
This design choice yields a consistently varying action space, 
implying that if we add one brick to the current state of brick combination, 
we can efficiently define the next action space.
We want to emphasize that with only six $\twobyfour$ bricks, 
we can create \emph{915,103,765 combinations}~\citep{EilersS2016amm}.
Accordingly, our choice of the primitive does not make our problem a trivial task; 
instead, every decision of where we place the next primitive can
deteriorate the quality of the final result because there exists a plethora of wrong paths.\looseness=-1

With our specific choice of unit primitives, we can define an action space for determining 
the next action and evaluating the future states.
However, since every assembly step gradually expands the action space, 
na\"ive approaches to defining the growing action space are not appropriate for our problem; 
an action space with redundant actions~\citep{ZahavyT2018neurips} is not applicable due to a varying action space, 
and an action sampling approach~\citep{HeJ2016emnlp} is also not suitable due to nominal or invalid actions.
Thus, we define a \emph{successive} action space composed of a two-step decision: 
(i) choosing a \emph{pivot brick} and (ii) choosing an \emph{offset} from the pivot brick.

Before explaining a pivot brick, we first assume a simplified assembly scenario 
that follows an Eulerian path\footnote{It is a path that visits all the vertices without revisiting the edges visited before.} -- a new brick is always placed by connecting to the last assembled brick.
This enables us to define a finite action space though most actions are infeasible to perform with the Eulerian path.
To broaden the search space by generalizing the Eulerian path, one of previously assembled bricks is chosen as a pivot brick.
Then, $\ours$ decides an offset from the pivot brick, which describes how the next brick is placed relative to the pivot.
Thanks to the homogeneous brick type, the number of available offsets is finite and consistent where we do not consider the validity of such offsets in a certain state -- for $\twobyfour$ bricks, 
there exist a maximum of 92 available offsets.

Due to the disallowance of overlap between bricks, 
our agent must consider \emph{invalid actions} among available actions 
during assembly.
Identifying the validity of actions from the current brick combination becomes intractable as more bricks are placed; 
the complexity of this process is $\calO(|A_{\off}| t^2)$, 
where $A_{\off}$ is an action space for offsets and $t$ is the cardinality of assembled bricks at a given step; 
see the supplementary material for qualitative results on this complexity.
Such expensive overhead for validation naturally leads us to 
adopt an action validity prediction network that learns to identify invalid actions 
with a single forward pass.

As described in~\secref{sec:introduction} and this section, 
our problem has interesting but challenging characteristics derived from the assumptions on 
discrete placement, a connectivity rule, disallowance of overlap, 
and ultimately invalid actions.
We, therefore, present a comparison to other existing studies in terms of state, 
supervision (or a reward function), conditioning, target objects, and 
action validation, as shown in~\tabref{tab:comparisons}. 
Compared to the previous work~\citep{HamrickJB2018cogsci,BapstV2019icml,KimJ2020ml4eng,ThompsonR2020neuripsw}, 
our method $\ours$ constructs an object in 3D with incomplete target information and pretrained validity prediction component; see Sections~\ref{sec:experiments} 
and \ref{sec:related_work} for a detailed description.

%% file: sections/03_method.tex
\section{Brick-by-Brick\label{sec:method}}

\begin{figure}[t]
\centering
\includegraphics[width=0.995\textwidth]{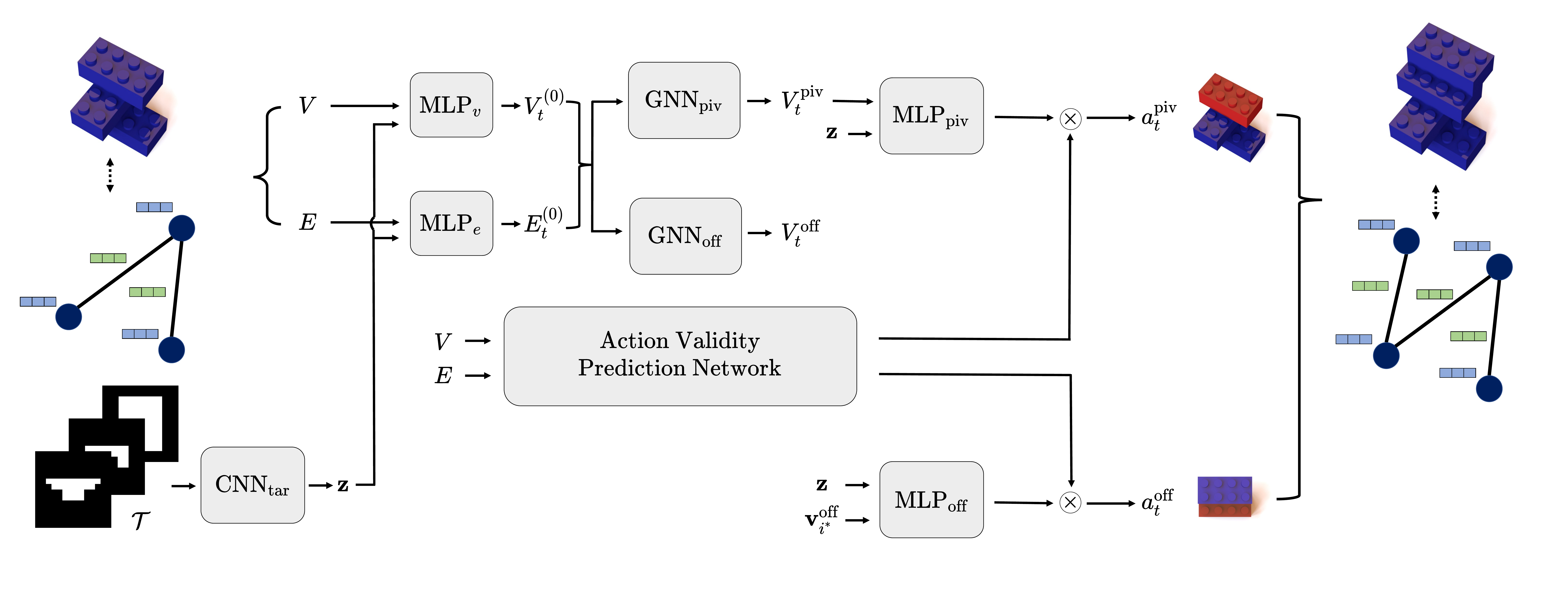}
\caption{Overview of our proposed method $\ours$. A state input $s_t = (G_t, \calT)$ is embedded and passed through CNNs, GNNs, and MLPs to predict an action $a_t$ that consists of a pivot brick indicator $a_t^{\piv}$ and an offset from the pivot brick $a_t^{\off}$. Moreover, an action validity prediction network helps to filter invalid actions from the action space where we determine the next brick to assemble. The red brick in both actions of $a_t^{\piv}$ and $a_t^{\off}$ indicates the chosen brick. See \secref{sec:method} for the details.}
\label{fig:overview}
\end{figure}

In this section, we briefly introduce the definition of RL 
and the corresponding framework for combinatorial construction, 
where an agent places a brick sequentially.
We then explain the details of our model that learns to select 
appropriate actions given only partial information of the desired target 
so that the assembled 3D object resembles the target.
Moreover, to cope with a vast number of invalid actions in the process of assembly, 
we propose an action validity prediction network. See \figref{fig:overview} for the overall pipeline of our method $\ours$.

\paragraph{Definition.}
In a standard RL framework, there exists an agent that interacts with an environment 
by iteratively making decisions given an observation of the environment.
This follows general decision making procedure of a Markov decision process (MDP), 
where a transition function satisfies the Markov property, 
i.e., $p(s_{t+1}|s_{0},s_{1},\ldots,s_{t},a_{t}) = p(s_{t+1}|s_{t},a_{t})$,
where $s_t$ and $a_t$ are a state and an action at timestep $t$, respectively.

In our problem setting, we only consider a finite horizon MDP formally defined 
as a tuple of $(S, A, P, R, \gamma)$, where $S=\{s_t\}$ is a set of states, $A=\{a_t\}$ a set of actions, $R: S \times A \to \bbR$ a reward function, 
$P: S \times A \to S$ a transition function, 
and $\gamma \in [0, 1)$ a discount factor.
The goal of the agent is to learn a policy $\pi(a_{t}|s_{t})$ 
that maximizes the expected future cumulative reward.
We introduce our detailed MDP formulation for combinatorial construction 
in the following sections.

\subsection{Problem Formulation\label{subsec:prob}}

Given target information $\calT$, the agent aims to construct a 3D target object $\bT$ by assembling bricks sequentially, one brick for each $t$-th step.
Each $t$-th brick is represented by its pose $(\bx_t, d_t)$, where $\bx_t \in \bbZ^3$ is the center coordinate of the brick in a 3D space and $d_t \in \{0,1\}$ denotes one of two possible directions, meaning that its longer axis is aligned along either $x$ axis or $y$ axis in the 3D space.\looseness=-1

\paragraph{Target Information.} We are given as target information $\calT$ 
a set of binary images of a target object, which may correspond to incomplete and partial information about the target. 
In practice, this setup with partial information is more realistic than accessing full information of a 3D target shape. Our task thus is to create a sequence of unit bricks 
by inferring a target object 
from abstract information in a combinatorial manner.

\paragraph{State.}
Each $t$-th state $s_t$ of the MDP is represented by a tuple of 
a directed graph $G_t$ composed of $t$ bricks and target information $\calT$, i.e., $s_t=(G_t, \calT)$.
The graph is defined as 
$G_t = (V_t, E_t)$, 
where $V_{t} = \{\bv_i\}_{i=1}^{t}$ is a set of $t$ bricks, i.e., $\bv_i=(\bx_i, d_i) \in \bbZ^{4}$, 
and $E_{t} = \{\be_{ij}\}_{i,j=1}^{t}$ is a set of the offset vectors between two connected nodes, 
i.e., $\be_{ij} = (\bx_i - \bx_j, d_i \oplus d_j) \in \bbZ^4$.
Note that nodes are connected by edges according to sequential actions and relative offsets in pose are used for edge features in order to induce translational and orientational equivariance.
Since all the edges are bi-directional, we omit the arrows when displaying graphs in \figref{fig:overview}.

\paragraph{Action.}
We define a successive action space of choosing the pivot brick first and the corresponding offset next. 
Formally, with $t$ bricks assembled, 
we define an action $a_t = (a^{\piv}_t, a^{\off}_t)$ 
where $a^{\piv}_t$ is to select a pivot brick 
and $a^{\off}_t$ is to select an offset with respect to the pivot brick.
The pose of the next brick is then 
$(\bx^{\piv} + \Delta \bx, (d^{\piv} + \Delta d) \mod 2)$, 
where $(\bx^{\piv}, d^{\piv})$ and $(\Delta \bx, \Delta d)$ 
are determined by $a^{\piv}_t$ and $a^{\off}_t$, respectively. In choosing actions $a^{\piv}_t$ and $a^{\off}_t$, we exclude invalid actions: (i) choosing a pivot brick near which no additional brick can be placed and (ii) choosing an offset for the next brick that overlaps with existing bricks.
To select a valid action in the action space with a vast number of such invalid ones, we predict invalid actions in advance using the action validity prediction network and exclude them from action candidates preemptively; 
we mask out all the probabilities of invalid actions, re-normalize the distribution over actions $a_t$, and sample an action from the distribution.

\paragraph{Transition Function.}
Given state $s_t$ and action $a_t$, our transition function  $p(s_{t+1}|s_t,a_t)$ 
is designed to determine the next state $s_{t+1}$ by deterministically 
updating $s_{t}$ based on $a_t$.
The node of the new brick $\bv_{t+1}$ is created so that $V_{t+1} = V_{t} \cup \{ \bv_{t+1} \}$. 
The edges between the new brick and existing bricks in physical contact via studs are created  
so that $E_{t+1} = E_{t} \cup \{ \be_{(i)(t+1)} \}_{i \in \calN_{t+1}} \cup \{ \be_{(t+1)(i)} \}_{i \in \calN_{t+1}}$, 
where $\calN_{t+1}$ denotes the set of bricks in direct contact with the new brick $\bv_{t+1}$. 
As the result, the graph in the state $s_{t + 1}$ is updated to $G_{t + 1} = (V_{t + 1}, E_{t + 1})$.

\paragraph{Reward Function.}
In contrast to the tasks where a direct reward evaluation is readily available,
it is not trivial to quantify the object assembled by
combinatorial construction, 
especially, in the context of graph generative model~\citep{LiaoR2019neurips}.
To mitigate such an issue, we exploit the property of a voxel representation.
Given a desired object, 
we first create voxels in a closed space 
and determine the occupancy of voxels with a target object, 
after normalizing it to the bottom center of voxels.
We then transform the combination of currently assembled bricks  
into the occupancy of the voxels and measure the overlap with the target object:
\begin{equation}
    \Delta\textrm{IoU}(\bC_{t}, \bT) = \frac{\textrm{vol}(\bC_{t} \cap \bT)}{\textrm{vol}(\bC_{t} \cup \bT)} - \frac{\textrm{vol}(\bC_{t-1} \cap \bT)}{\textrm{vol}(\bC_{t-1} \cup \bT)},\label{eqn:delta_iou}
\end{equation}
where $\bC_t$, $\bC_{t-1}$, and $\bT$ are 
the occupied voxels at timestep $t$, timestep $t - 1$, and a desired target, respectively. In addition, vol$(\cdot)$ is a function that measures a volume.
The step-wise reward function is then $\Delta\textrm{IoU}$ 
if the new brick overlaps at least 50\% with the occupied voxels of target object and $0$ otherwise. 
Consequently, our agent will learn sequential placement of bricks 
to construct the object, without explicit supervision by maximizing \eqnref{eqn:delta_iou}, 
as will be described in the subsequent section.\looseness=-1

\subsection{Sequential Construction\label{subsec:actionsel}}

In this section, we describe how we process 
$G_t$ and $\calT$, which comprise a state, with different types of neural networks such as convolutional neural networks and graph neural networks.
The overview of this construction procedure is illustrated in~\figref{fig:overview}.

\paragraph{Node and Target Embeddings.}
Given a state $s_t = (G_t, \calT)$ 
where $G_t = (V_t, E_t)$, 
we first use a convolutional neural network (CNN) to extract features $\bz$ from the target:
\begin{equation}
    \bz = \cnn_{\textrm{tar}}(\calT).
    \label{eqn:cnn_tar}
\end{equation}
If the partial information is given as a set of images, the feature $\bz$ is obtained by first applying CNN to each image separately and then concatenating the outputs to a single vector. 

For node and edge features, an MLP embeds 
them with the target feature $\bz$:
\begin{equation}
    \bv_{i}^{(0)} = \mlp_v([\bv_{i}, \bz]), \quad \textrm{and} \quad \be_{ij}^{(0)} = \mlp_e([\be_{ij}, \bz]),\label{eqn:init_node_edge}
\end{equation}
for all $i, j \in \{1, \ldots, t\}$, where $[,]$ and $^{(0)}$ 
denote concatenation and the first layer, respectively.

Equations~\eqref{eqn:cnn_tar} and \eqref{eqn:init_node_edge} 
can be viewed as pre-processing inputs to feed in a graph neural network (GNN).
Inspired by \cite{BattagliaPW2018arxiv}, 
we apply a variant of graph networks (GNs)
in which a global graph feature is omitted.  
At the $\ell$-th layer of GNNs, we update edge features, aggregate the messages for each node, and update node features:
\begin{align}
    \be_{ij}^{(\ell + 1)} &= \mlp_e^{(\ell)}\left([\bv_i^{(\ell)}, \bv_j^{(\ell)}, \be_{ij}^{(\ell)}] \right), \label{eqn:gnn_propagation_1}\\
    \bm_i^{(\ell)} &= \sum_{j \in \calN_i} \aggregate \left( \be_{ij}^{(\ell + 1)} \right), \label{eqn:gnn_propagation_2}\\
    \bv_i^{(\ell + 1)} &= \mlp_v^{(\ell)}\left([\bv_i^{(\ell)}, \bm_i^{(\ell)}] \right),
    \label{eqn:gnn_propagation_3}
\end{align}
where $\calN_i$ is the neighborhood nodes of $\bv_i$, and \aggregate$(\cdot)$ is the aggregation function that computes a message for each node by aggregating the features of its neighboring nodes.
Note that $\mlp_v$, $\mlp_e$, $\mlp_v^{(\ell)}$, and $\mlp_e^{(\ell)}$ have their own learnable parameters.

\paragraph{Action Selection.}
In order to enrich representations for predicting $a_t$,
we employ two separate GNNs: $\gn_{\piv}$ and $\gn_{\off}$, 
with $L$ layers in total, 
to produce sets of node embeddings, $V_t^{\piv}$ and $V_t^{\off}$ 
for pivots and offsets:
\begin{equation}
    V_t^{\piv} = \gn_{\piv}(V_t^{(0)}, E_t^{(0)}), \quad \textrm{and} \quad V_t^{\off} = \gn_{\off}(V_t^{(0)}, E_t^{(0)}),
\end{equation}
where $V_t^{(0)} = \{\bv_i^{(0)}\}_{i=1}^{t}$ and $E_t^{(0)} = \{\be_{ij}^{(0)}\}_{i,j=1}^{t}$ are the sets of node and edge features 
obtained by \eqnref{eqn:init_node_edge}.
Each layer of both GNNs updates node and edge features using Equations~\eqref{eqn:gnn_propagation_1}, \eqref{eqn:gnn_propagation_2}, and~\eqref{eqn:gnn_propagation_3}.
Finally, the set of node and edge features, $V_t^{\piv}$ and $V_t^{\off}$, along with a target feature $\bz$ are used to decide the next action $a_t^{\piv}$ and $a_t^{\off}$:
\begin{equation}
    p\big(a_t^{\piv}\big) = \sigma\Big(\mlp_{\piv}\big([V_t^{\piv}, \bz] \big)\Big),
    \quad \textrm{and} \quad p\big(a_t^{\off}\big) = \sigma\Big(\mlp_{\off}\big([\bv_{i^*}^{\off}, \bz] \big)\Big),\label{eqn:actions}
\end{equation}
where $\sigma$ is a softmax function and $\bv_{i^*}^{\off}$ is the node feature 
selected by the index $i^*$ of $a_t^{\piv}$.

\paragraph{Action Validation.}
To tackle the issue of a vast number of invalid actions, we propose to learn an action validity prediction network. Previous work adopts direct forwarding~\citep{BapstV2019icml,ThompsonR2020neuripsw} or sampling \& checking~\citep{KimJ2020ml4eng}.
In the direct forwarding approach~\citep{BapstV2019icml,ThompsonR2020neuripsw}, the agent directly selects an action without any prior processing. This typically suffers from the early termination of episodes since the sequence of actions terminates with a deadlock as soon as the agent selects an invalid action. In the sampling \& checking approach~\citep{KimJ2020ml4eng}, valid actions are collected by sampling a random set of actions and checking the validity for each of them directly. This requires a high cost of iterative checking and the actions are limited to a small collection of checked actions.

In contrast to these approaches, we train a separate module to predict a large set of valid actions, enabling the agent to sufficiently 
explore the action space.
We train a GNN, of which 
the node-wise output predicts validity confidences for its candidate actions, i.e., candidate actions for the corresponding brick.
The structures of both pivot and offset validity prediction networks are identical 
to the networks described in~\eqnref{eqn:actions}, 
but the last activation is a sigmoid function and no target feature $\bz$ is used.
Importantly, these networks can be pretrained by the ground-truth 
action validity, which is obtained from randomly-assembled objects, 
and such a pretrained network can be used in training an actor-critic network, \emph{without re-training}.

Note that, if our action validity prediction network fails to filter invalid actions 
and one of such actions is selected by the agent, the corresponding episode is terminated.
Unlike the direct forwarding approach, training the agent with the action validity prediction network does not suffer from early termination as the validity prediction network masks out the majority of invalid actions.

\paragraph{Training.}
We adopt the proximal policy optimization (PPO) algorithm~\cite{SchulmanJ2017arxiv}, 
which is one of the state-of-the-art on-policy algorithms.
In particular, we optimize the clipped surrogate objective over parameters $\bstheta$:
\begin{equation}
    \calL(\bstheta) = \bbE \left[\min(r_t(\bstheta)\hat{A}_t, \textrm{clip}(r_t(\bstheta), 1 - \epsilon, 1 + \epsilon)\hat{A}_t) \right],
    \label{eqn:ppo}
\end{equation}
where $r_{t}(\bstheta)$ is a probability ratio between the previous and updated policy, 
$\textrm{clip}$ is a clipping function between the second and the third arguments, 
and $\hat{A}_{t}$ is an advantage function~\citep{SchulmanJ2015arxiv}.
To calculate the advantage of a state $s_{t}$, 
our model employs a value network, 
$\mlp_{\textrm{val}}([\mu \big( V_t^{\piv} \big), \mu \big(V_t^{\off} \big), \bz])$,
where $\mu(\cdot)$ is a global average function over instances in a given set.

%% file: sections/04_experiments.tex
\section{Experimental Results\label{sec:experiments}}

We evaluate our image-conditioned 3D object assembly of $\ours$ 
in three scenarios: (i) MNIST construction, 
(ii) randomly-assembled object construction, 
and (iii) ModelNet construction.

\begin{figure}[t]
\centering
\subfigure[MNIST Class 0]{
\includegraphics[width=0.28\columnwidth]{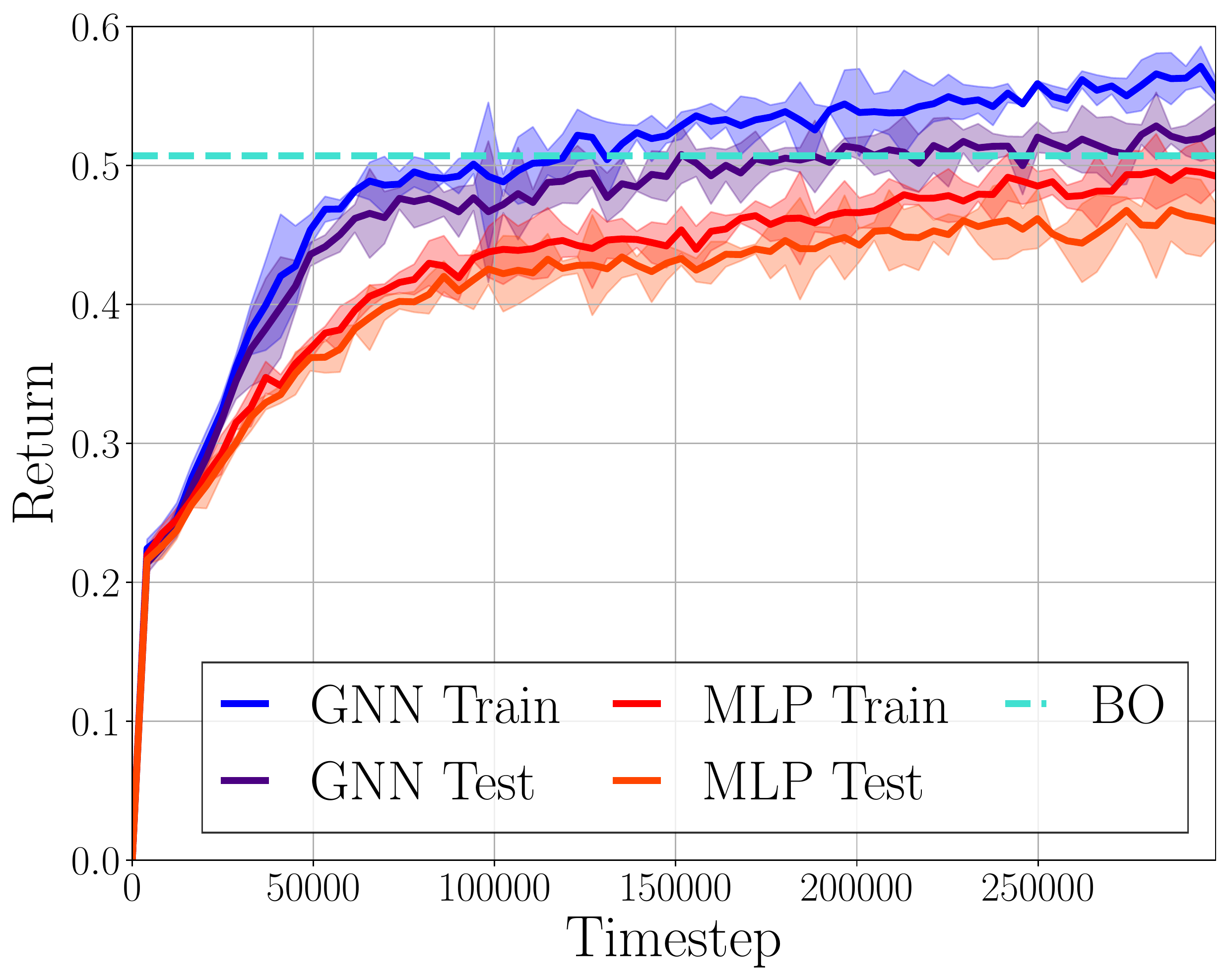}
\label{fig:mnist_easy}
}
\quad
\subfigure[Randomly-Assembled]{
\includegraphics[width=0.28\columnwidth]{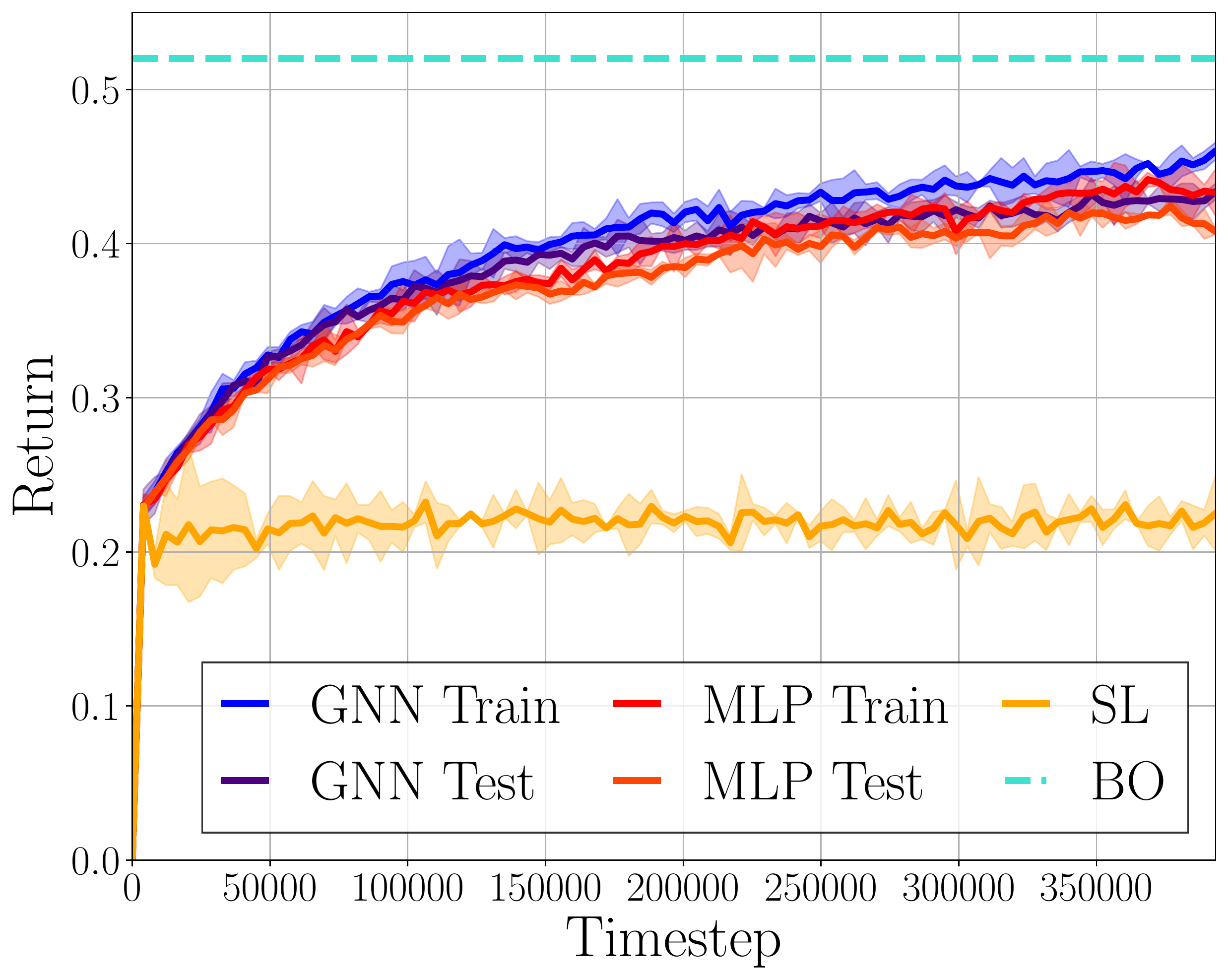}
\label{fig:3dobject_easy}
}
\quad
\subfigure[ModelNet Airplane]{
\includegraphics[width=0.28\columnwidth]{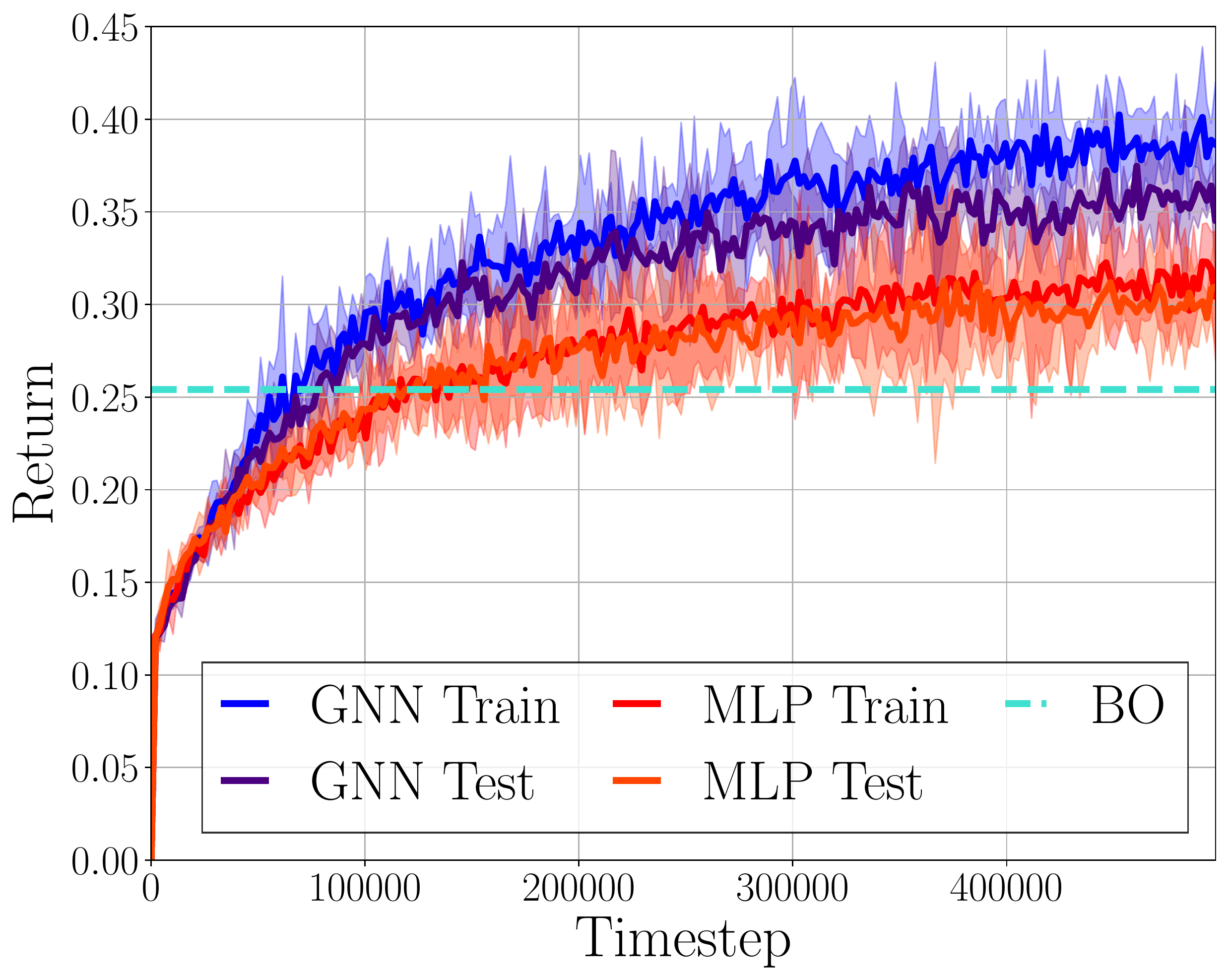}
\label{fig:3dobject_hard}
}
\caption{Episode return vs. timestep in different setups. The return values in training and test episodes are reported by repeating 3 times with different seeds.}
\label{fig:episode_return}
\end{figure}

\begin{figure}[t]
\centering
\subfigure[Pivot ROC]{
    \includegraphics[width=0.225\columnwidth]{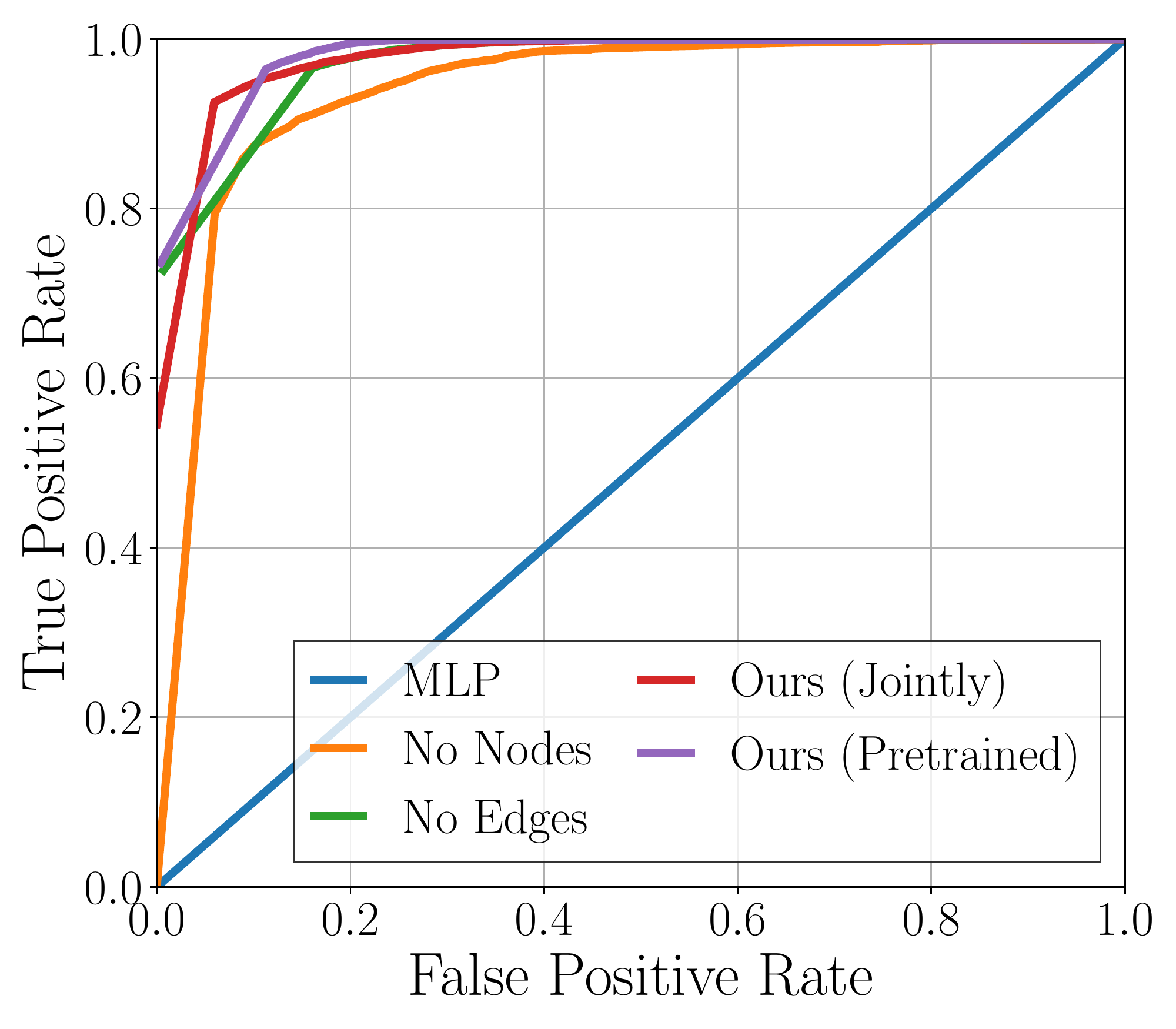}
    \label{fig:roc_pivot}
}
\subfigure[Pivot PR]{
    \includegraphics[width=0.225\columnwidth]{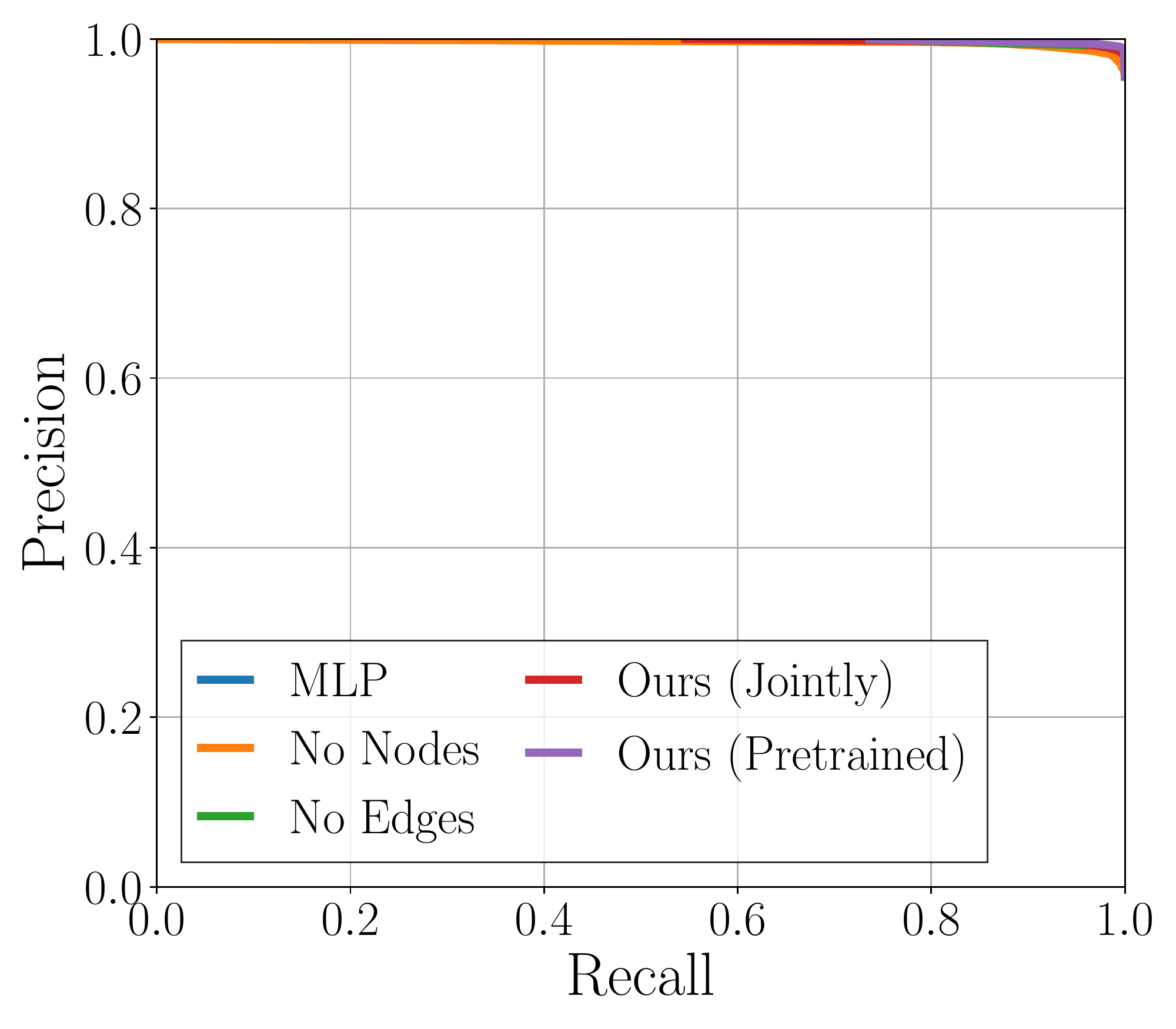}
    \label{fig:pr_pivot}
}
\subfigure[Offset ROC]{
    \includegraphics[width=0.225\columnwidth]{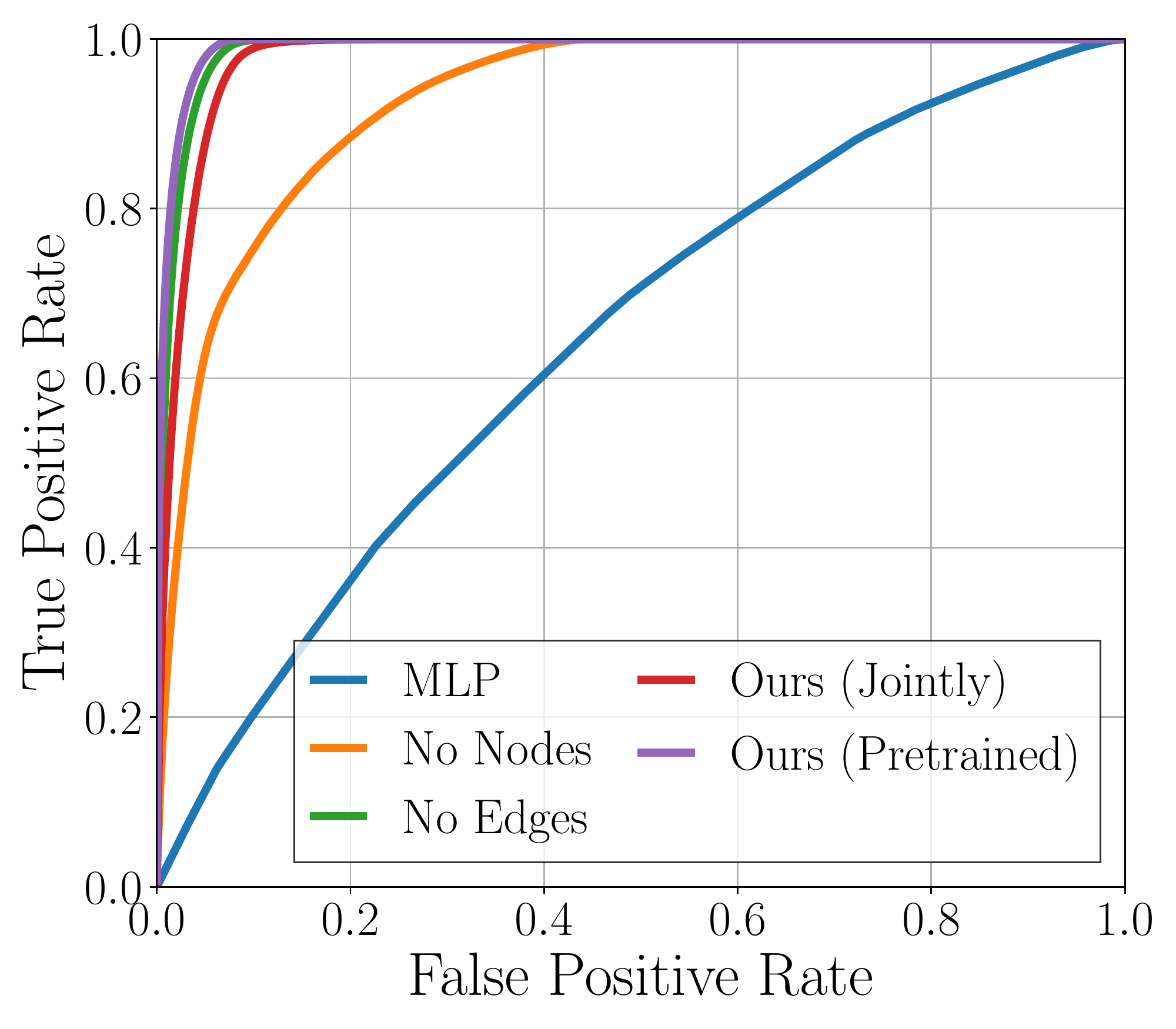}
    \label{fig:roc_offset}
}
\subfigure[Offset PR]{
    \includegraphics[width=0.225\columnwidth]{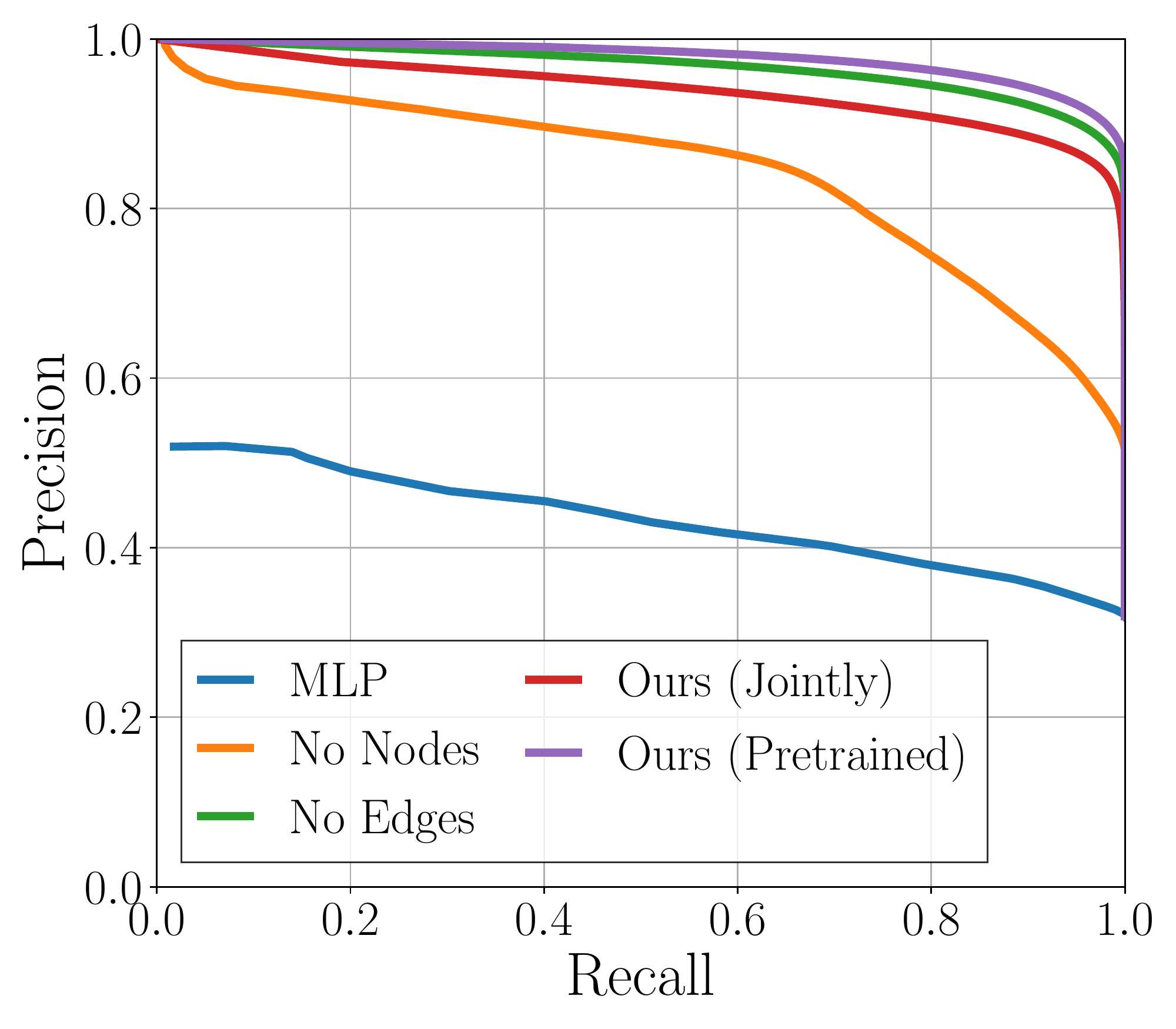}
    \label{fig:pr_offset}
}
\caption{ROC and PR curves for the action validity prediction network. All the results are measured using the test dataset of randomly-assembled objects.}
\label{fig:roc_pr}
\end{figure}

For an evaluation metric, we measure the episode return or IoU between the 
constructed object and the desired target at the end of each episode:
\begin{equation}
\textrm{IoU}(\bC_{N}, \bT) = \frac{\textrm{vol}(\bC_{N} \cap \bT)}{\textrm{vol}(\bC_{N} \cup \bT)},
\end{equation}
where $N$ is the total number of bricks and $\bT$ is the voxel representation of the target object. 
The maximum number of bricks to be placed depends on $\calT$ 
and is pre-defined.
After exhausting the brick budget, 
we terminate the episode and compute the final IoU.
Unless otherwise specified, 
we report the average performance over 3 random seeds, 
each of which is trained for a fixed timestep budget.

\begin{figure}[t]
\centering
\subfigure{
    \includegraphics[width=0.09\columnwidth]{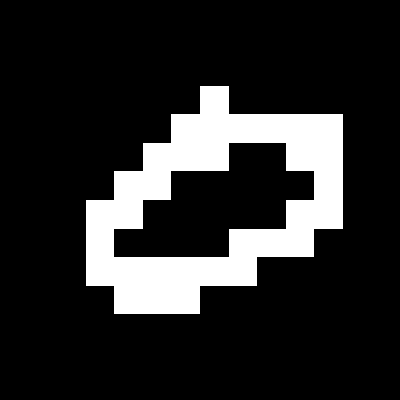}
    \label{fig:mnist_target_0_0}
}
\subfigure{
    \includegraphics[width=0.12\columnwidth]{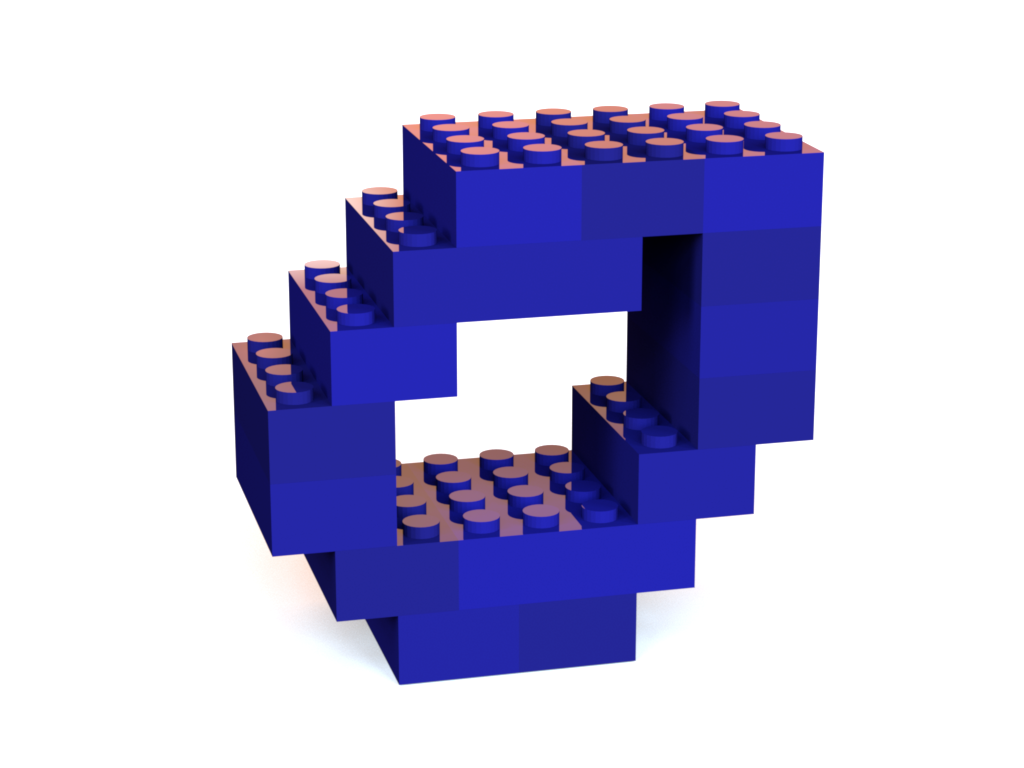}
    \label{fig:mnist_constructed_0_0}
}
\subfigure{
    \includegraphics[width=0.09\columnwidth]{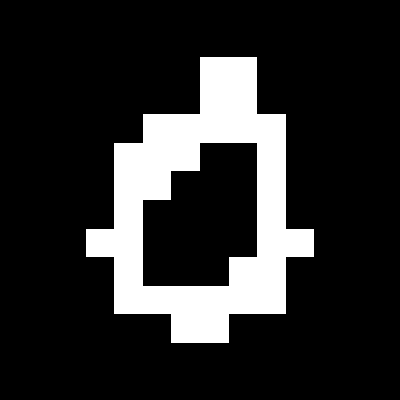}
    \label{fig:mnist_target_0_1}
}
\subfigure{
    \includegraphics[width=0.12\columnwidth]{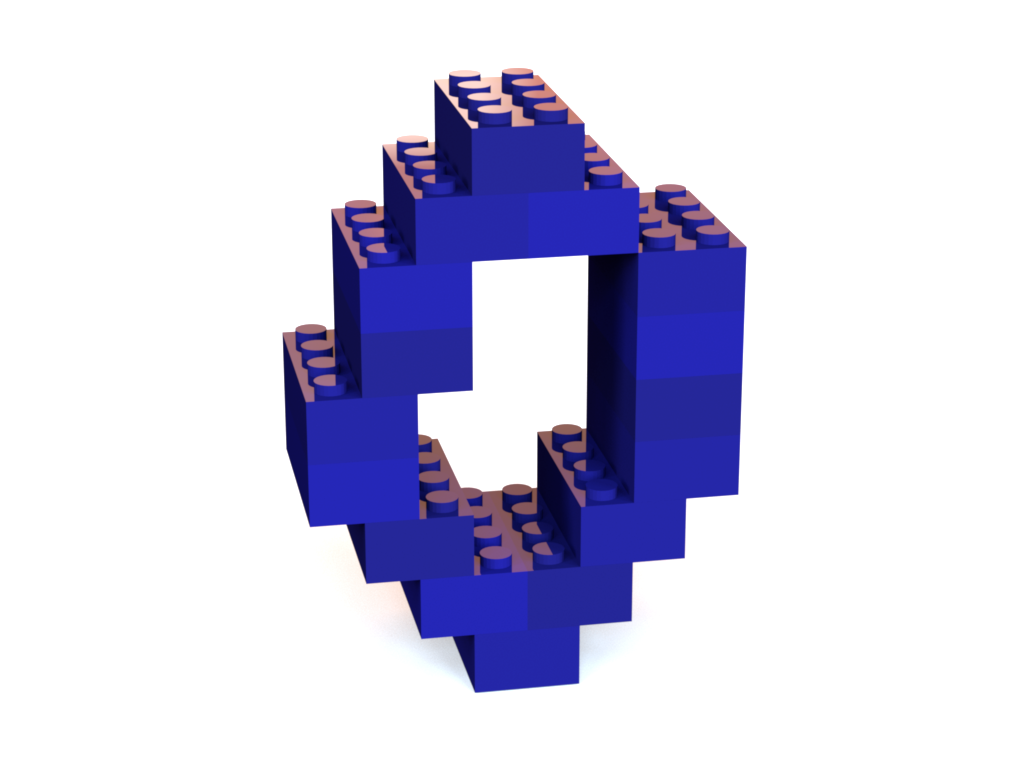}
    \label{fig:mnist_constructed_0_1}
}
\subfigure{
    \includegraphics[width=0.09\columnwidth]{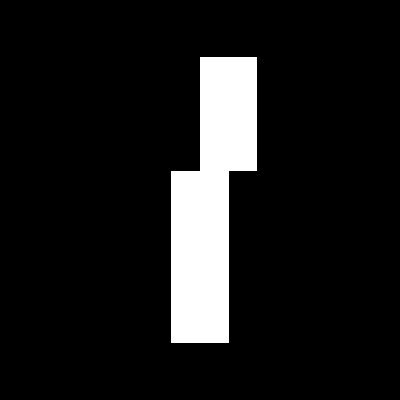}
    \label{fig:mnist_target_1_0}
}
\subfigure{
    \includegraphics[width=0.12\columnwidth]{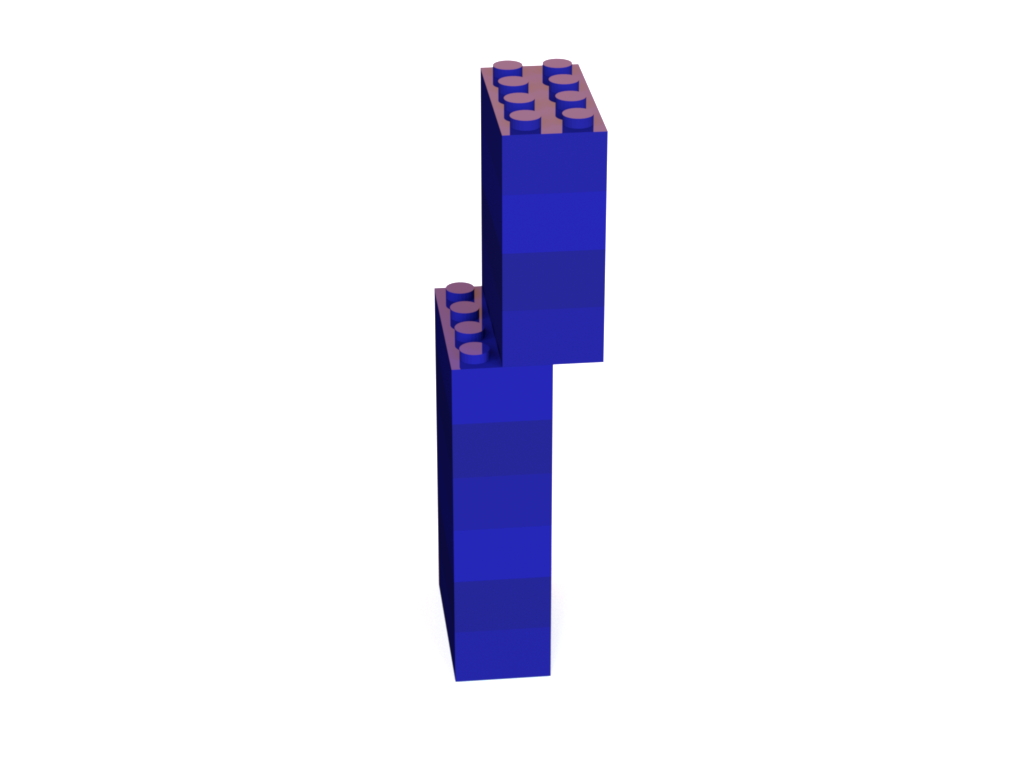}
    \label{fig:mnist_constructed_1_0}
}
\subfigure{
    \includegraphics[width=0.09\columnwidth]{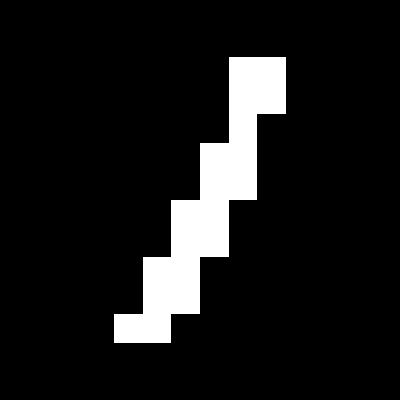}
    \label{fig:mnist_target_1_1}
}
\subfigure{
    \includegraphics[width=0.12\columnwidth]{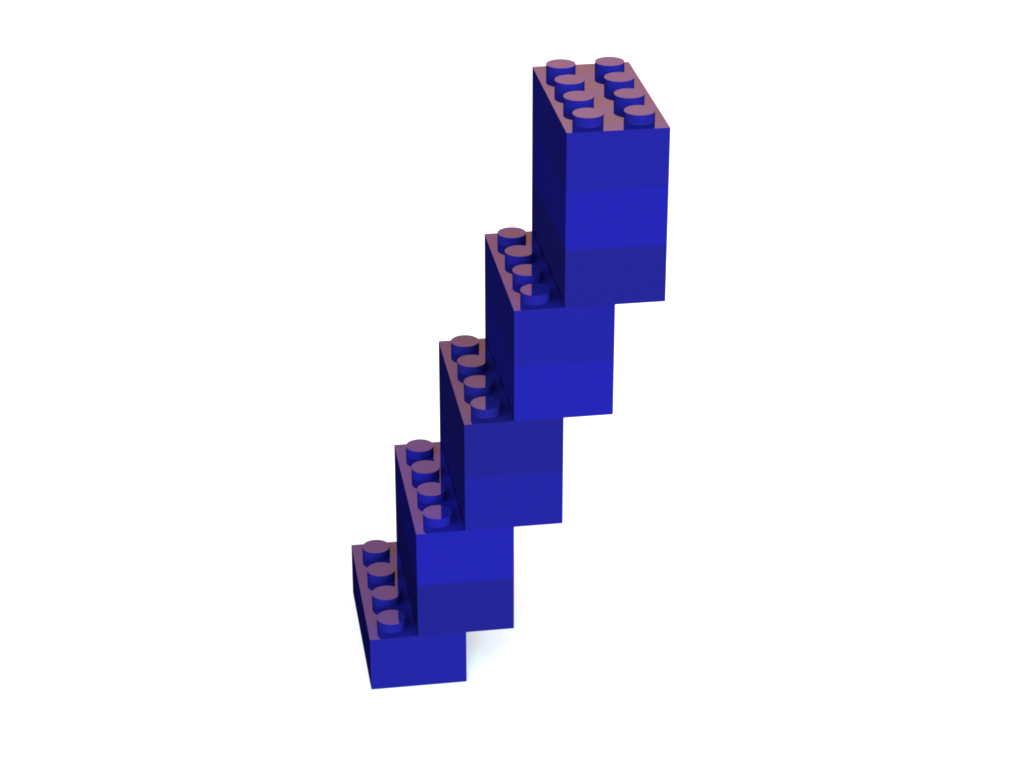}
    \label{fig:mnist_constructed_1_1}
}
\subfigure{
    \includegraphics[width=0.09\columnwidth]{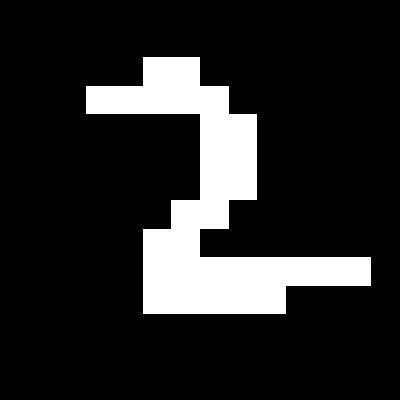}
    \label{fig:mnist_target_2_0}
}
\subfigure{
    \includegraphics[width=0.12\columnwidth]{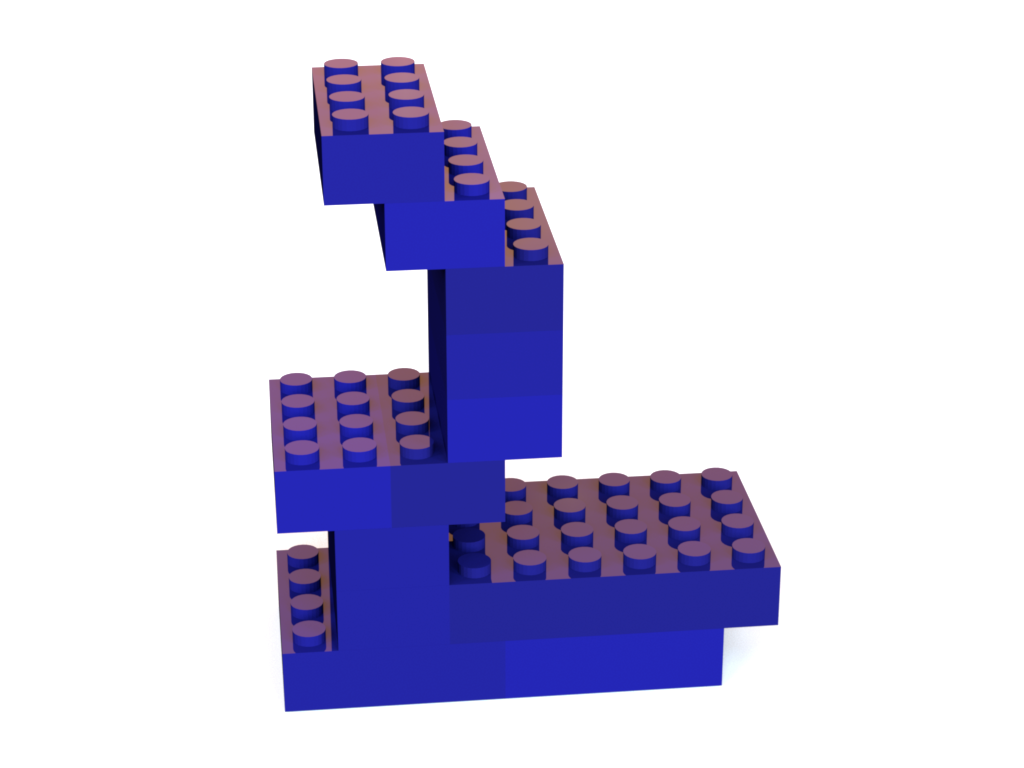}
    \label{fig:mnist_constructed_2_0}
}
\subfigure{
    \includegraphics[width=0.09\columnwidth]{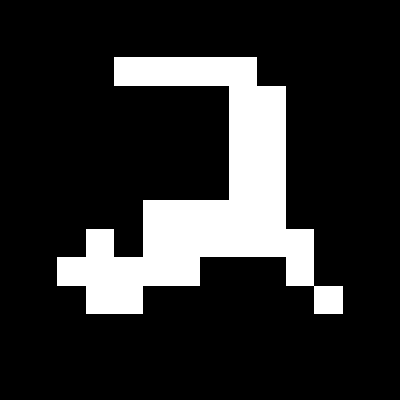}
    \label{fig:mnist_target_2_1}
}
\subfigure{
    \includegraphics[width=0.12\columnwidth]{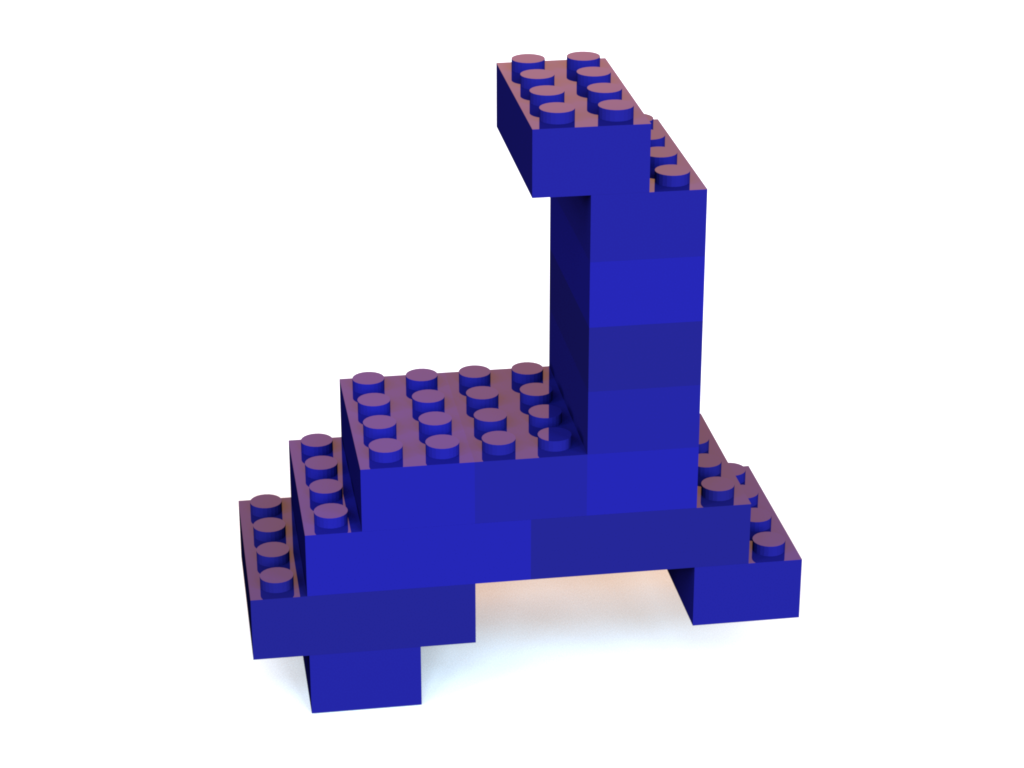}
    \label{fig:mnist_constructed_2_1}
}
\subfigure{
    \includegraphics[width=0.09\columnwidth]{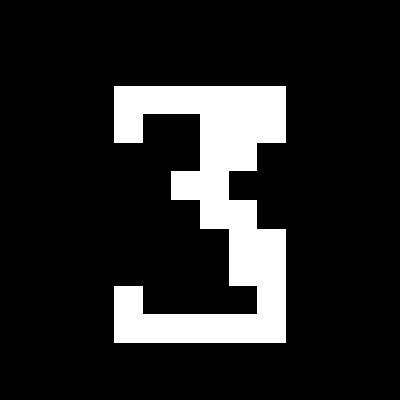}
    \label{fig:mnist_target_3_0}
}
\subfigure{
    \includegraphics[width=0.12\columnwidth]{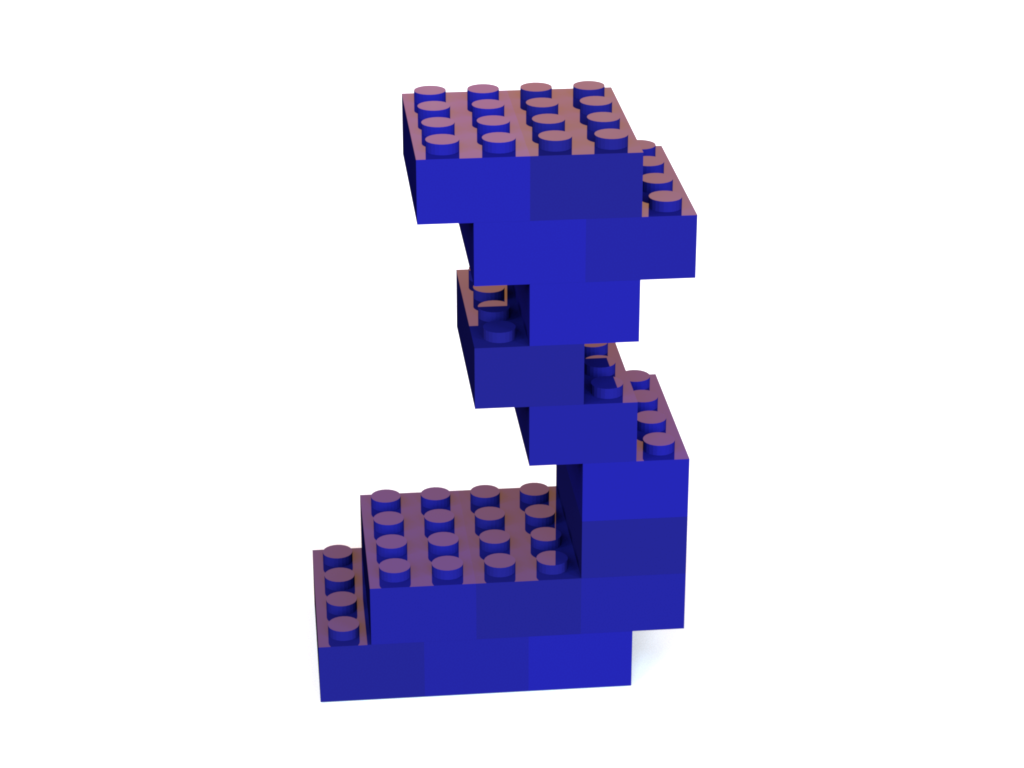}
    \label{fig:mnist_constructed_3_0}
}
\subfigure{
    \includegraphics[width=0.09\columnwidth]{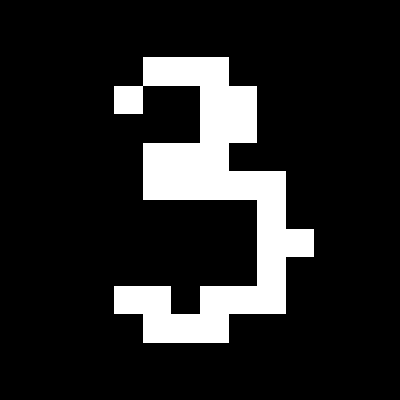}
    \label{fig:mnist_target_3_1}
}
\subfigure{
    \includegraphics[width=0.12\columnwidth]{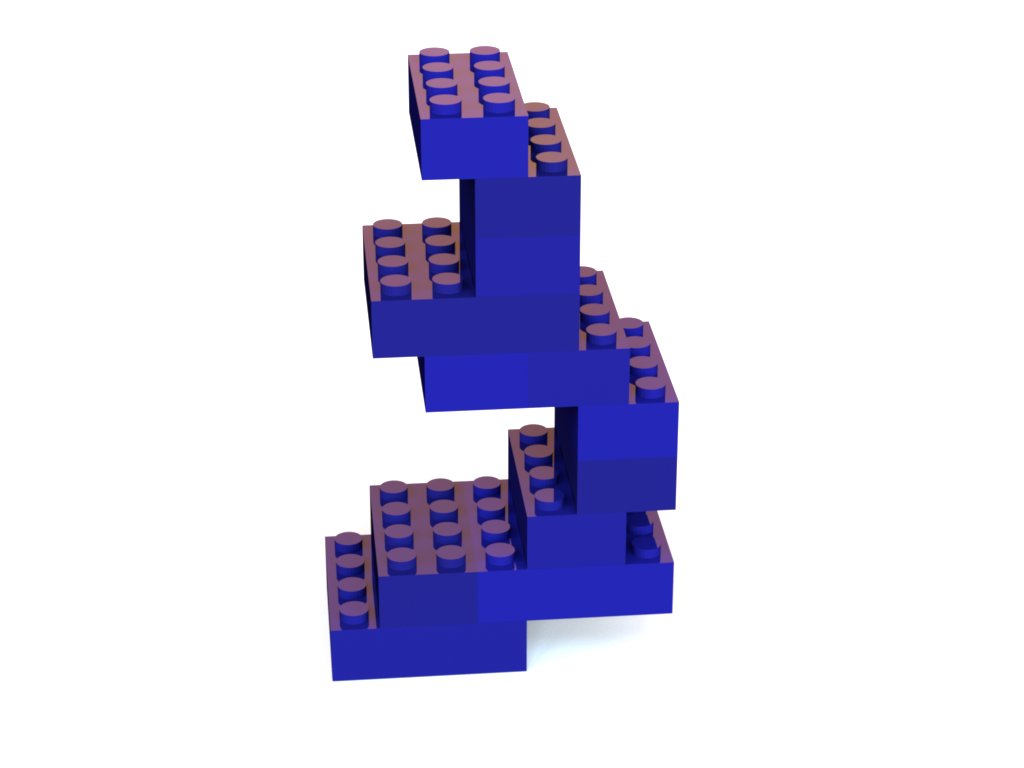}
    \label{fig:mnist_constructed_3_1}
}
\subfigure{
    \includegraphics[width=0.09\columnwidth]{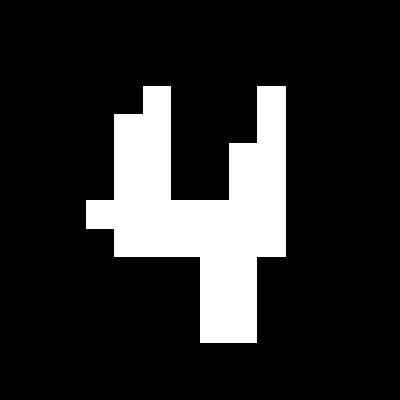}
    \label{fig:mnist_target_4_0}
}
\subfigure{
    \includegraphics[width=0.12\columnwidth]{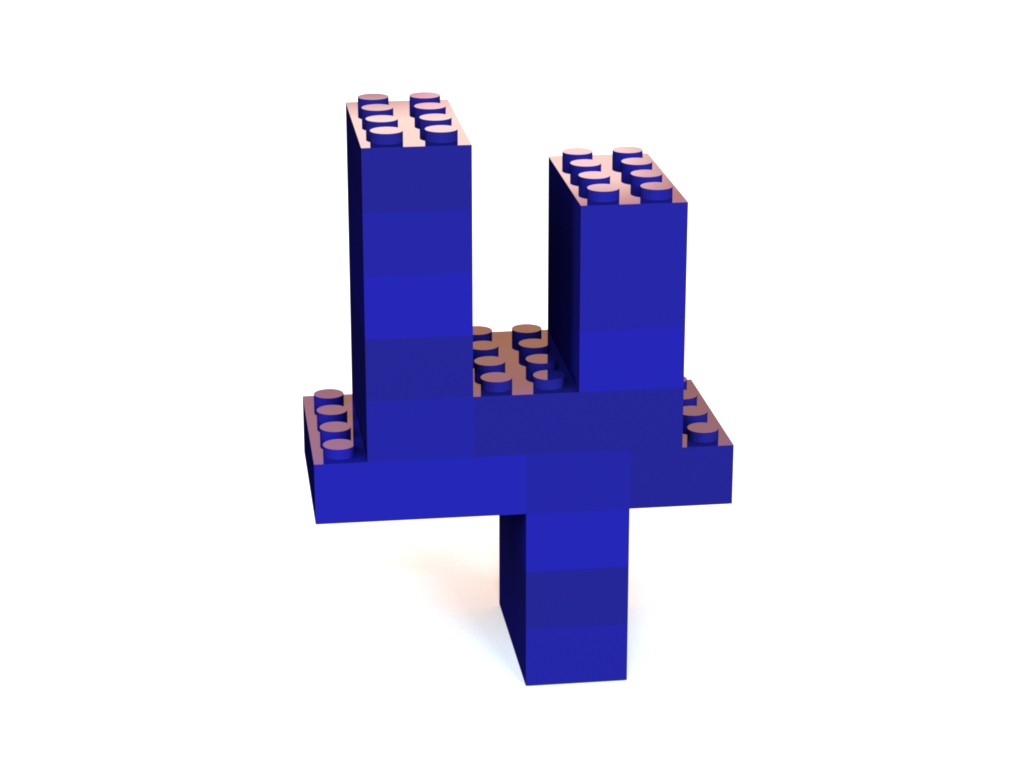}
    \label{fig:mnist_constructed_4_0}
}
\subfigure{
    \includegraphics[width=0.09\columnwidth]{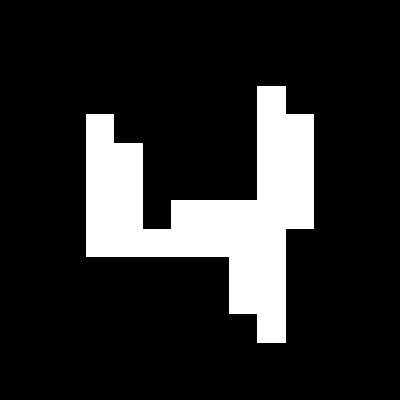}
    \label{fig:mnist_target_4_1}
}
\subfigure{
    \includegraphics[width=0.12\columnwidth]{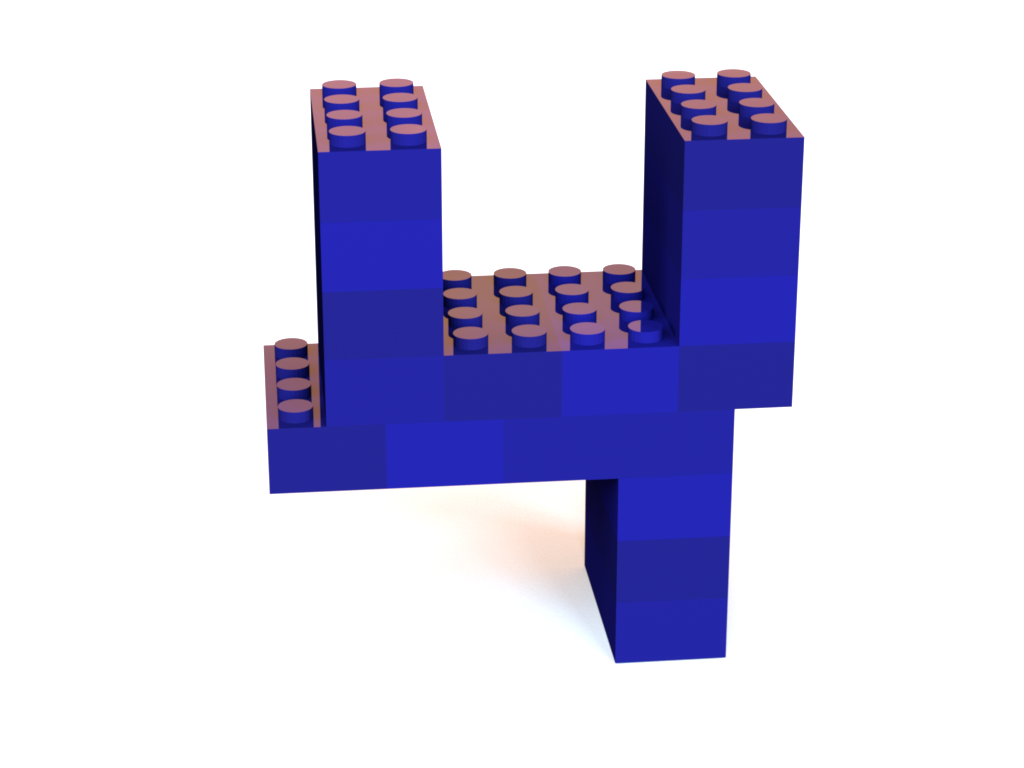}
    \label{fig:mnist_constructed_4_1}
}
\subfigure{
    \includegraphics[width=0.09\columnwidth]{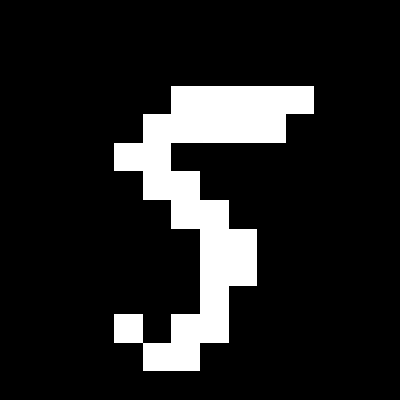}
    \label{fig:mnist_target_5_0}
}
\subfigure{
    \includegraphics[width=0.12\columnwidth]{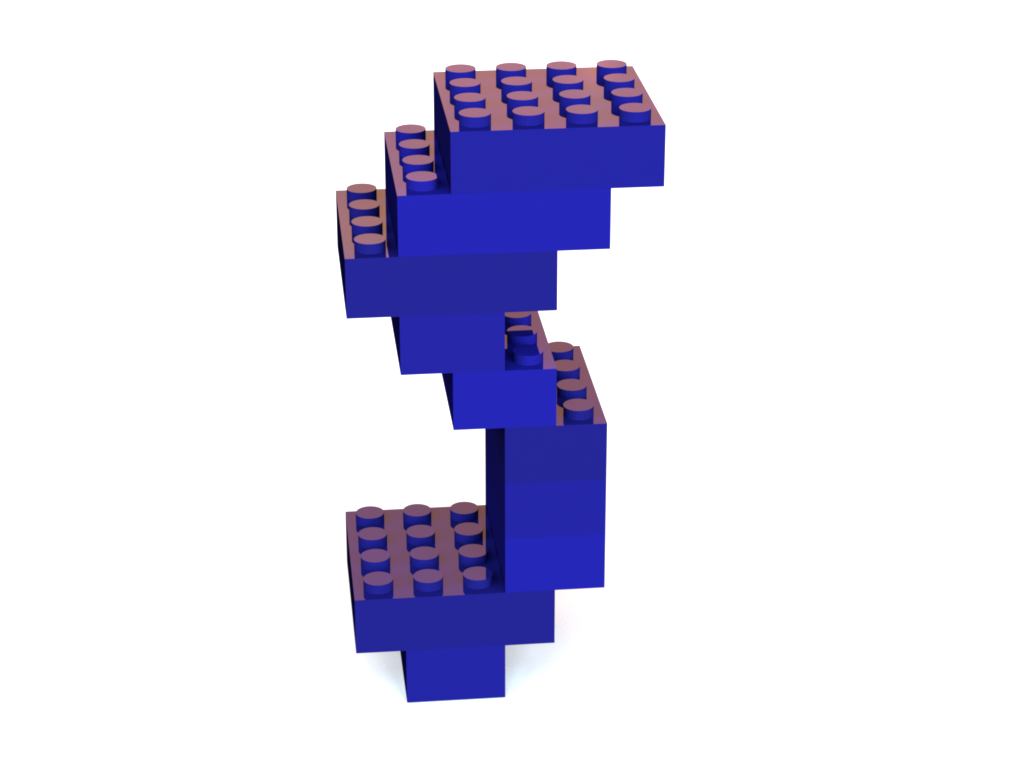}
    \label{fig:mnist_constructed_5_0}
}
\subfigure{
    \includegraphics[width=0.09\columnwidth]{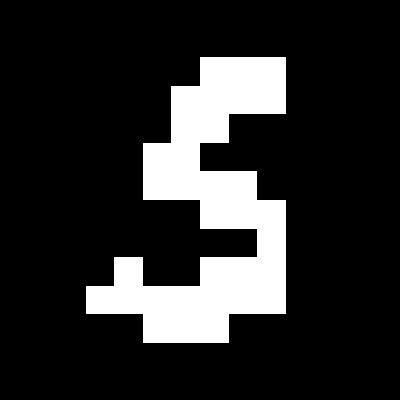}
    \label{fig:mnist_target_5_1}
}
\subfigure{
    \includegraphics[width=0.12\columnwidth]{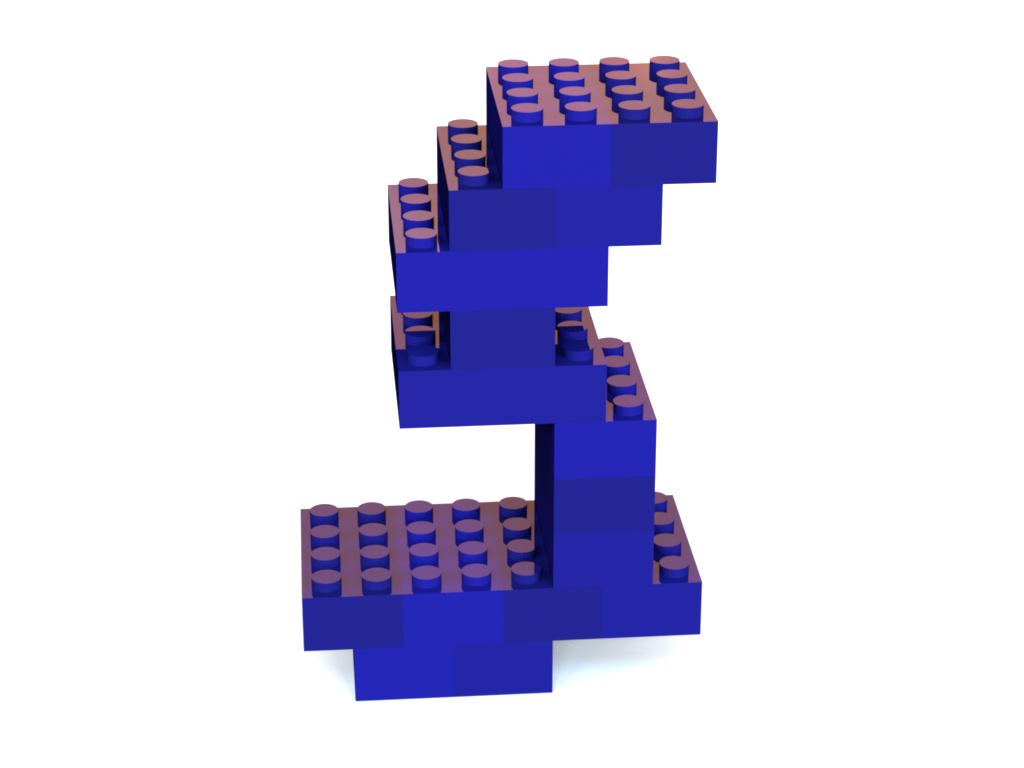}
    \label{fig:mnist_constructed_5_1}
}
\subfigure{
    \includegraphics[width=0.09\columnwidth]{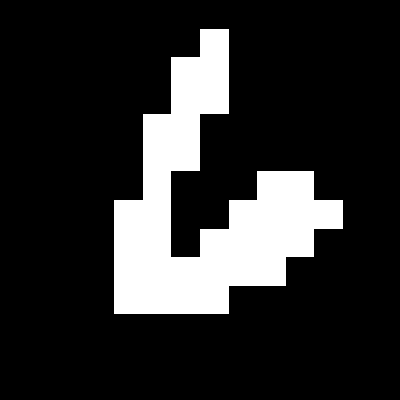}
    \label{fig:mnist_target_6_0}
}
\subfigure{
    \includegraphics[width=0.12\columnwidth]{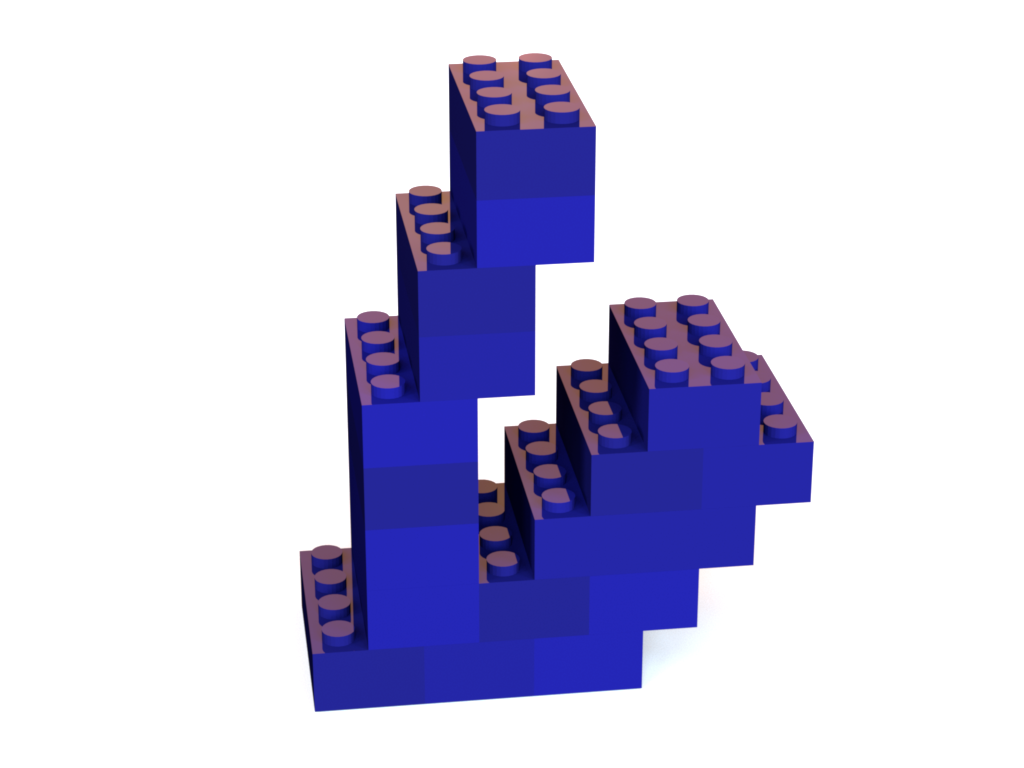}
    \label{fig:mnist_constructed_6_0}
}
\subfigure{
    \includegraphics[width=0.09\columnwidth]{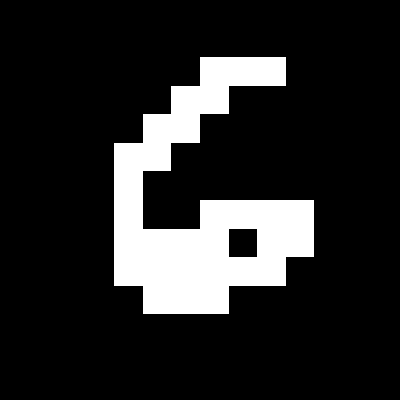}
    \label{fig:mnist_target_6_1}
}
\subfigure{
    \includegraphics[width=0.12\columnwidth]{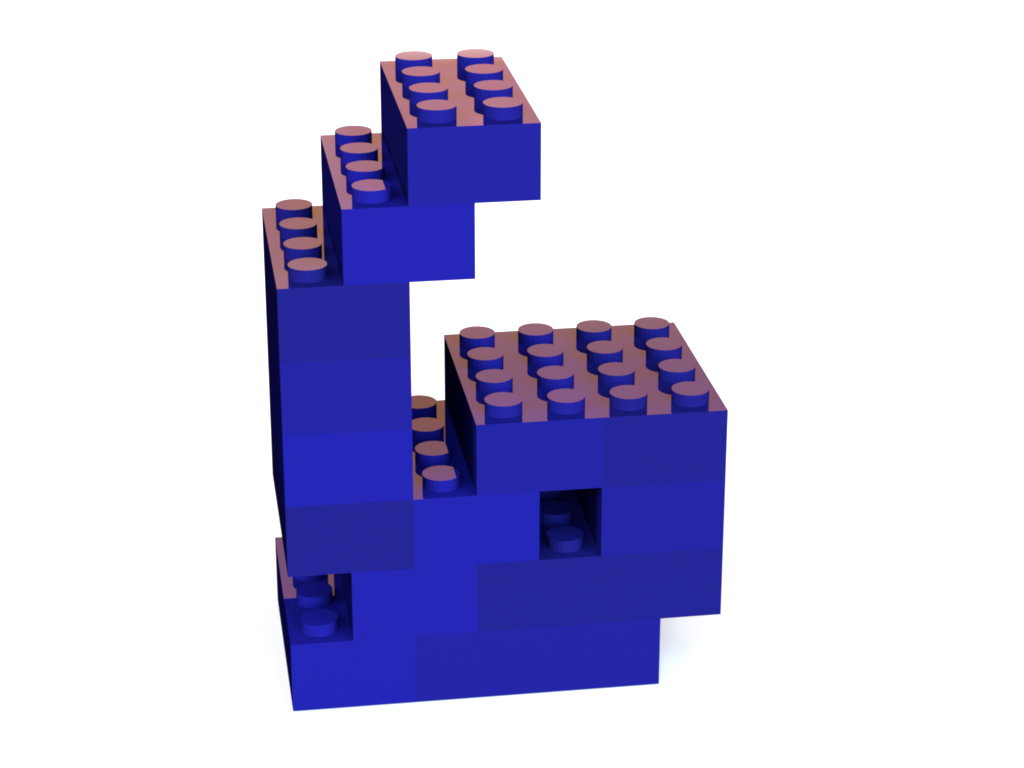}
    \label{fig:mnist_constructed_6_1}
}
\subfigure{
    \includegraphics[width=0.09\columnwidth]{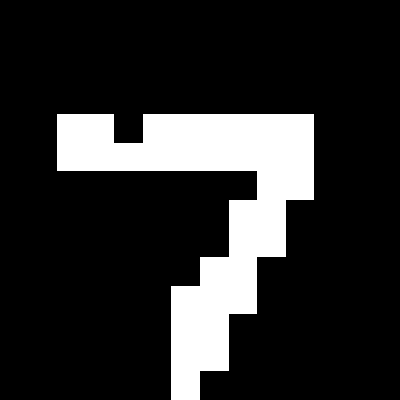}
    \label{fig:mnist_target_7_0}
}
\subfigure{
    \includegraphics[width=0.12\columnwidth]{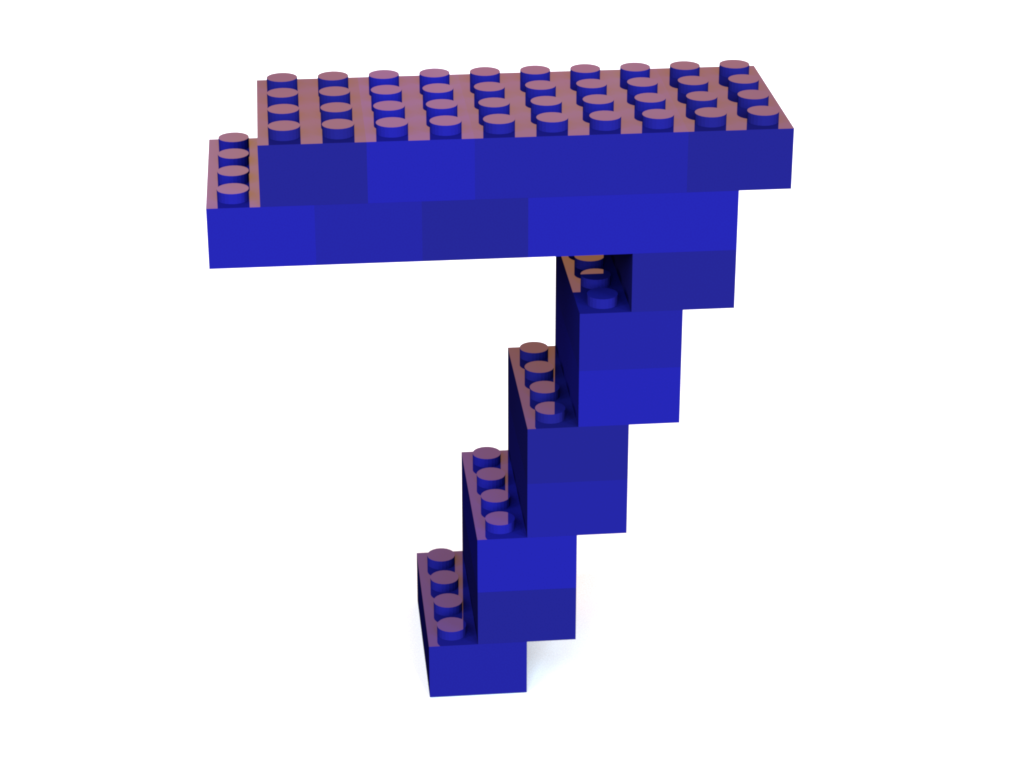}
    \label{fig:mnist_constructed_7_0}
}
\subfigure{
    \includegraphics[width=0.09\columnwidth]{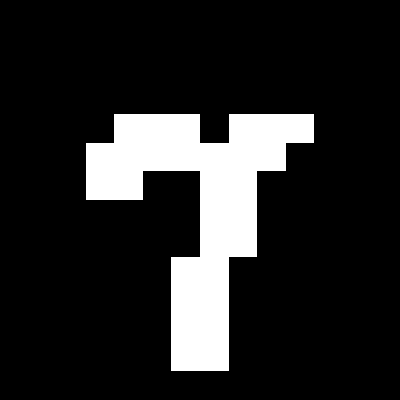}
    \label{fig:mnist_target_7_1}
}
\subfigure{
    \includegraphics[width=0.12\columnwidth]{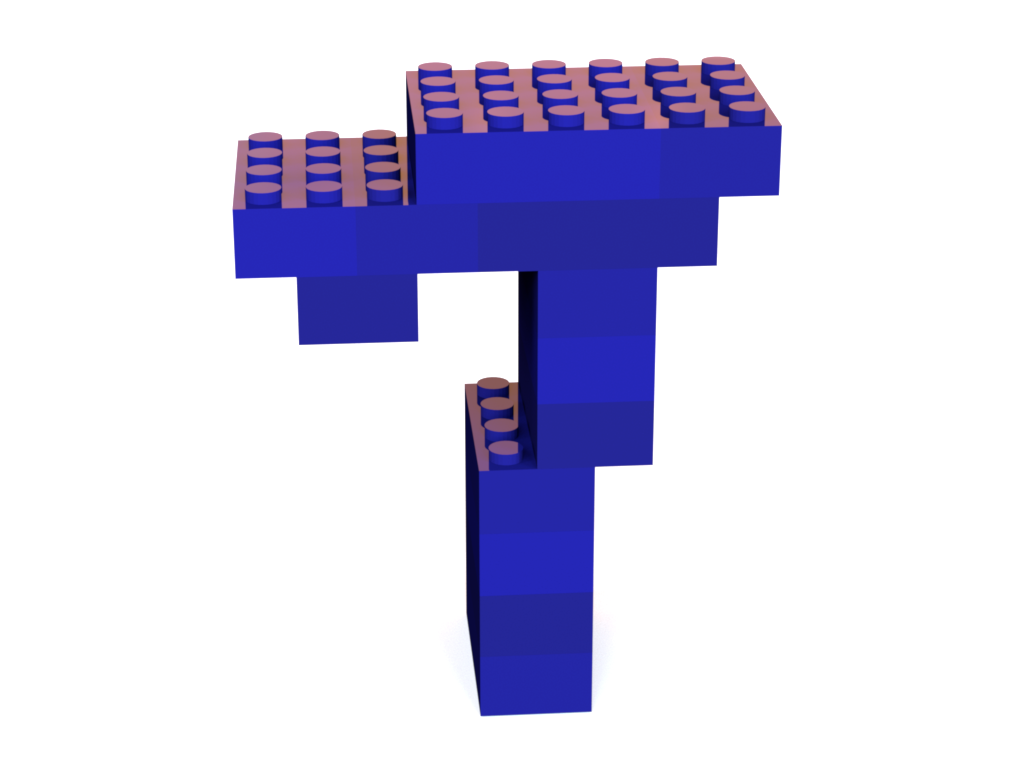}
    \label{fig:mnist_constructed_7_1}
}
\subfigure{
    \includegraphics[width=0.09\columnwidth]{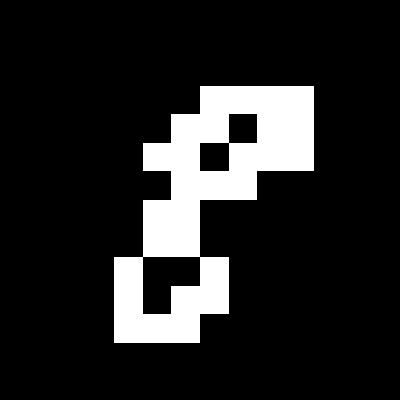}
    \label{fig:mnist_target_8_0}
}
\subfigure{
    \includegraphics[width=0.12\columnwidth]{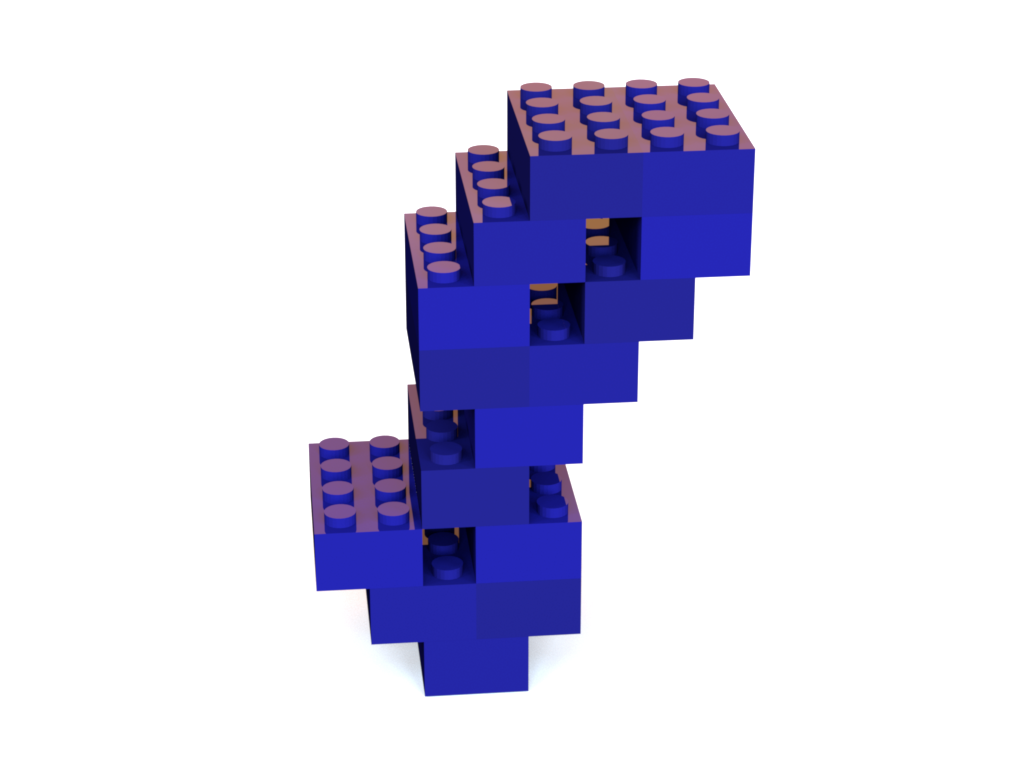}
    \label{fig:mnist_constructed_8_0}
}
\subfigure{
    \includegraphics[width=0.09\columnwidth]{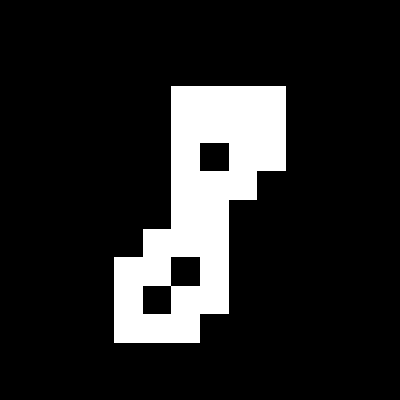}
    \label{fig:mnist_target_8_1}
}
\subfigure{
    \includegraphics[width=0.12\columnwidth]{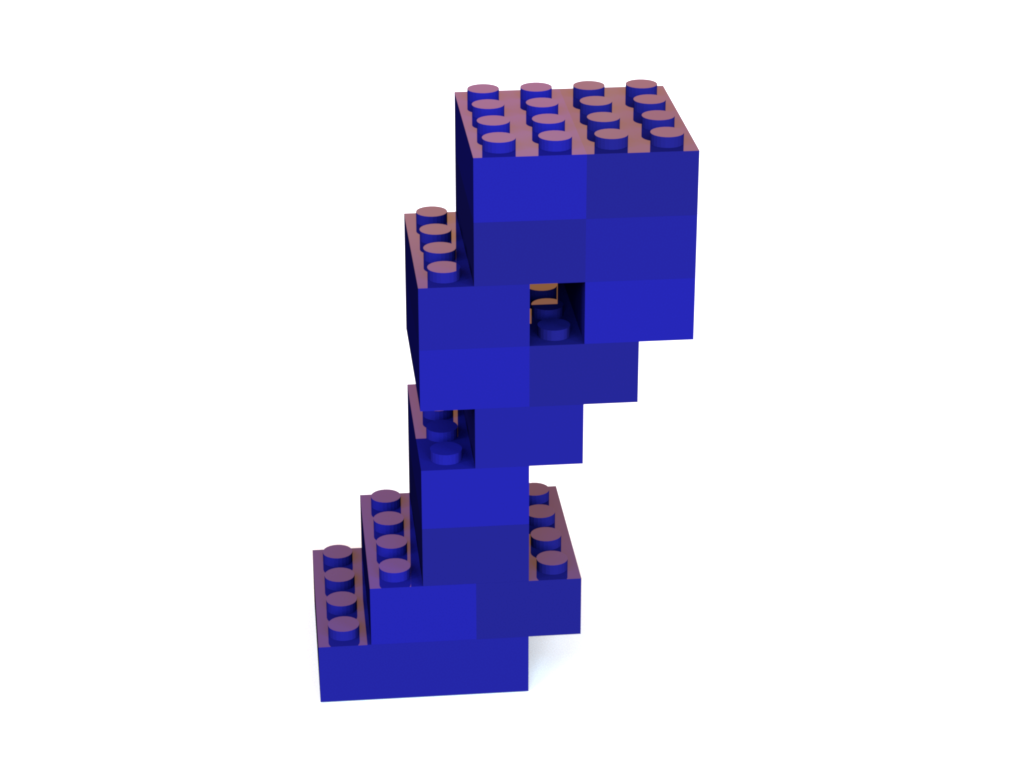}
    \label{fig:mnist_constructed_8_1}
}
\subfigure{
    \includegraphics[width=0.09\columnwidth]{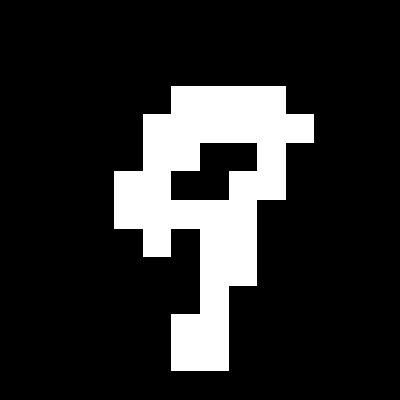}
    \label{fig:mnist_target_9_0}
}
\subfigure{
    \includegraphics[width=0.12\columnwidth]{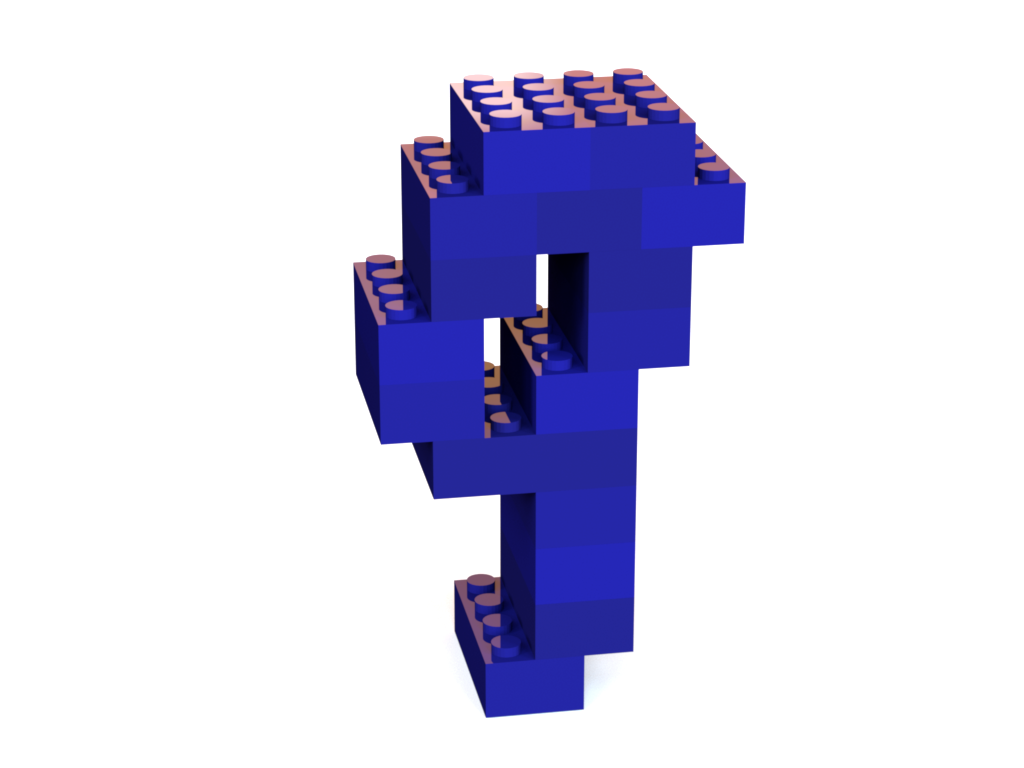}
    \label{fig:mnist_constructed_9_0}
}
\subfigure{
    \includegraphics[width=0.09\columnwidth]{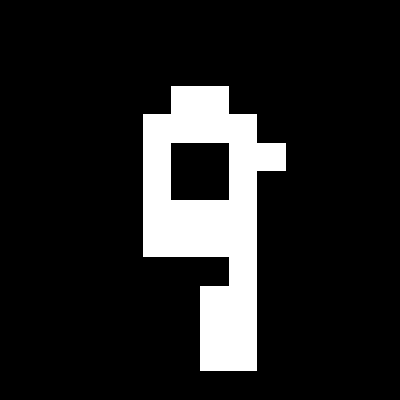}
    \label{fig:mnist_target_9_1}
}
\subfigure{
    \includegraphics[width=0.12\columnwidth]{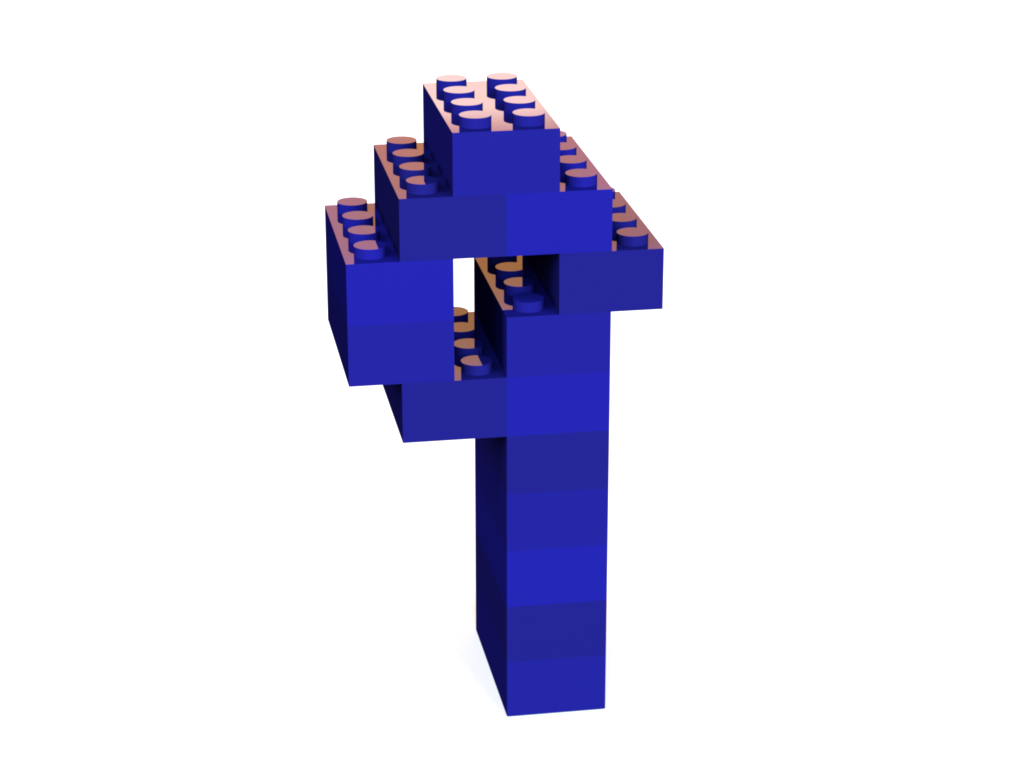}
    \label{fig:mnist_constructed_9_1}
}
\caption{Qualitative results on MNIST construction. Our model is separately trained on each class, and target images are unseen during training.}
\label{fig:mnist}
\end{figure}

To show the effectiveness of our method, we first analyze our action validity prediction network and test other baseline methods and $\ours$ in different scenarios.
In all construction tasks, we compare $\ours$ to the MLP-based model 
where all GNNs are replaced with MLPs, 
and to the Bayesian optimization-based approach (BO) 
that sequentially optimizes the step-wise reward in terms of IoU 
to search for an optimal construction sequence.
As presented in~\tabref{tab:comparisons}, 
BO uses exact volumetric information for both training and test target objects 
because it cannot assemble an object with only partial information. 
For each scenario, the episode returns of the BO model are averaged 
over both training and test datasets.
In addition, we compare $\ours$ to the supervised learning method trained 
with the cross-entropy loss between predicted and ground-truth sequences, 
specifically, in the randomly-assembled object construction.
Since sequence-level supervision is used, 
the performance of the supervised learning method 
is only measured on the test dataset.
Details can be found in the supplementary material.

\paragraph{Action Validity Prediction Network.}
We test our action validity prediction network by creating training and test datasets.
The training dataset is composed of 200,000 brick combinations and their ground-truth
action validity, and the test dataset is composed of 30,000 brick combinations and their ground-truth action validity.
Importantly, the range of the size of a brick combination in the training dataset is $[1, 20]$, 
and the range in the test dataset is $[1, 30]$.
While the test dataset contains larger brick combinations than
the training dataset, the performance of the action validity prediction network in terms of
precision and recall is satisfactory, predicting reliable validity confidences even for actions in unseen ranges,
as presented in~\figref{fig:roc_pr}.
Our GNN outperforms MLP as well as 
GNN baselines, which do not have either node features or edge features.
In addition, the pretrained network, which is reusable in different scenarios, is slightly better than the action validity prediction network 
that is jointly trained with training episodes.
See the supplementary material for more details of the action validity prediction network.

\paragraph{MNIST Construction.}
In each episode, an agent is provided with an image from the MNIST dataset 
and is provided to create a 3D object resembling the digit target.
Similar to~\citep{SalakhutdinovRR2008icml}, we binarize the MNIST dataset 
to convert a real-valued number to either 0 or 1, 
for brevity of the calculation of IoU.
To create a 3D target object with a 2D MNIST image, we first rescale an image 
to half of the original size and then expand an image along the channel dimension,
in order to assemble with $\twobyfour$ bricks, i.e.,
an image of size $28 \times 28$ is transformed to a 3D object of size $14 \times 14 \times 4$.
Furthermore, we limit possible offset candidates to 6 different types 
of which the values according to the channel dimension are fixed to the same value.
Training and test datasets are established by choosing one of the ten classes 
in the binarized MNIST and splitting images from that class.
In particular, 500 images from one of available classes 
are chosen, further divided into 400 samples 
for a training dataset and 100 samples for a test dataset.

Due to a space limit, we only report the average reward performance on class 0 in~\figref{fig:mnist_easy}.
Results for other classes are available in the supplementary material.
The gap between the training and test sets on episode returns is marginal,
which implies that our model $\ours$ generalizes to unseen targets well.
In addition, both our training and test results are better compared to both the BO and MLP-based models.
We visualize constructed objects for the test dataset of classes 0 through 9 in~\figref{fig:mnist}. More qualitative results are also provided in the supplementary material. In general, our agent successfully constructs objects of unseen instances.
This can be understood as our agent catches distinctive details 
in the target information and reflects in a construction process.

\begin{figure}[t]
\centering
\subfigure{
    \includegraphics[width=0.125\columnwidth]{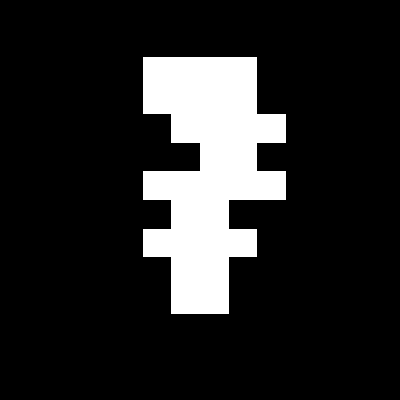}
    \includegraphics[width=0.125\columnwidth]{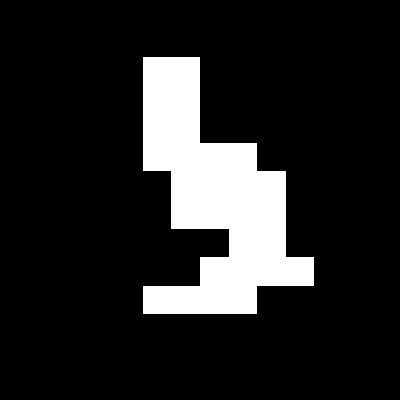}
    \includegraphics[width=0.125\columnwidth]{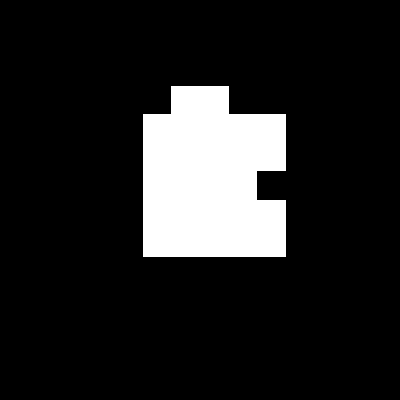}
    \label{fig:artificial_target_1}
}
\subfigure{
    \includegraphics[width=0.165\columnwidth]{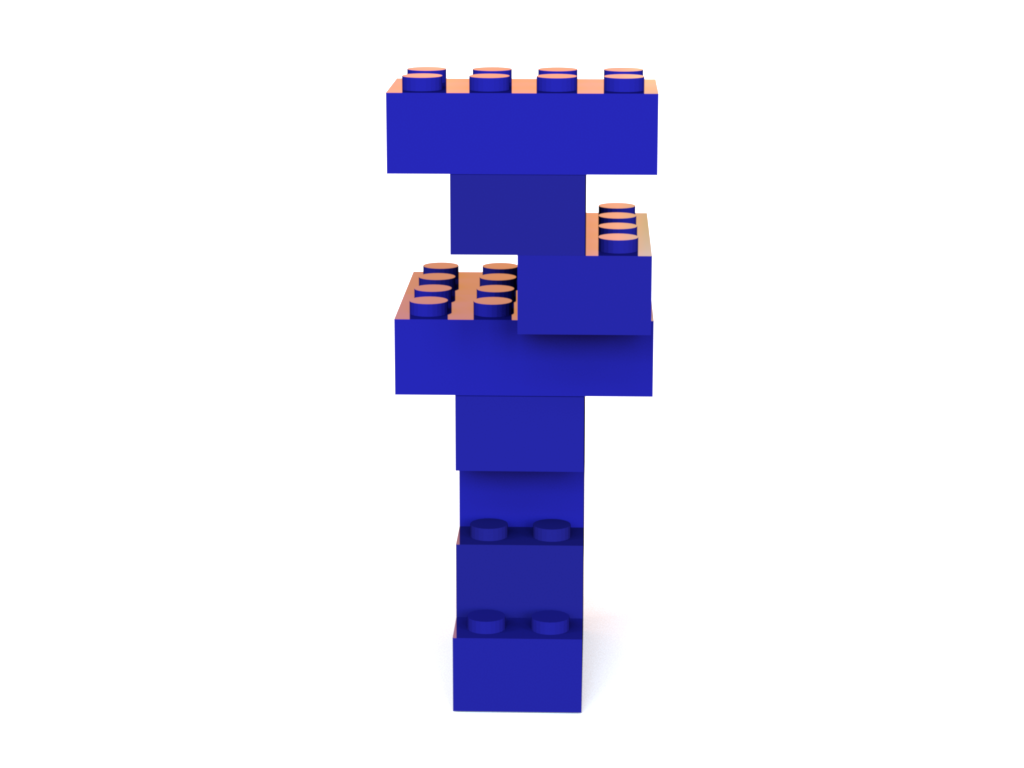}
    \includegraphics[width=0.165\columnwidth]{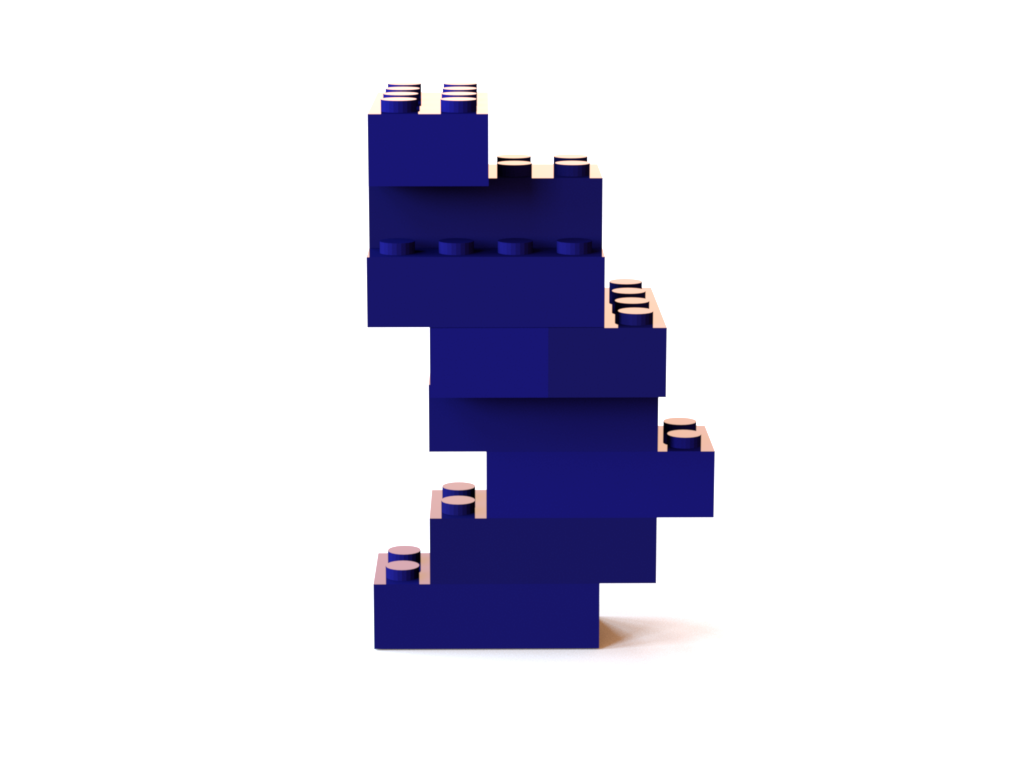}
    \includegraphics[width=0.165\columnwidth]{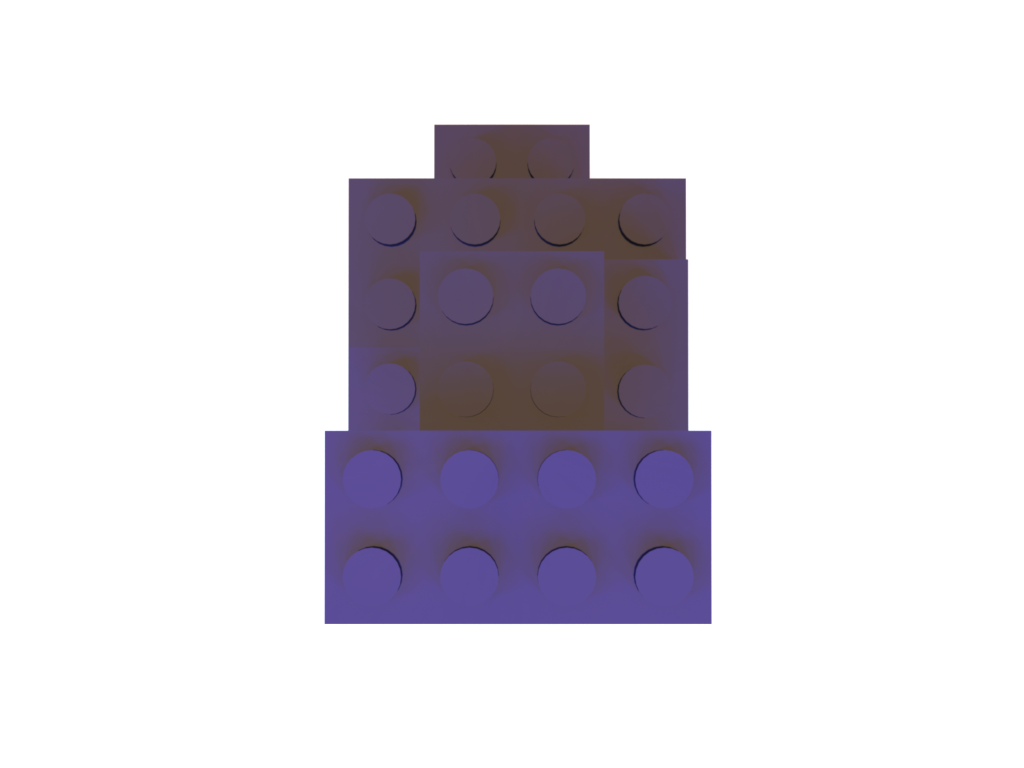}
    \label{fig:artificial_constructed_1}
}
\subfigure{
    \includegraphics[width=0.125\columnwidth]{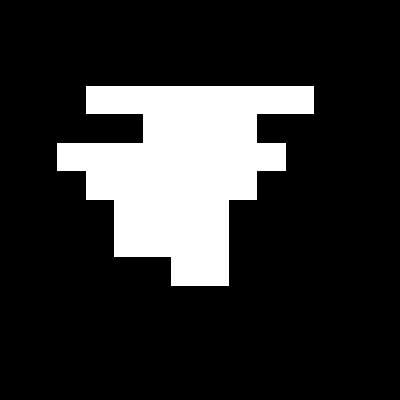}
    \includegraphics[width=0.125\columnwidth]{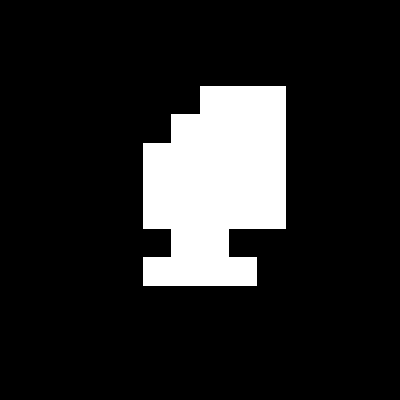}
    \includegraphics[width=0.125\columnwidth]{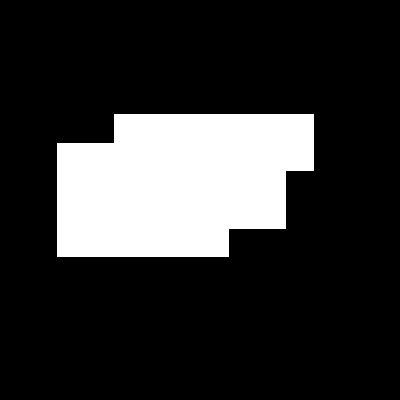}
    \label{fig:artificial_target_2}
}
\subfigure{
    \includegraphics[width=0.165\columnwidth]{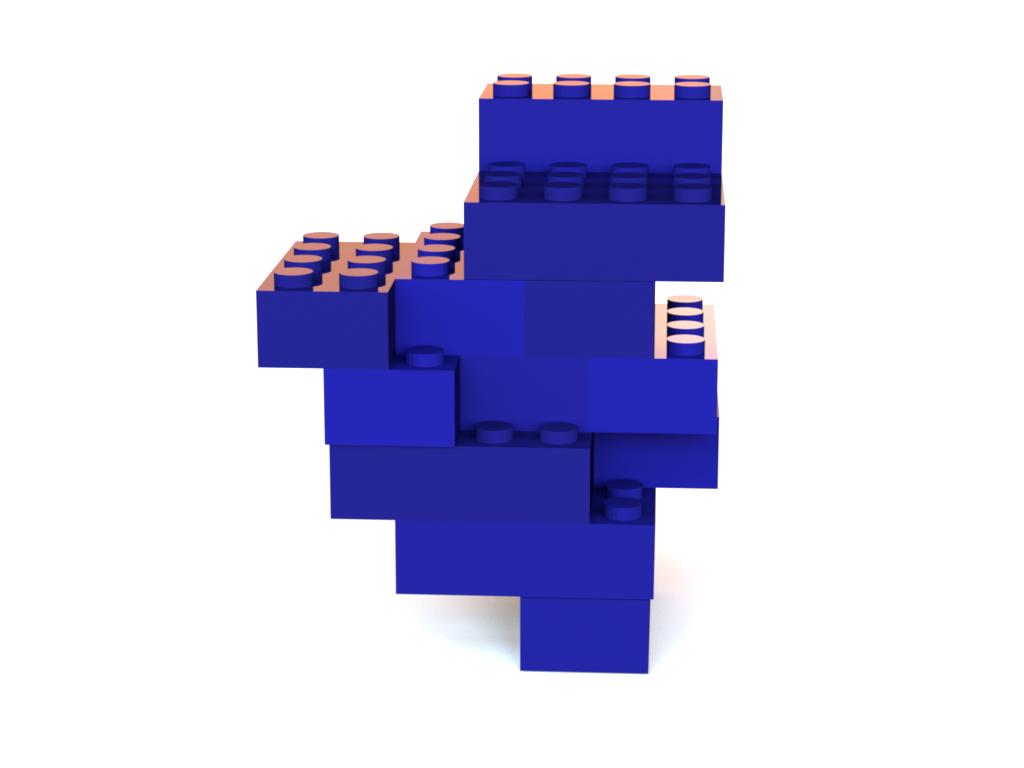}
    \includegraphics[width=0.165\columnwidth]{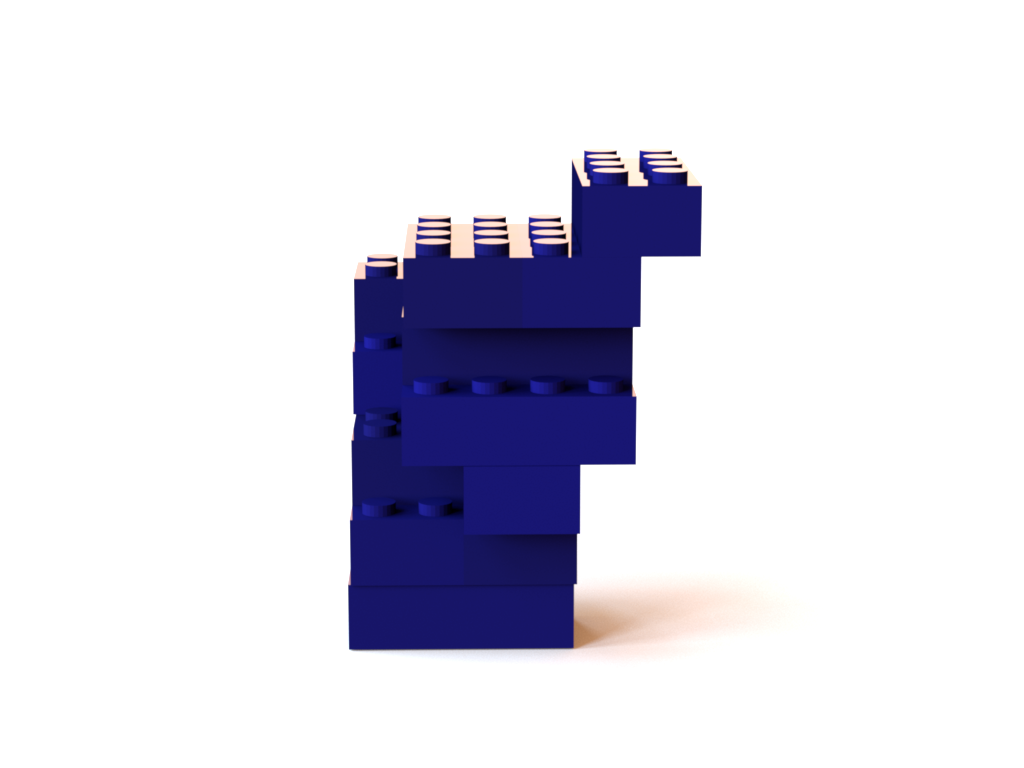}
    \includegraphics[width=0.165\columnwidth]{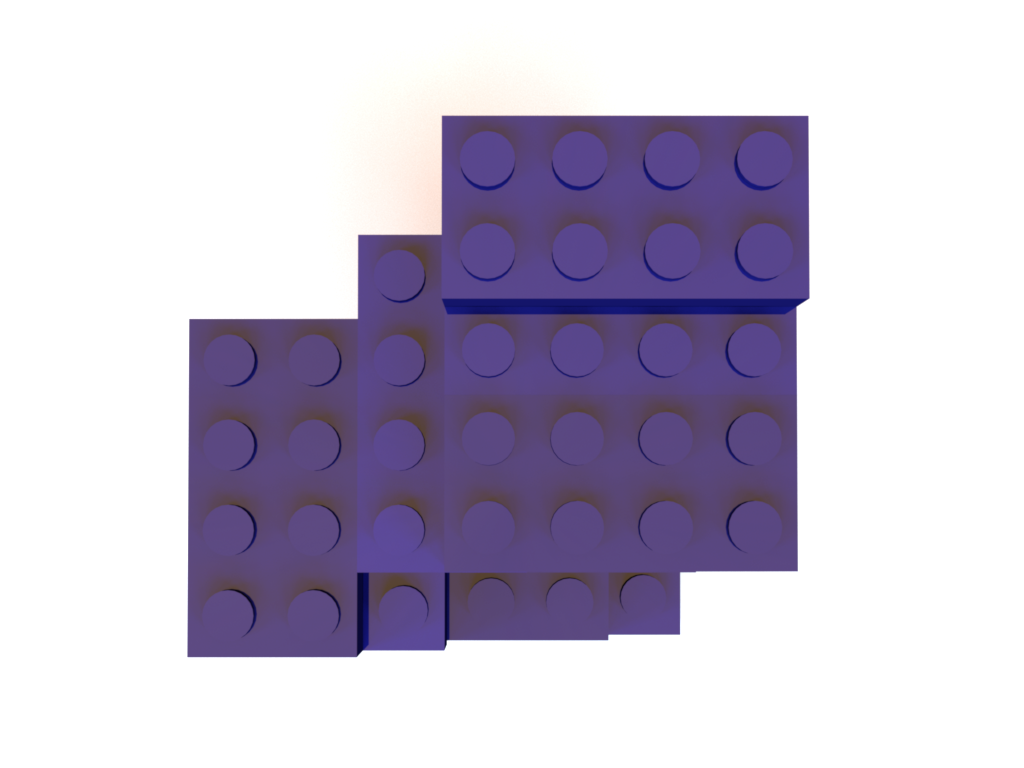}
    \label{fig:artificial_constructed_2}
}
\subfigure{
    \includegraphics[width=0.125\columnwidth]{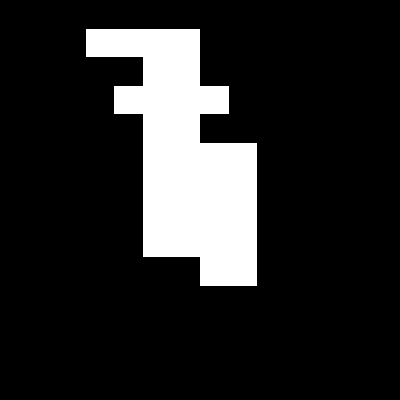}
    \includegraphics[width=0.125\columnwidth]{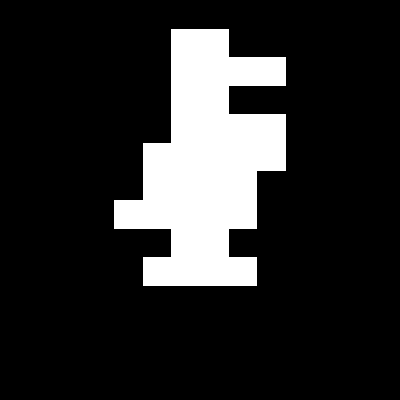}
    \includegraphics[width=0.125\columnwidth]{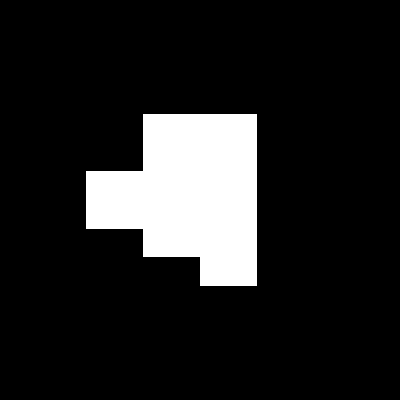}
    \label{fig:artificial_target_3}
}
\subfigure{
    \includegraphics[width=0.165\columnwidth]{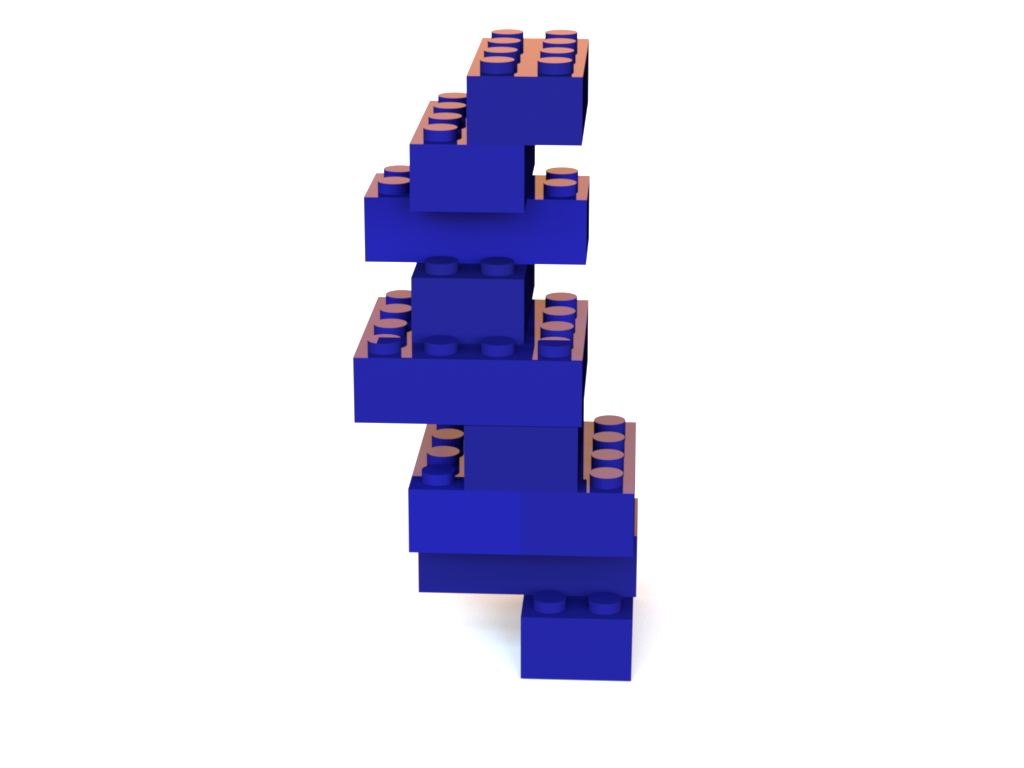}
    \includegraphics[width=0.165\columnwidth]{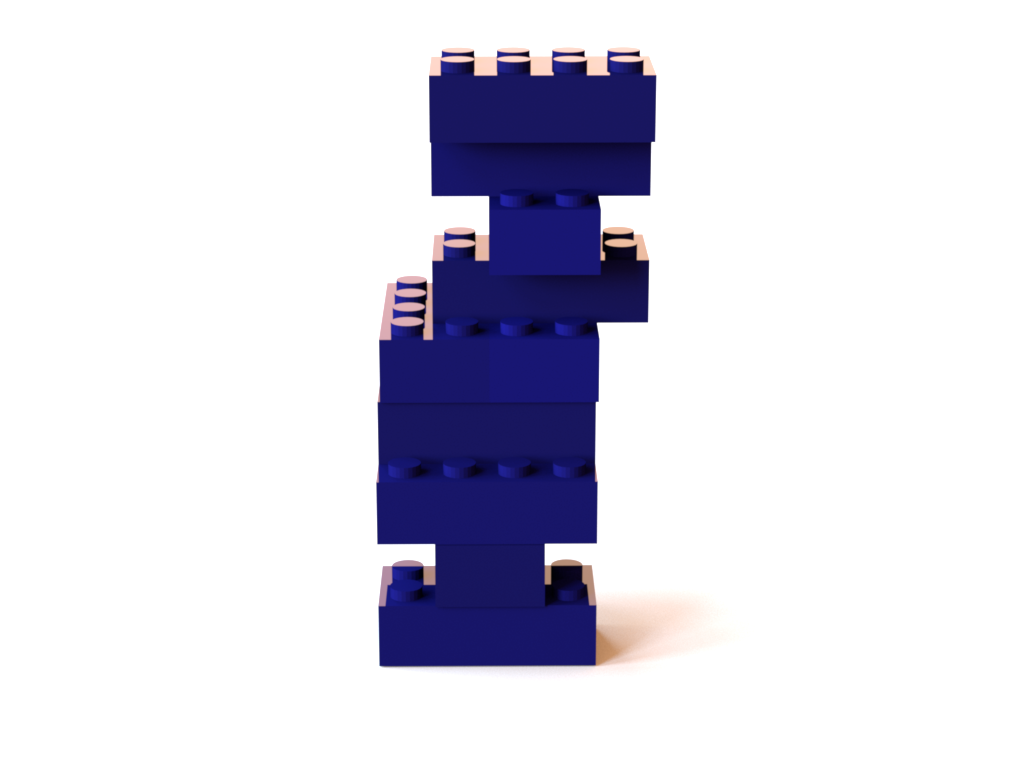}
    \includegraphics[width=0.165\columnwidth]{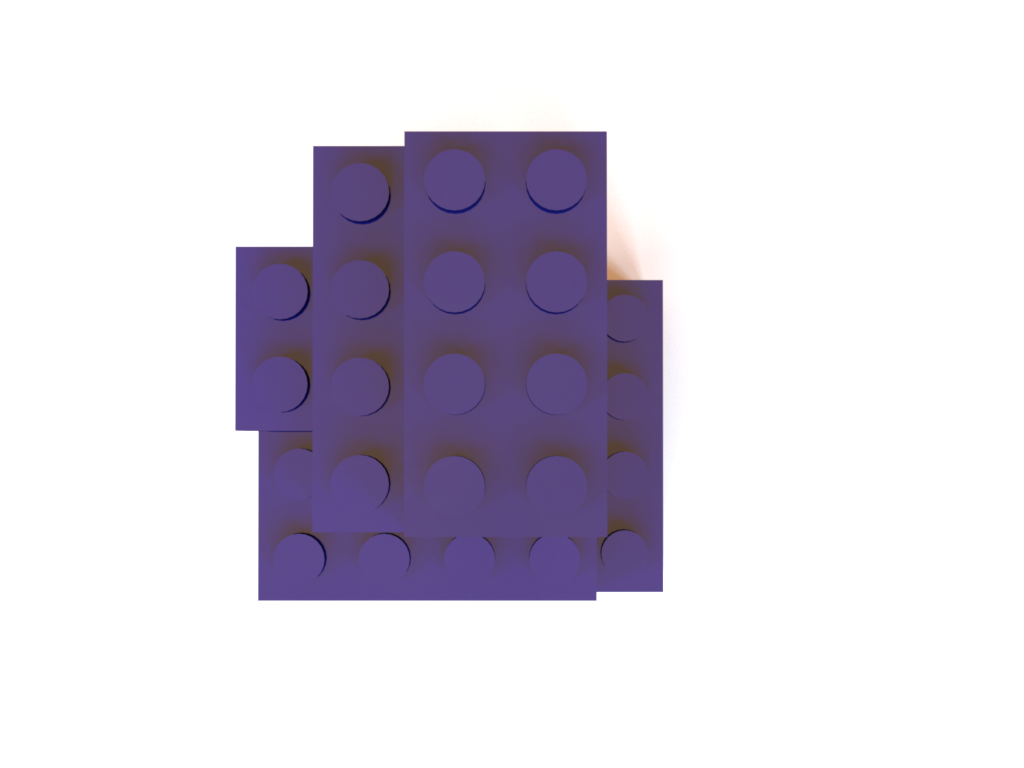}
    \label{fig:artificial_constructed_3}
}
\setcounter{subfigure}{0}
\subfigure[Target images]{
    \includegraphics[width=0.125\columnwidth]{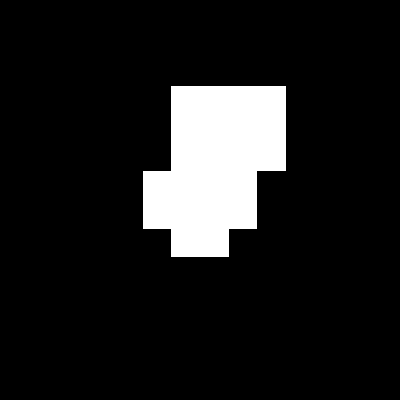}
    \includegraphics[width=0.125\columnwidth]{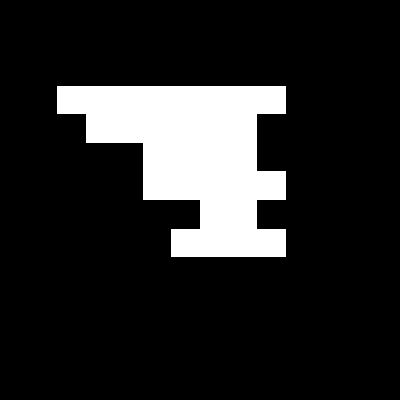}
    \includegraphics[width=0.125\columnwidth]{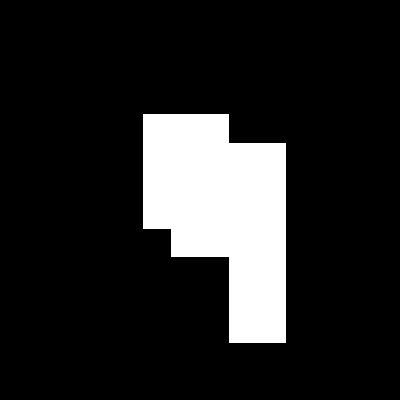}
    \label{fig:artificial_target_4}
}
\subfigure[Constructed objects from three viewpoints]{
    \includegraphics[width=0.165\columnwidth]{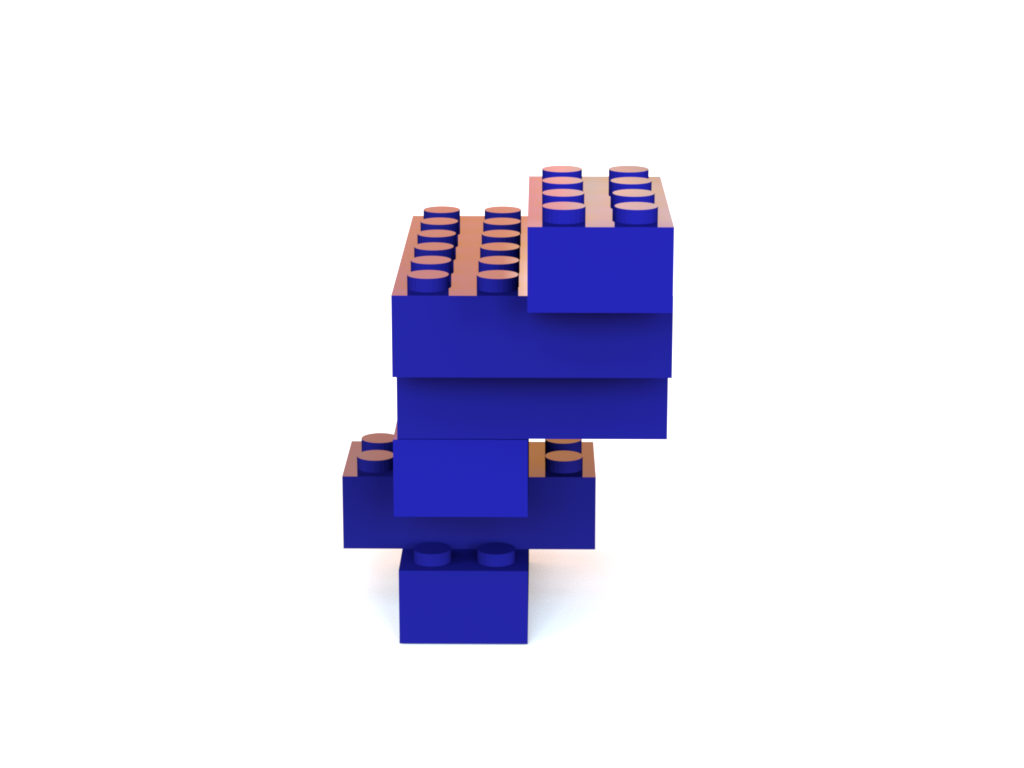}
    \includegraphics[width=0.165\columnwidth]{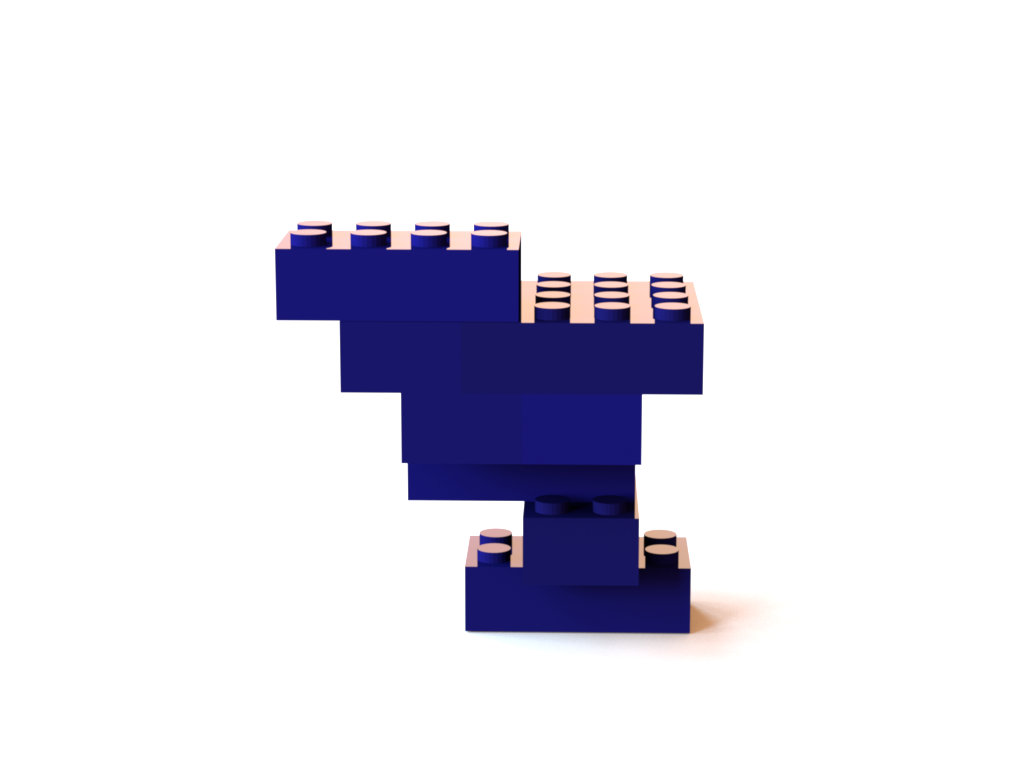}
    \includegraphics[width=0.165\columnwidth]{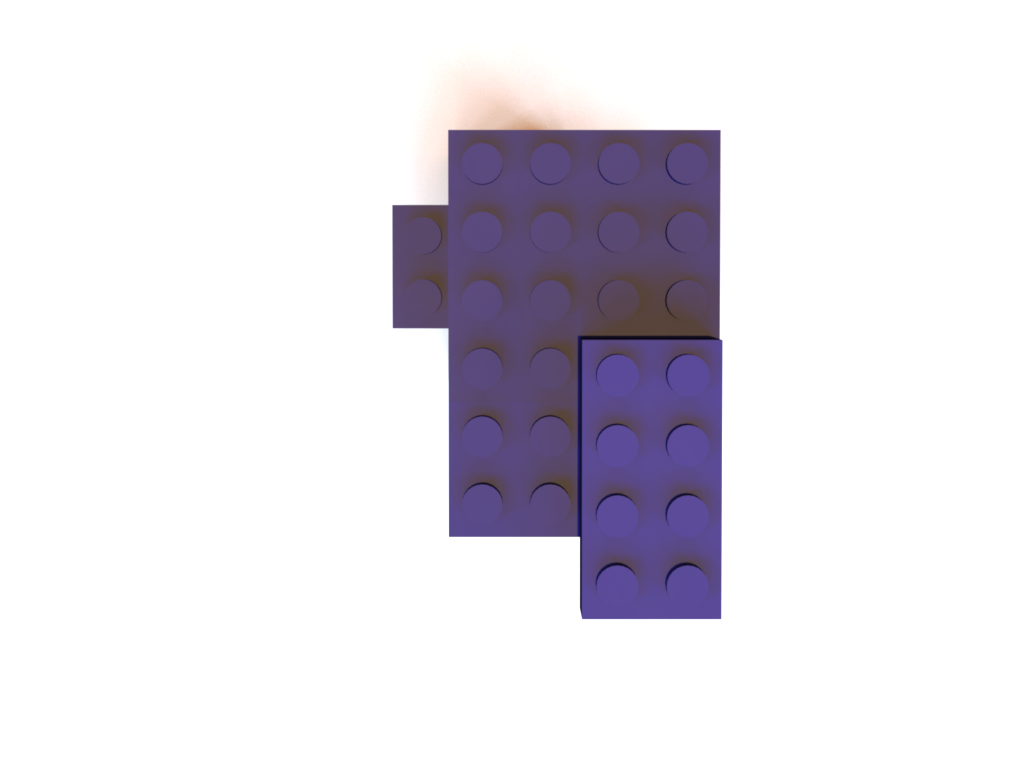}
    \label{fig:artificial_constructed_4}
}
\caption{Qualitative results on randomly-assembled object construction. Targets are obtained from the test dataset, and each row represents a pair of target information and constructed object.\label{fig:artificial}}
\end{figure}

\begin{figure}[t]
\centering
\subfigure[Airplane 1]{
    \includegraphics[width=0.10\columnwidth]{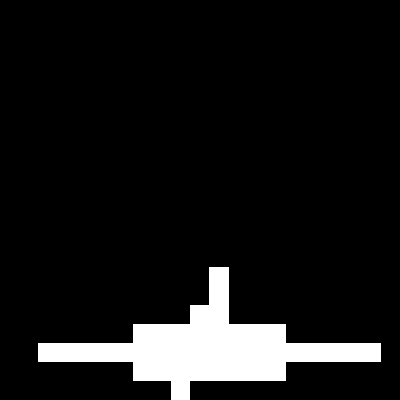}
    \includegraphics[width=0.10\columnwidth]{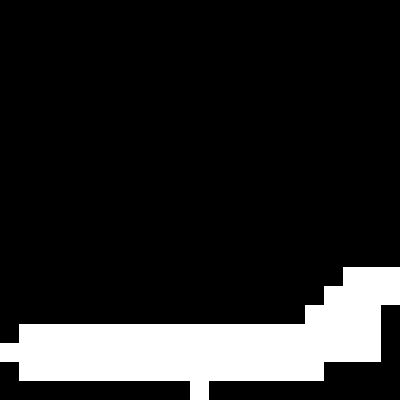}
    \includegraphics[width=0.10\columnwidth]{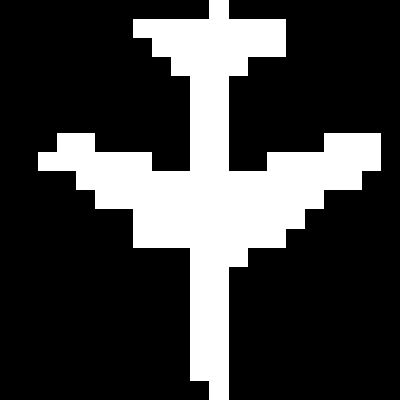}
    \includegraphics[width=0.15\columnwidth]{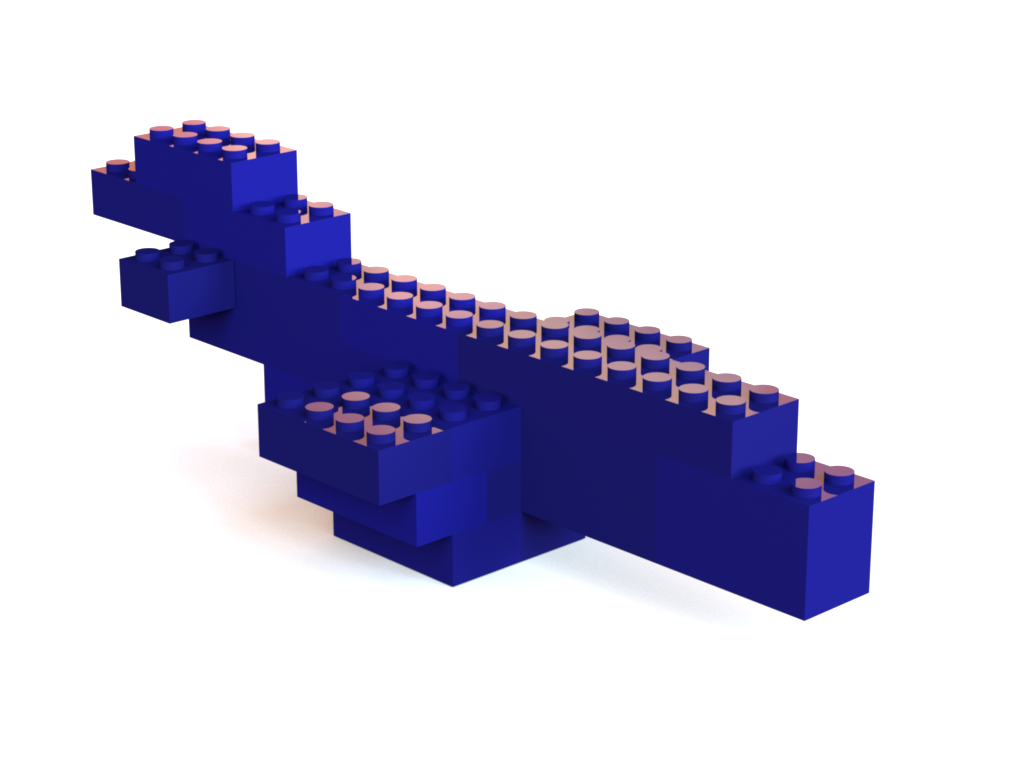}
    \label{fig:airplane_target_1}
}
\subfigure[Airplane 2]{
    \includegraphics[width=0.10\columnwidth]{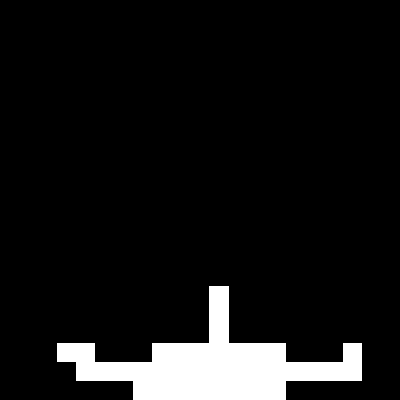}
    \includegraphics[width=0.10\columnwidth]{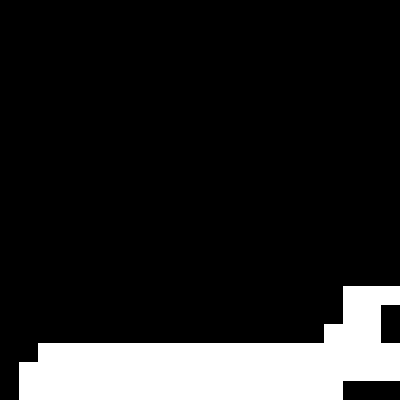}
    \includegraphics[width=0.10\columnwidth]{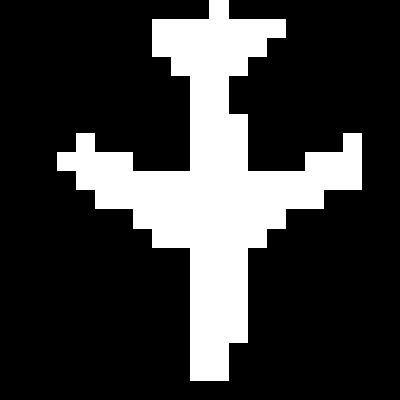}
    \includegraphics[width=0.15\columnwidth]{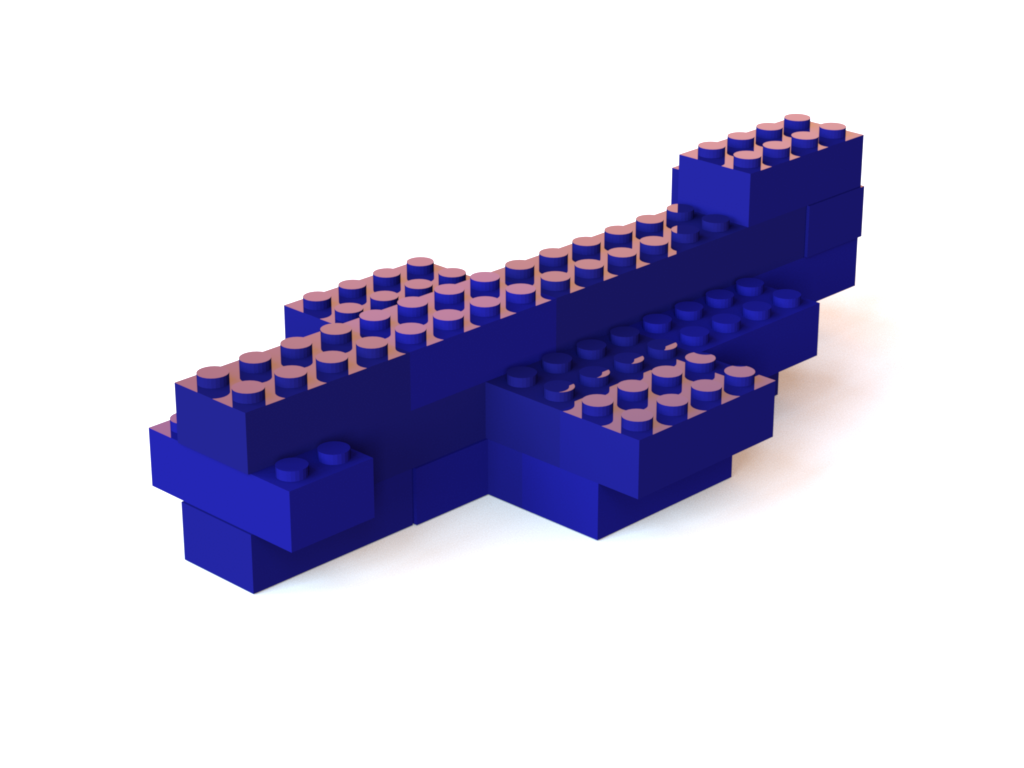}
    \label{fig:airplane_target_2}
}
\subfigure[Airplane 3]{
    \includegraphics[width=0.10\columnwidth]{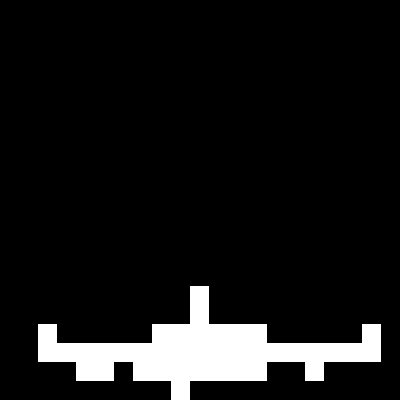}
    \includegraphics[width=0.10\columnwidth]{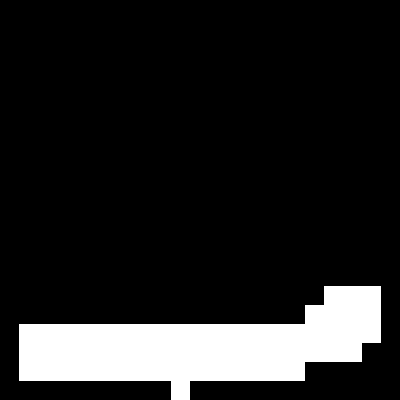}
    \includegraphics[width=0.10\columnwidth]{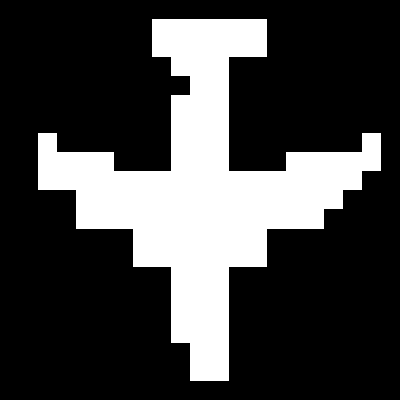}
    \includegraphics[width=0.15\columnwidth]{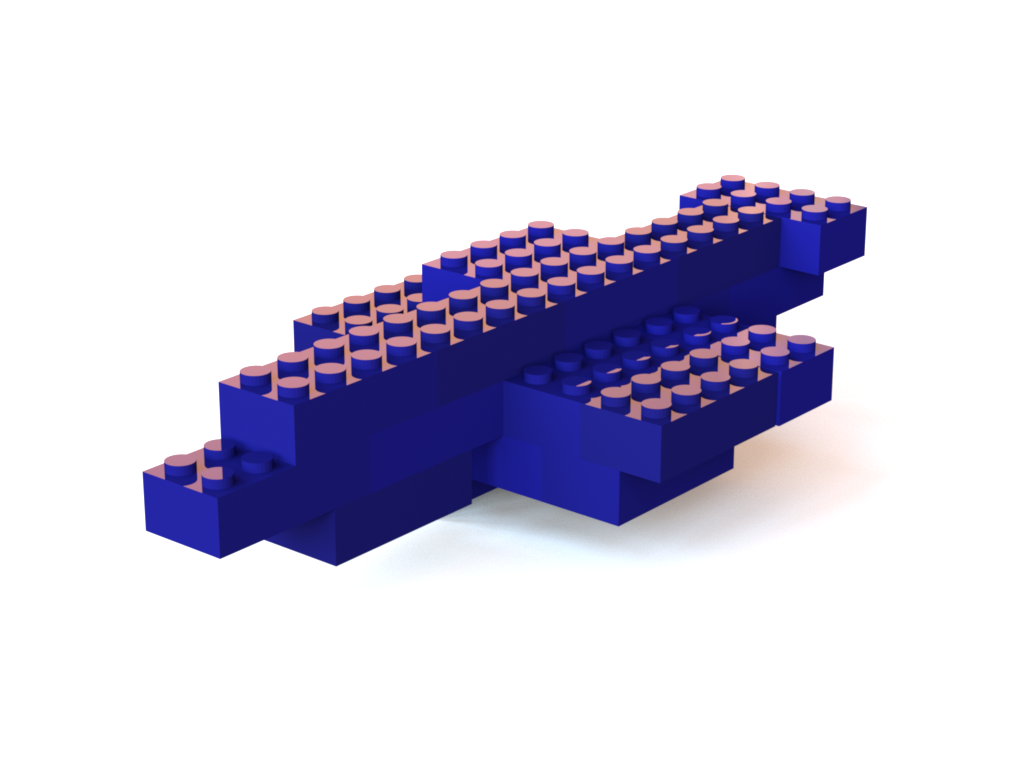}
    \label{fig:airplane_target_4}
}
\subfigure[Monitor 1]{
    \includegraphics[width=0.10\columnwidth]{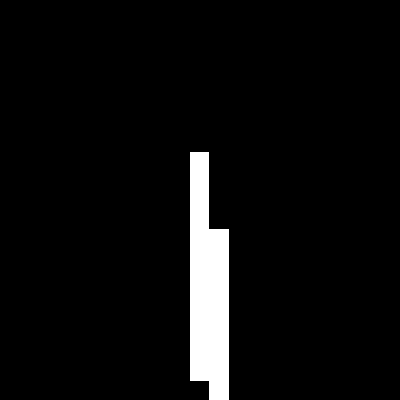}
    \includegraphics[width=0.10\columnwidth]{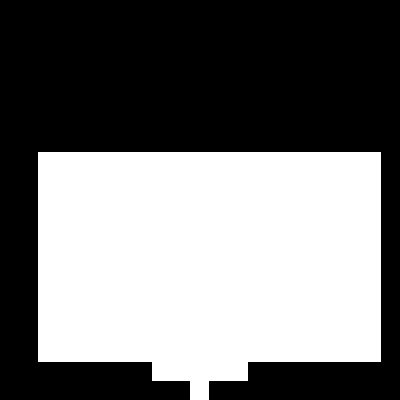}
    \includegraphics[width=0.10\columnwidth]{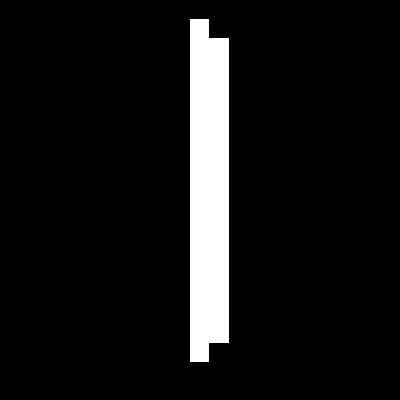}
    \includegraphics[width=0.15\columnwidth]{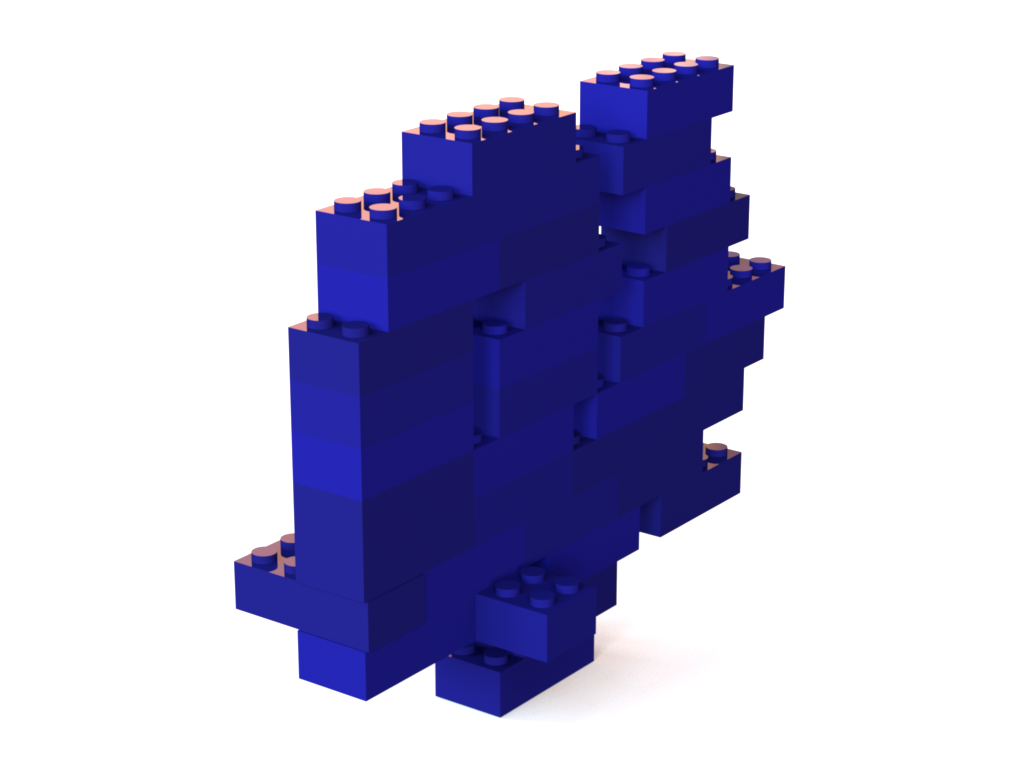}
    \label{fig:monitor_target_1}
}
\subfigure[Monitor 2]{
    \includegraphics[width=0.10\columnwidth]{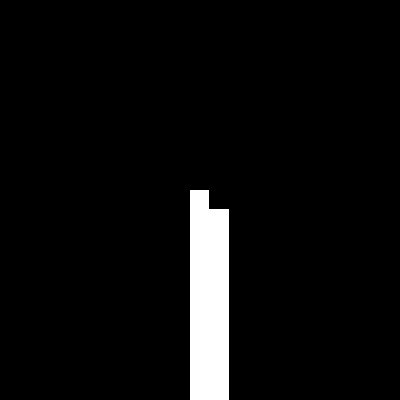}
    \includegraphics[width=0.10\columnwidth]{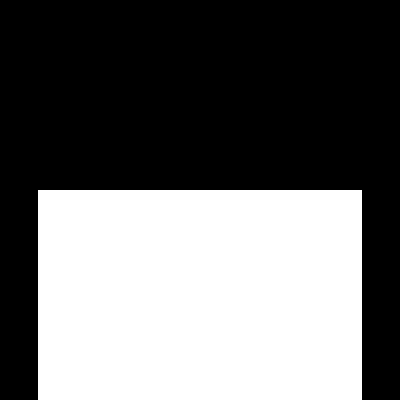}
    \includegraphics[width=0.10\columnwidth]{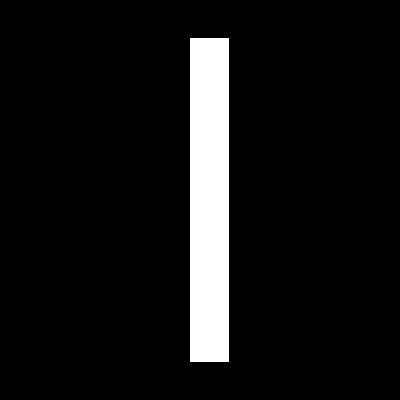}
    \includegraphics[width=0.15\columnwidth]{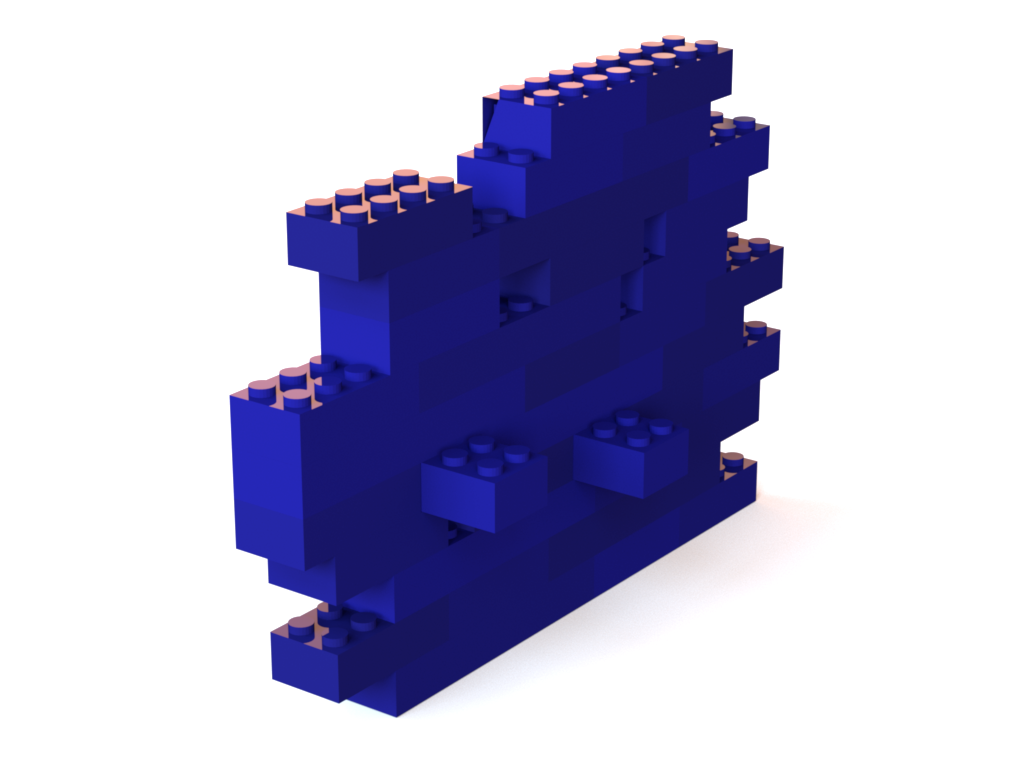}
    \label{fig:monitor_target_2}
}
\subfigure[Table]{
    \includegraphics[width=0.10\columnwidth]{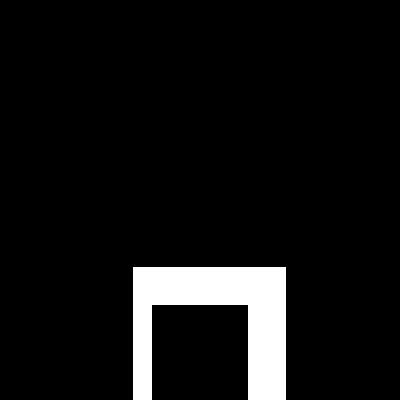}
    \includegraphics[width=0.10\columnwidth]{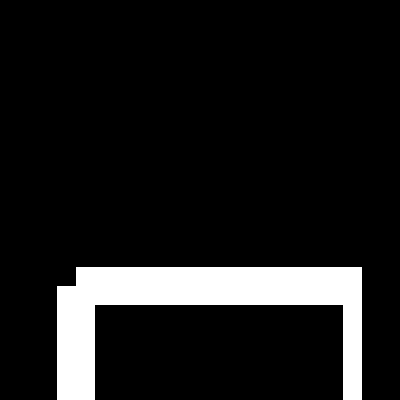}
    \includegraphics[width=0.10\columnwidth]{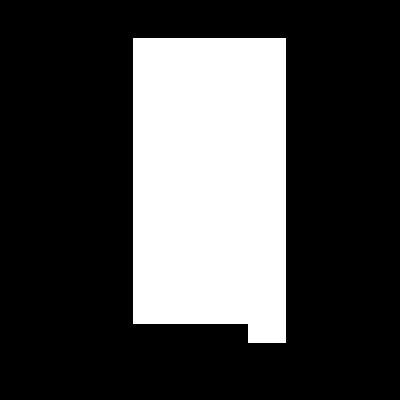}
    \includegraphics[width=0.15\columnwidth]{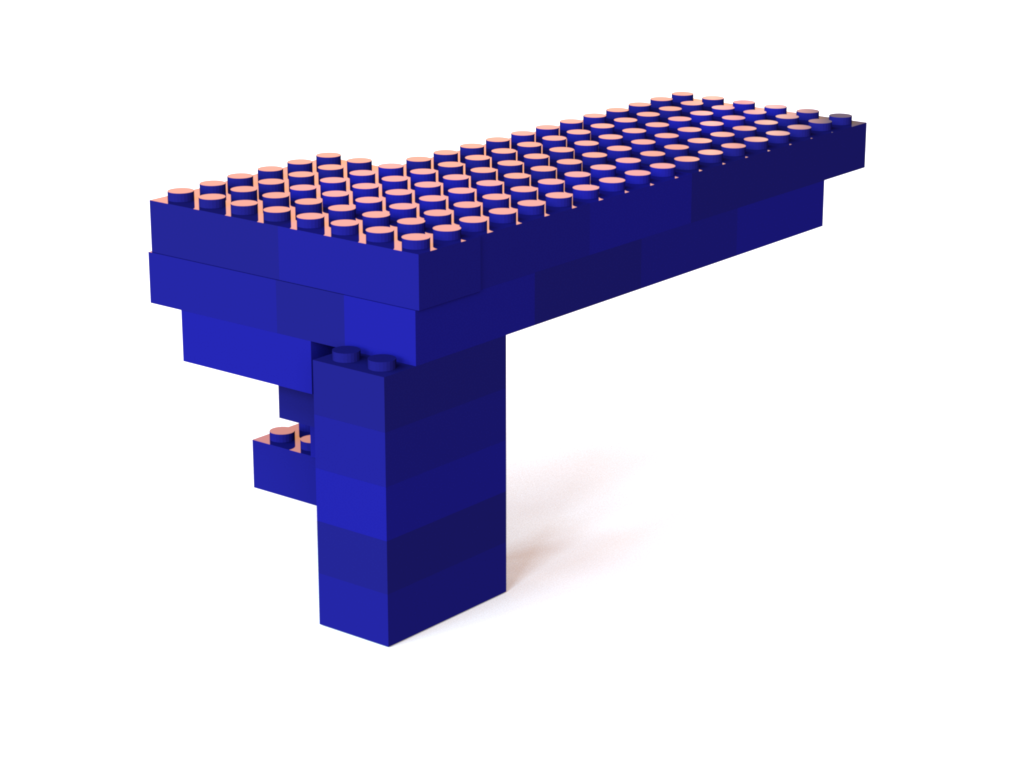}
    \label{fig:table_target_1}
}
\caption{Results on ModelNet construction. Targets are obtained from the test dataset. The first three panels and the last panel of each figure show the target images and the constructed example.}
\label{fig:modelnet}
\end{figure}

\paragraph{Randomly-Assembled Object Construction.}
Contrary to the experiments of MNIST construction, 
this task focuses on building objects that require more than one image 
to fully understand the structure.
Accordingly, the agent must construct an object with three $14 \times 14$ images 
from different viewpoints, which are initially given as the target information.
Objects in this experiment are artificially generated by connecting bricks at random.
The total number of bricks is also chosen randomly between 10 to 15.
For available offset types, we only utilize connection types 
that occupy four or more studs and only allow a new brick 
to be placed on top of the pivot brick so that the resulting target 
becomes more distinguishable; 
see the supplementary material for the details.
In this scheme, the total number of offsets is 16.
Finally, we sample images from 800 target objects for a training dataset 
while 200 target objects are used for a test dataset.

As shown in~\figref{fig:3dobject_easy}, 
our model achieves a return comparable to the MLP-based model 
while a slightly lower return compared to BO.
We conjecture that this is due to the relatively small number of used bricks compared to other test suites.
Nevertheless, our agent is still capable of 
associating the target object in 3D space from multiple images 
as illustrated in~\figref{fig:artificial}.
Our model learns to assemble bricks in a way that the resulting object 
successfully matches the initially given images, 
whereas the model trained with the supervised learning method does not generalize 
to unseen images of the test dataset.
This clearly demonstrates the effectiveness of applying RL 
compared to learning with sequence-level supervision.

\paragraph{ModelNet Construction.}
Similar to randomly-assembled object construction, 
the agent is given 3 images of a realistic target 
from the ModelNet dataset~\cite{WuZ2015cvpr}.
To adjust the difficulty of this task, 
we find an object that is able to limit the maximum budget of bricks to less than 60.
As a result, we choose airplane, monitor, and table categories from the ModelNet dataset.
Moreover, we use offset types that connect with four or more studs 
and allow a new brick to be placed above and below the pivot brick.
This task is the most challenging 
due to the excessive search space 
compared to MNIST construction and randomly-assembled object construction, 
and assesses the agent's ability to generate a real-world target object.

We provide training and testing curves for the airplane class in \figref{fig:3dobject_hard};
see the supplementary material for more results for monitor and table categories.
Despite the difficulty raised from the large search space and long sequence,
the result demonstrates that $\ours$ is capable of learning the construction process of real-world objects.
BO with a limited budget achieves lower return compared to $\ours$ since the search space is too big to explore with limited computation.
By comparing the constructed object to images of the desired target in~\figref{fig:modelnet}, 
it can be observed that $\ours$ generally captures overall shape of the target. Though, our model tends to struggle to catch fine-grained details such as wings of the airplane or legs of the table.
However, for example, a table with only three legs (i.e., one leg missing) or two legs in a diagonal direction would perfectly match with the same three input views.
It implies that if we provide more complete target information than three different views of target object, our agent can construct the target object more precisely; 
see the supplementary material for more detailed discussion on this limitation.

%% file: sections/05_related_work.tex
\section{Related Work\label{sec:related_work}}

In this section, we briefly cover related work on the task solved in this work.

\paragraph{3D Object Generation.}
Following the studies on 2D object generation, e.g., the work by~\citet{DosovitskiyA2016ieeetpami}, 
3D object generation is often achieved in holistic 
manner~\citep{WuJ2016neurips,AchlioptasP2018icml,HenzlerP2019iccv,YangG2019iccv}.
They generate a 3D object in a single feed-forward operation 
which limits exploitation of intermediate structures.
Compared to these holistic methods,
\citet{KimJ2020ml4eng} propose an approach to tackle a combinatorial assembly problem 
by using Bayesian optimization~\citep{BrochuE2010arxiv}, not a learning-based method.
Unlike~\citep{KimJ2020ml4eng}, \citet{ThompsonR2020neuripsw} apply a graph-structured 
generative model in the combinatorial 3D object generation task, 
by training to match a ground-truth sequence of LEGO bricks.

\paragraph{Graph-based Reinforcement Learning.}
A common learning-based technique for creating a graph is 
to use one of various models such as recurrent neural networks~\citep{YouJ2018icml,LiY2018arxiv}, 
adversarial networks~\citep{DeCaoN2018arxiv}, 
variational autoencoders~\citep{JinW2018icml,SamantaB2020jmlr,SimmG2020icml_a}, 
and Transformers~\citep{NashC2020icml}.
Unlike these directions, 
\citet{SimmG2020icml_b} solve this idea of generating molecules 
with RL such that generated molecules are placed 
in the Cartesian coordinate.
The key difference to our work is that we sequentially generate 3D shapes which have a much larger search space.
Furthermore, \citet{BapstV2019icml} show that an RL agent
can learn physical construction in 2D space, 
and utilize rich visual information as well as a graph-structured representation, 
in order to define a state and a search space.

\paragraph{Image-Conditioned Reinforcement Learning.}
\citet{GaninY2018icml} propose an approach to synthesizing a program for 2D images 
when either an unconditional or conditional scenario is assumed.
Their method generates an image by sequentially conducting an action in the MuJoCo environment.
\citet{NairAV2018neurips} propose a goal-conditioned RL approach the goal of which is provided by visual information.
\citet{HuangZ2019iccv} suggest a method to paint a palette with stokes where a target image is conditionally given, by utilizing an RL algorithm.

\paragraph{Brick Assembly Optimization.}
The brick assembly problem satisfying pre-defined constraints is a longstanding topic in computer graphics.
\citet{LeeS2015gecco} tackle LEGO brick layout optimization by a genetic algorithm.
Similarly, \citet{LuoSJ2015acmtgraphics} solve building sculptures safely with LEGO brick by stability aware refinement.
\citet{ZhangM2017cada} propose the method for generating component-based building instructions that is safe based on segmentation models.
\citet{KozakiT2016cad} tackle a similar problem of brick assembly from images with the octree voxel-based model.
This line of research tends to focus on directly utilizing the voxel representation of target object, instead of extracting the representation of target object from incomplete partial information.

%% file: sections/06_conclusion.tex
\section{Conclusion\label{sec:conclusion}}

We have proposed a novel problem formulation, combinatorial construction, 
which asks an agent to construct an object sequentially. 
We adopt RL by defining a state as graph-structured representation to express assembled bricks and their connections,
where incomplete target information is given.
In addition, we develop our algorithm with a successive action space 
that does not depend on the number of bricks already constructed 
and a reward function that measures overlap 
between a target object and the current state.
Through extensive experiments, we demonstrate that our method can construct objects 
in various construction scenarios, and provide the analysis 
of our action validity prediction network.\looseness=-1

%% file: sections/07_supplementary.tex
{\Large \textbf{Supplementary Material}}

\renewcommand{\theequation}{s.\arabic{equation}}
\renewcommand{\thetable}{s.\arabic{table}}
\renewcommand{\thefigure}{s.\arabic{figure}}
\renewcommand{\thealgorithm}{s.\arabic{algorithm}}
\renewcommand{\thesection}{S.\arabic{section}}
\renewcommand{\thesubsection}{S.\arabic{section}.\arabic{subsection}}

\setcounter{equation}{0}
\setcounter{table}{0}
\setcounter{figure}{0}
\setcounter{algorithm}{0}
\setcounter{section}{0}
\setcounter{subsection}{0}

\vspace{10pt}

In this material, we first describe the importance of
action validity prediction networks.
Then, we introduce the details of the benchmarks, provide the model architecture, 
and present the additional experimental results, which are missing in the main article.
Finally, we discuss limitations and societal impacts of our work in the last section.

\section{Action Validity Prediction Network}

\begin{figure}[ht]
    \centering
    \includegraphics[width=0.4\textwidth]{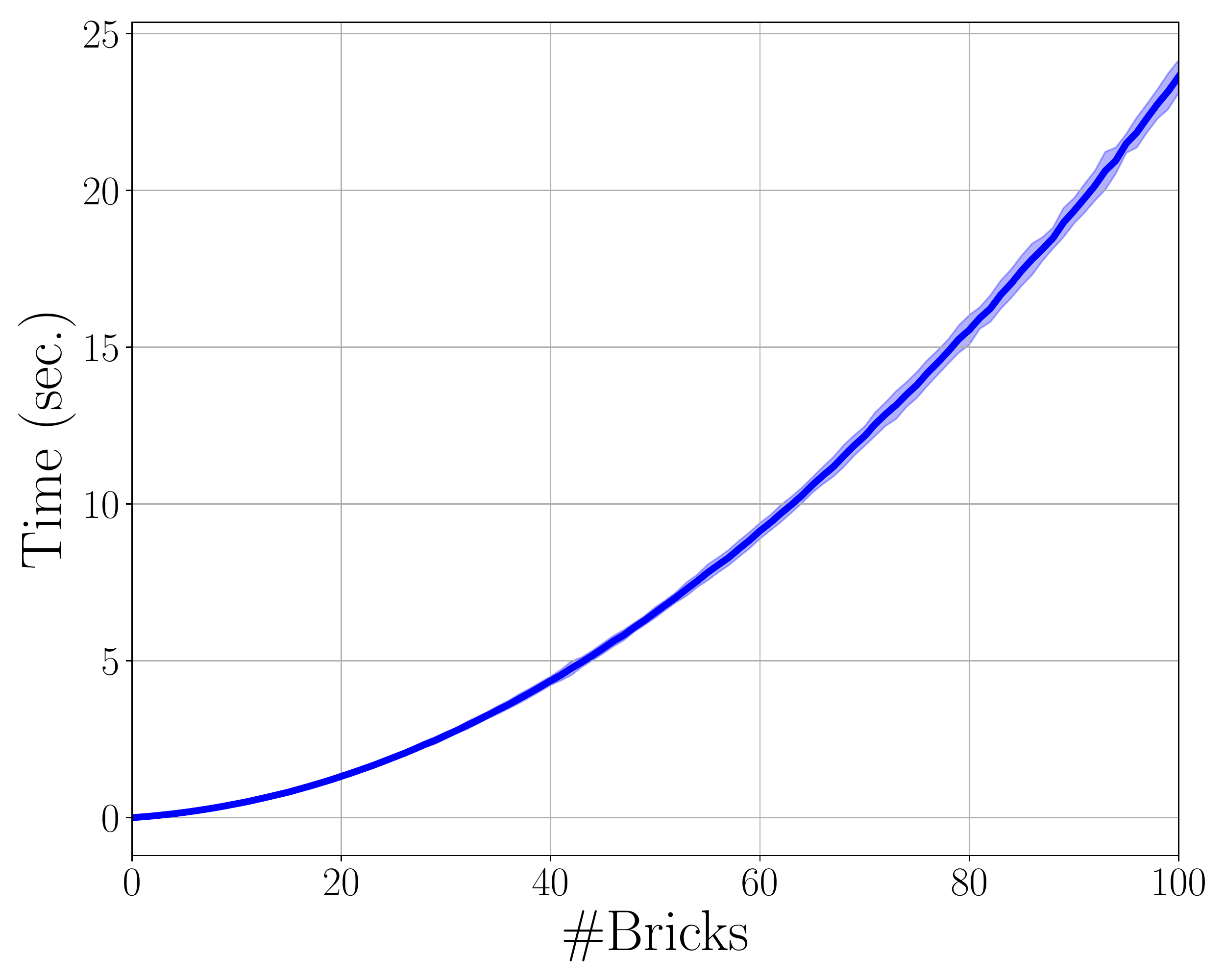}
    \caption{Results of wall-clock time for computing the ground-truth action validity. We repeat 10 times and plot $\pm 1.96$ standard deviation.\label{fig:time_valid}}
\end{figure}

\input{tables/tab_action_validation}
\input{tables/tab_avn}

Compared to the construction cases with ground-truth action validity,
the cases with our action validity prediction network are
beneficial in terms of computational costs.
We present the results of wall-clock time 
for computing the ground-truth action validity in~\figref{fig:time_valid}.
It shows that computing the action validity for a combination of 100 bricks 
needs more than 20 seconds.
Moreover, we summarize the comparisons between possible action validation approaches 
as shown in~\tabref{tab:action_validation}.

As described in the main article, 
our action validity prediction network can be pretrained using the episodes 
obtained from the randomly-assembled object construction task 
and only requires a single forward pass 
to compute the action validity in inference time.
In addition to these, we show the results on predicting invalid actions 
by an action validity prediction network in~\tabref{tab:avn}.
The results shown in~\figref{fig:roc_pr} and this table 
demonstrate that our pretrained network is effective 
in predicting action validity.
For the jointly-trained validity prediction network, 
we assume that the oracle agent decides the next actions by obtaining them from the training dataset,
which implies that all the 10,000 episodes in the training dataset are used to train the action validity prediction network with a single epoch.

\section{Details of Benchmarks}

At the beginning of each episode, the agent starts with a single brick placed 
at the origin with the direction 0 regardless of the type of experiment it is being tested on.
Specifically, the agent is given a graph with a single node feature 
of $[0, 0, 0, 0]$ 
and the edge feature matrix of zero values along with target information.
All the hyperparameters are described in~Tables~\ref{tab:mnist_hyperparam} 
and \ref{tab:other_hyperparam}.
Below, we present the additional details distinctive for each benchmark.

\subsection{MNIST Construction}

\begin{figure}[ht]
    \centering
    \includegraphics[width=0.16\textwidth]{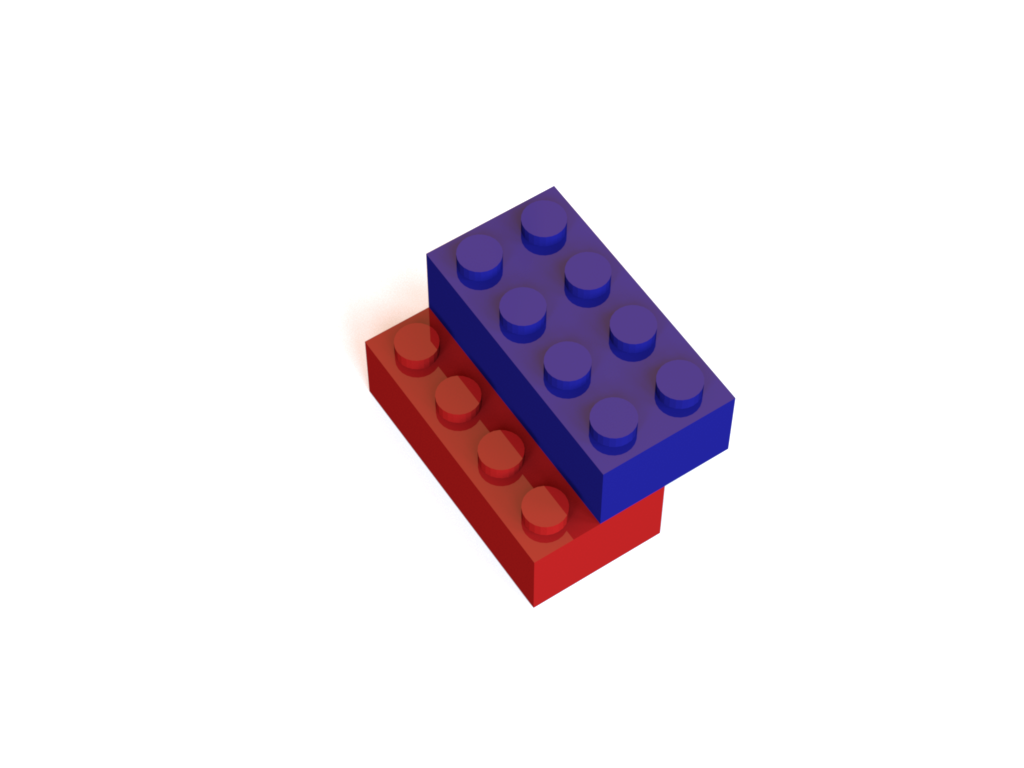}
    \includegraphics[width=0.16\textwidth]{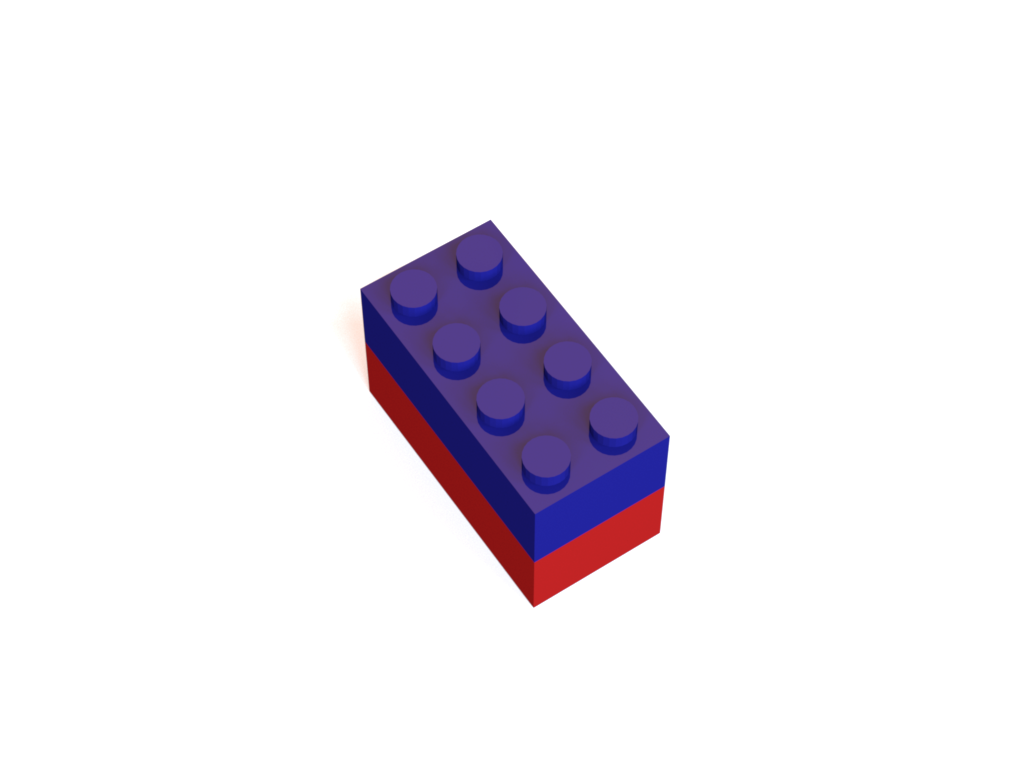}
    \includegraphics[width=0.16\textwidth]{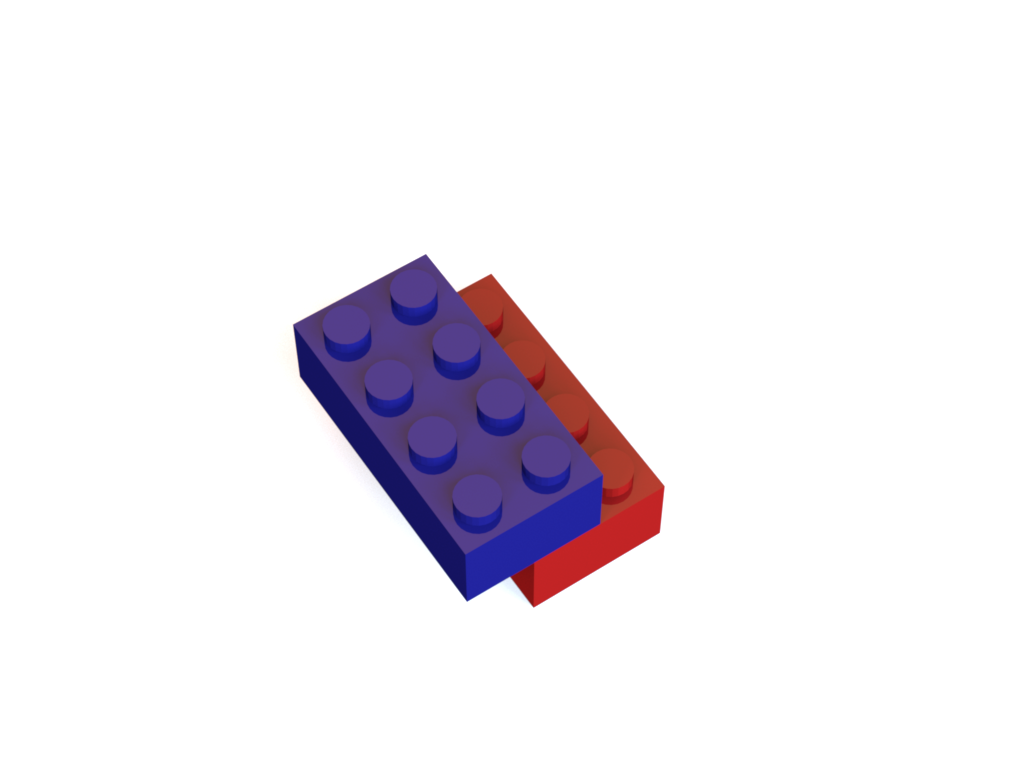}
    \includegraphics[width=0.16\textwidth]{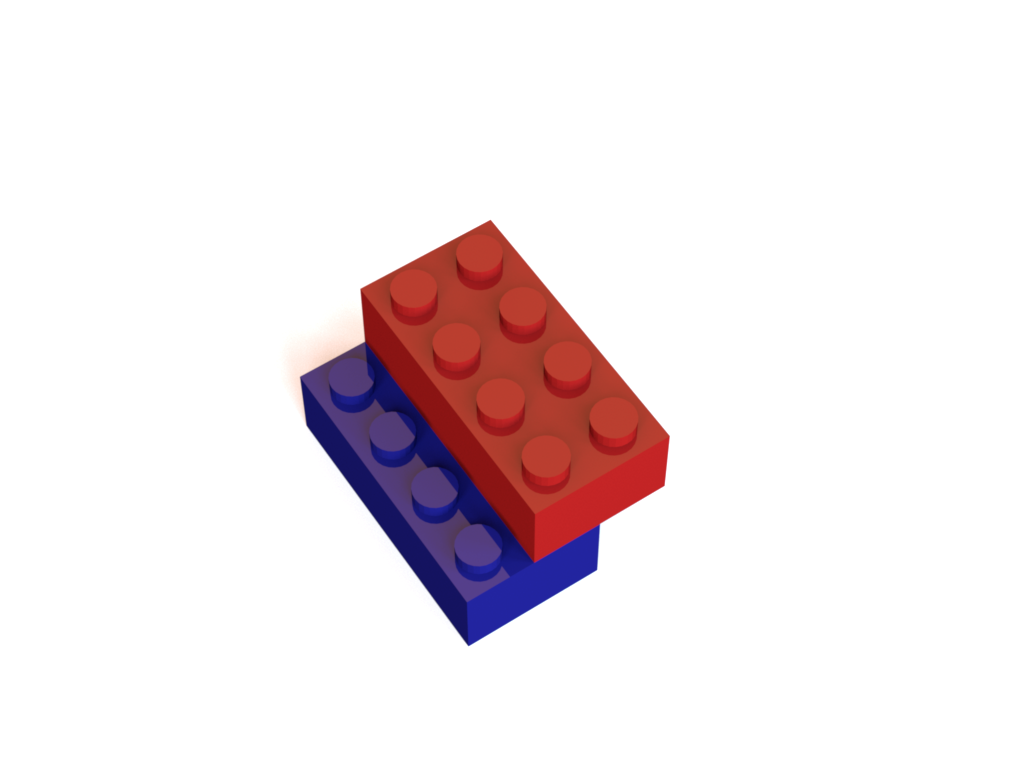}
    \includegraphics[width=0.16\textwidth]{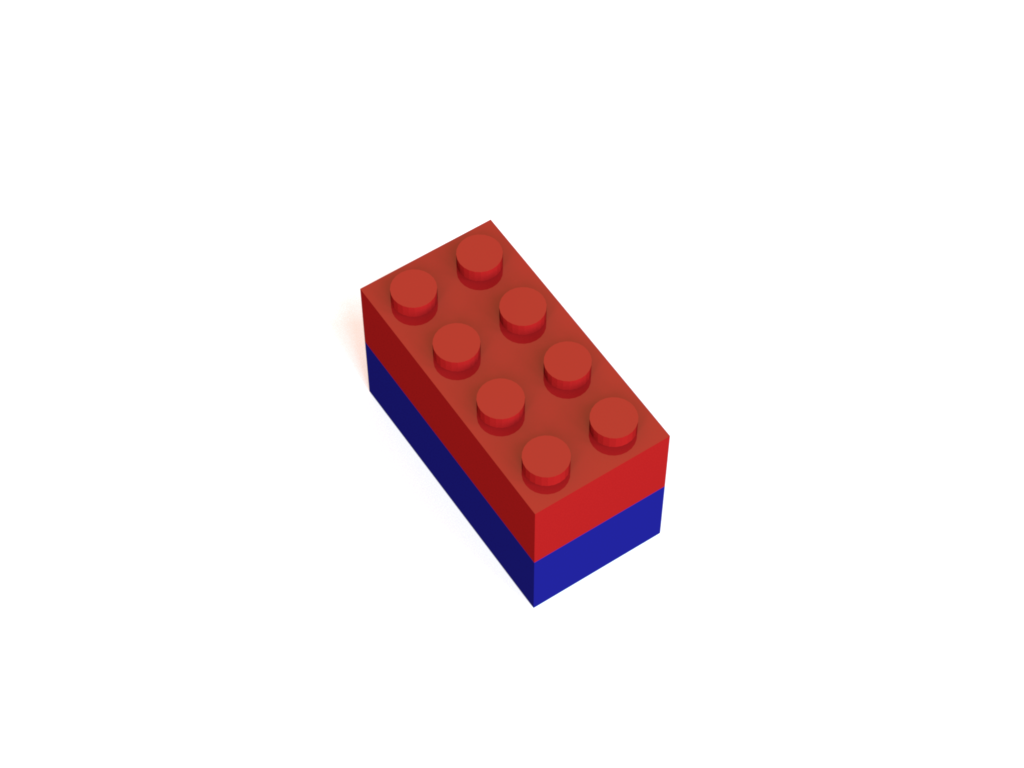}
    \includegraphics[width=0.16\textwidth]{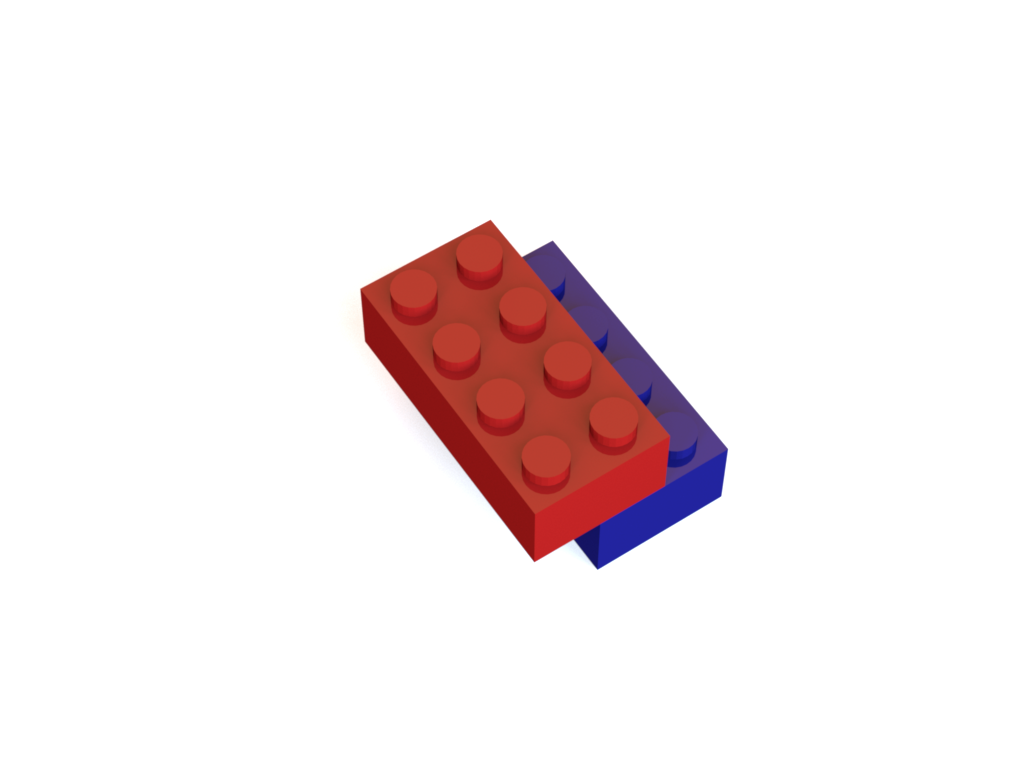}
    \caption{Visualization of available offsets for MNIST construction.\label{fig:sup_mnist_easy_offset}}
\end{figure}

Available offsets are visualized in~\figref{fig:sup_mnist_easy_offset}.
Since new brick (colored in dark blue) can be placed below the pivot brick (colored in red), 
the total number of offsets is 6.

Brick budget for each instance is set to 110\% of the total number of the pixels 
that have value 1 in a target MNIST image.

\subsection{Randomly-Assembled Object Construction}

\begin{figure}[ht]
\centering
\subfigure{
    \includegraphics[width=0.14\textwidth]{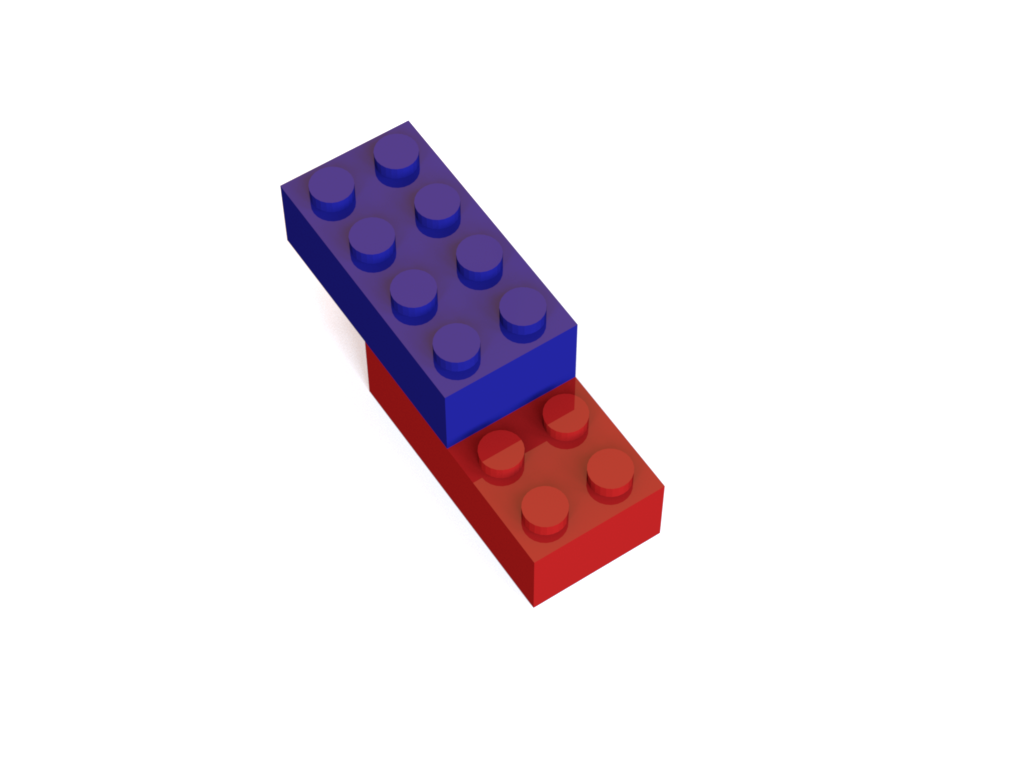}
}
\subfigure{
    \includegraphics[width=0.14\textwidth]{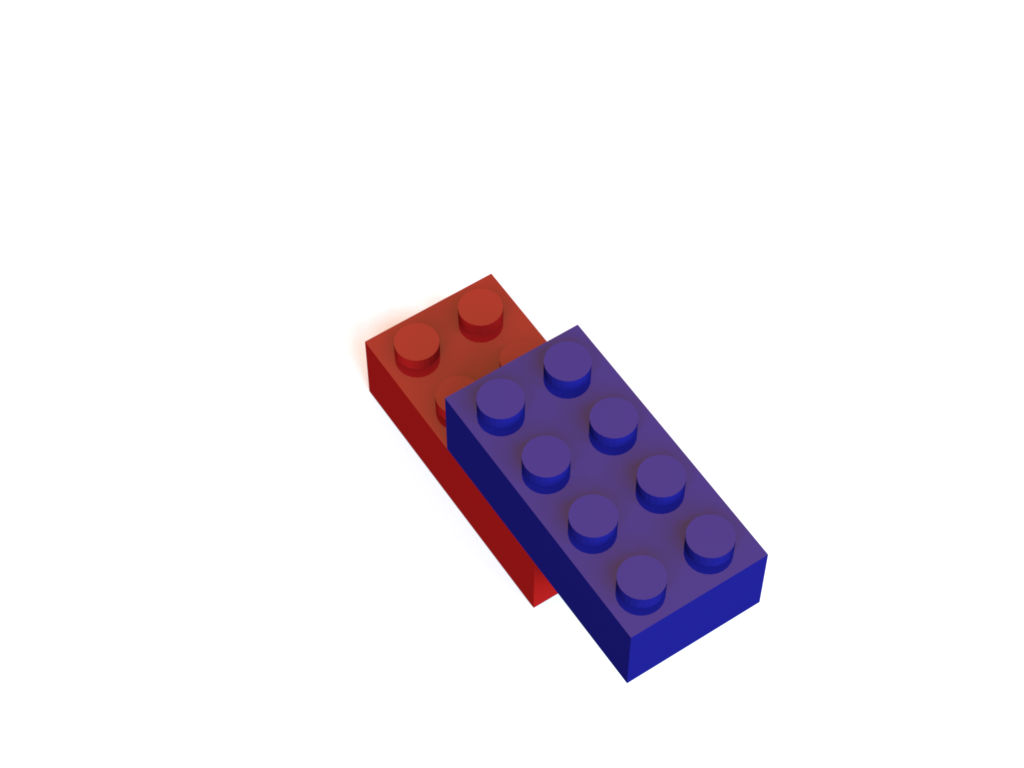}
}
\subfigure{
    \includegraphics[width=0.14\textwidth]{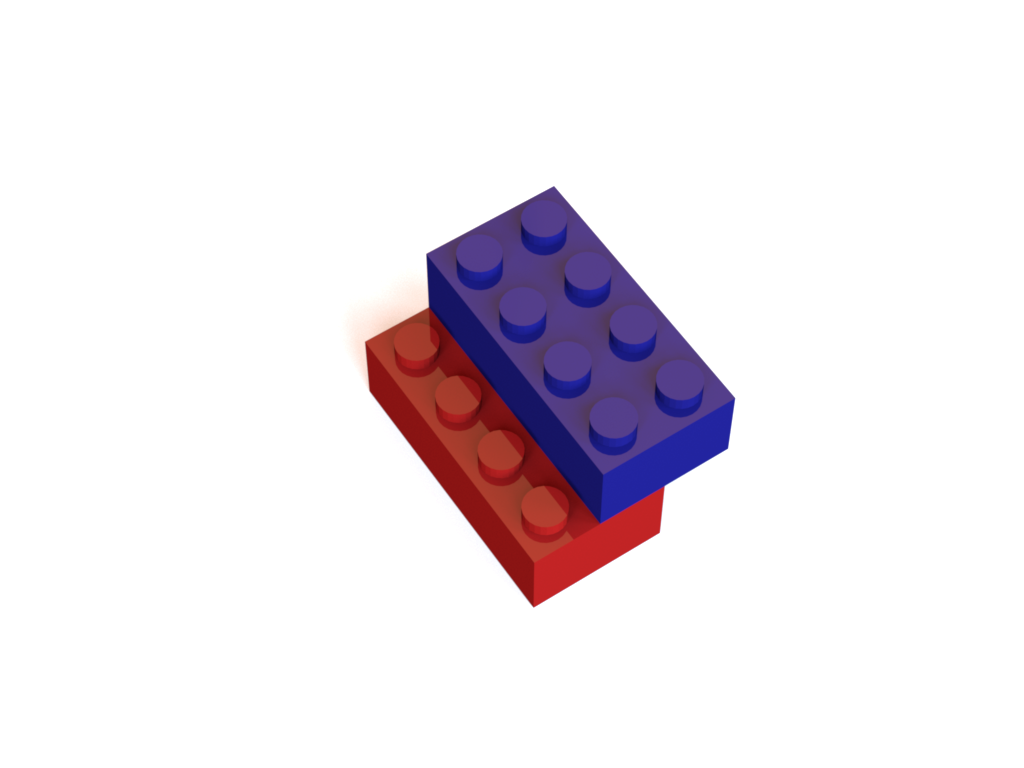}
}
\subfigure{
    \includegraphics[width=0.14\textwidth]{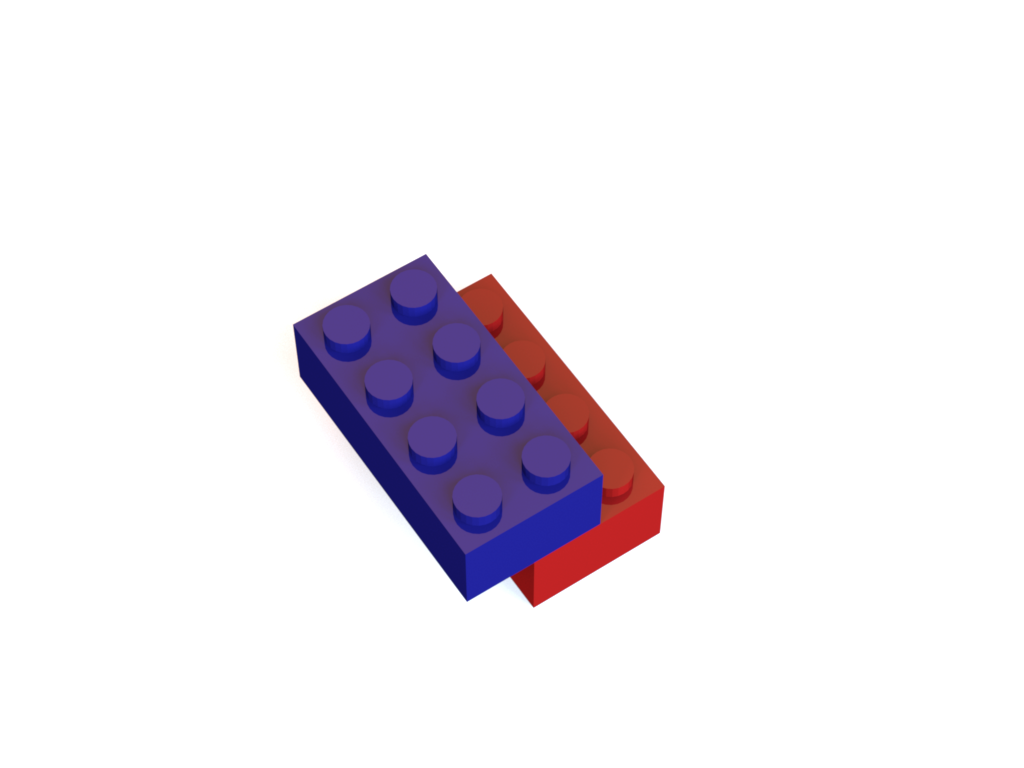}
}
\subfigure{
    \includegraphics[width=0.14\textwidth]{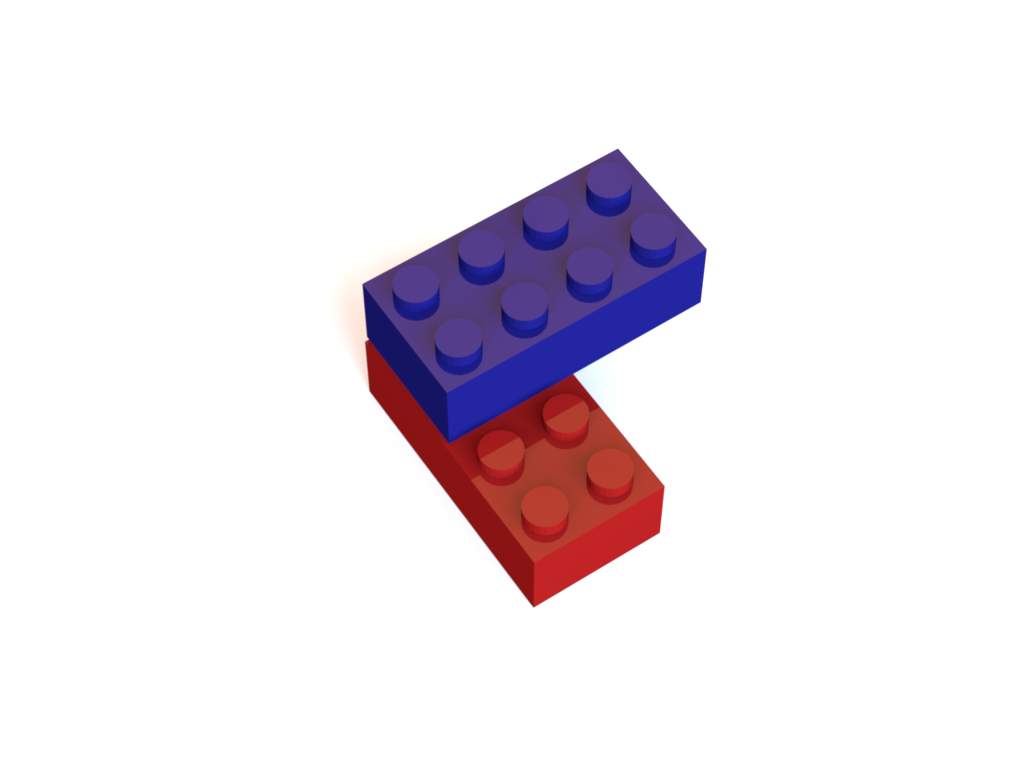}
}
\subfigure{
    \includegraphics[width=0.14\textwidth]{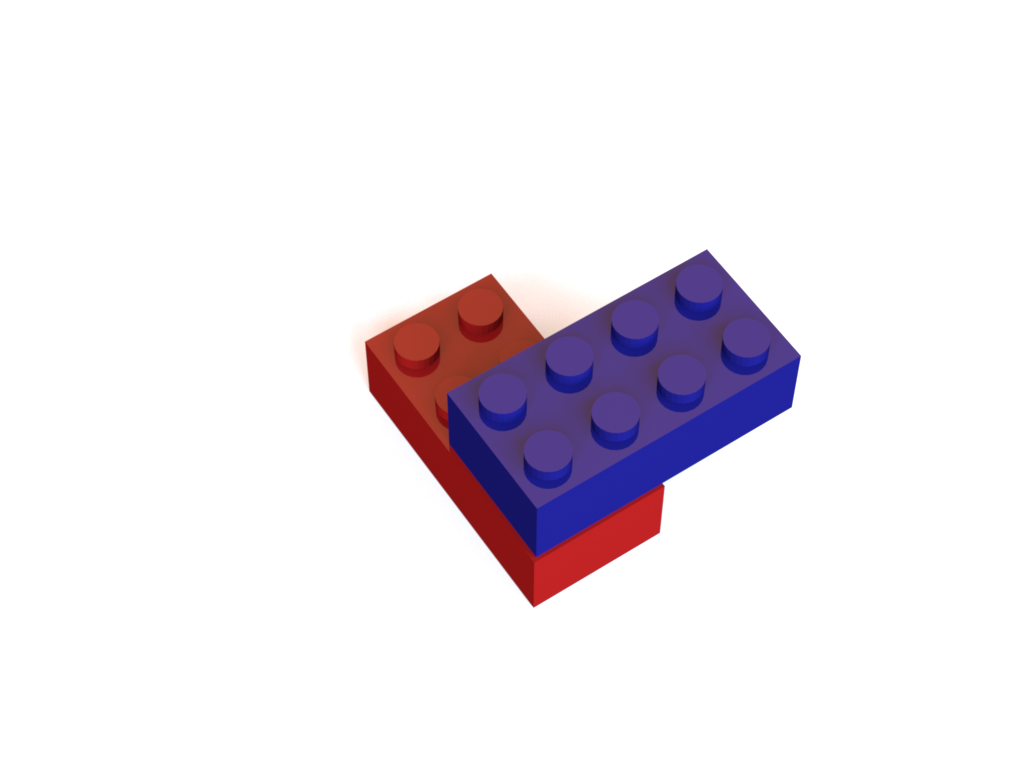}
}
\subfigure{
    \includegraphics[width=0.14\textwidth]{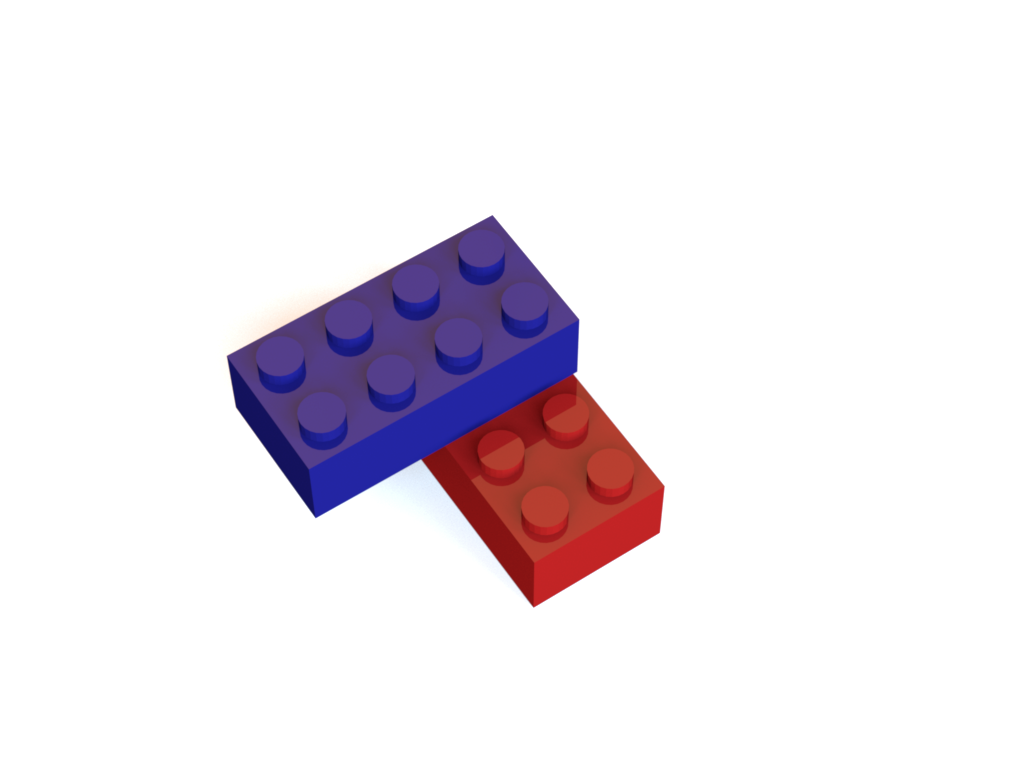}
}
\subfigure{
    \includegraphics[width=0.14\textwidth]{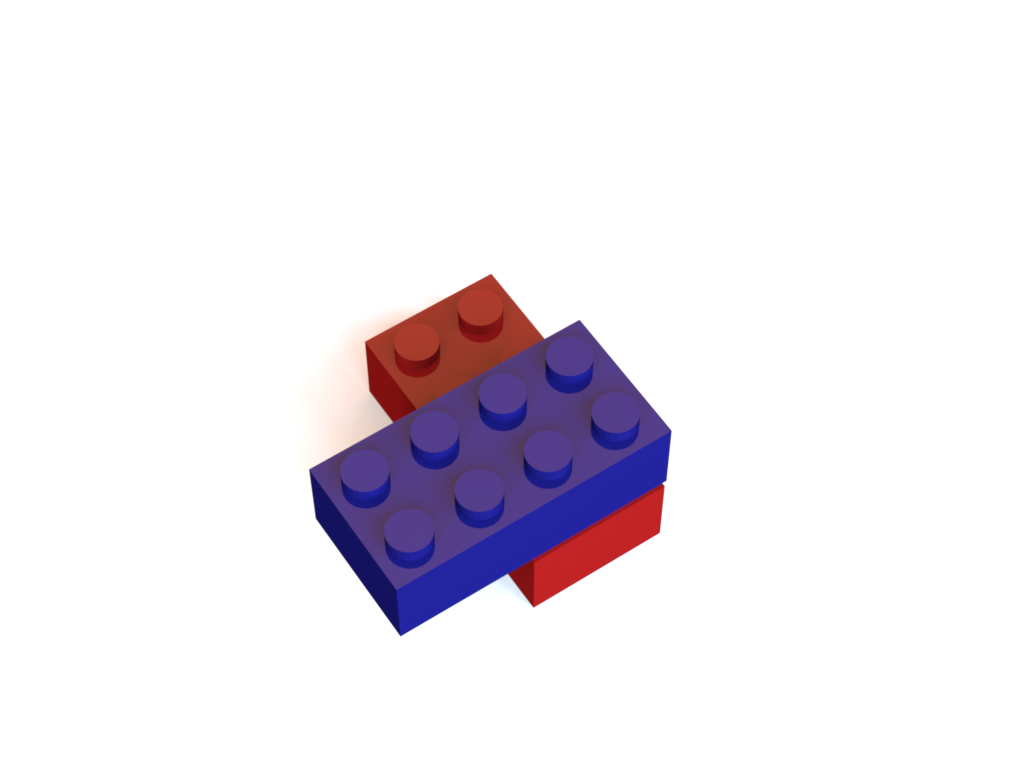}
}
\subfigure{
    \includegraphics[width=0.14\textwidth]{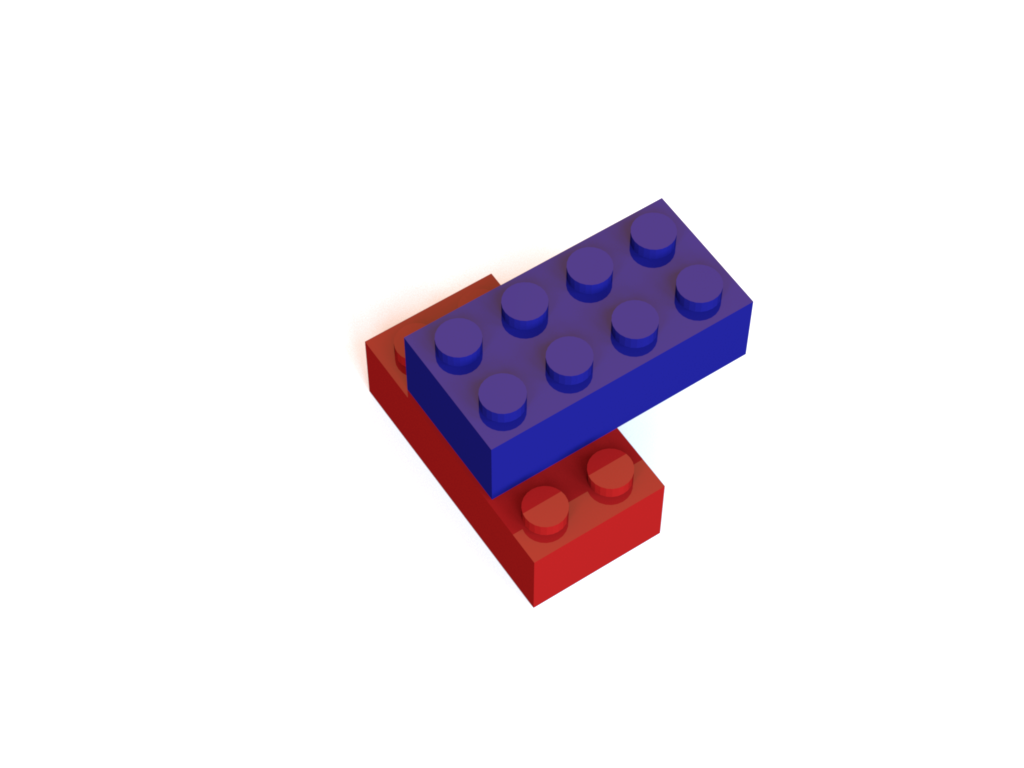}
}
\subfigure{
    \includegraphics[width=0.14\textwidth]{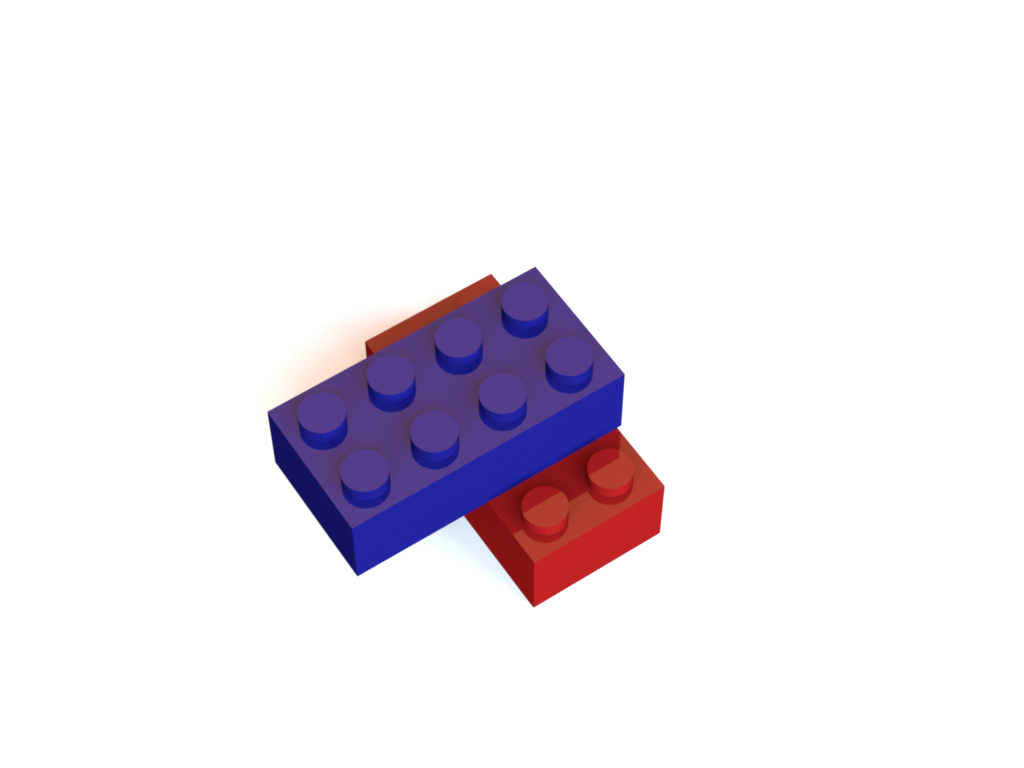}
}
\subfigure{
    \includegraphics[width=0.14\textwidth]{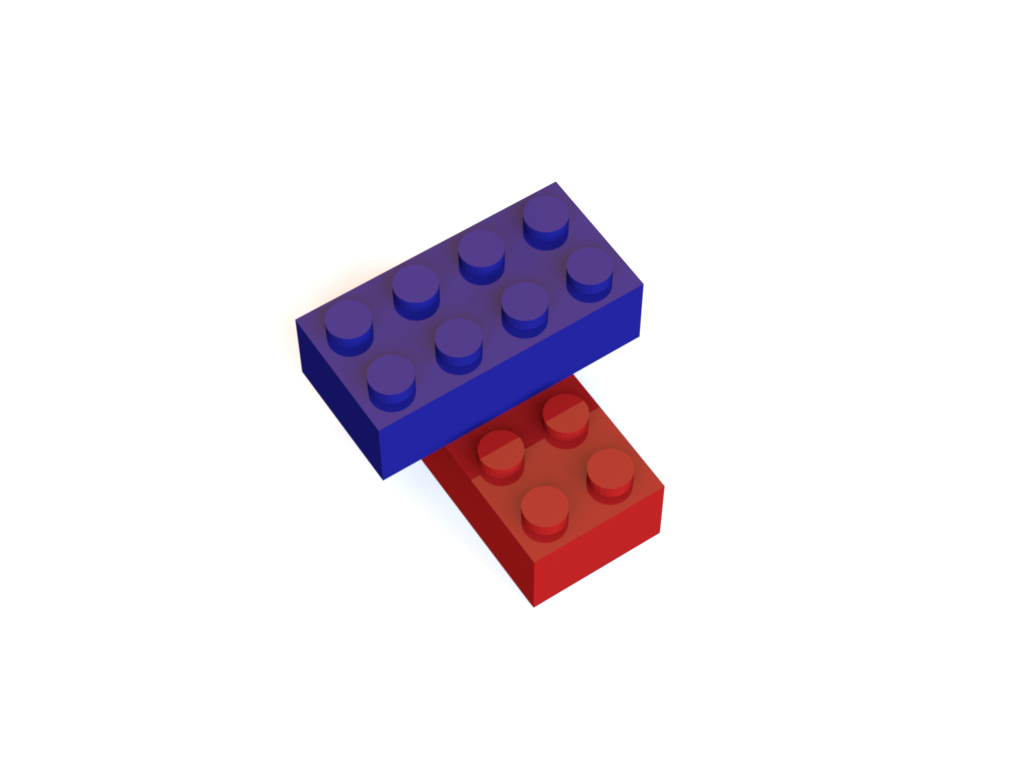}
}
\subfigure{
    \includegraphics[width=0.14\textwidth]{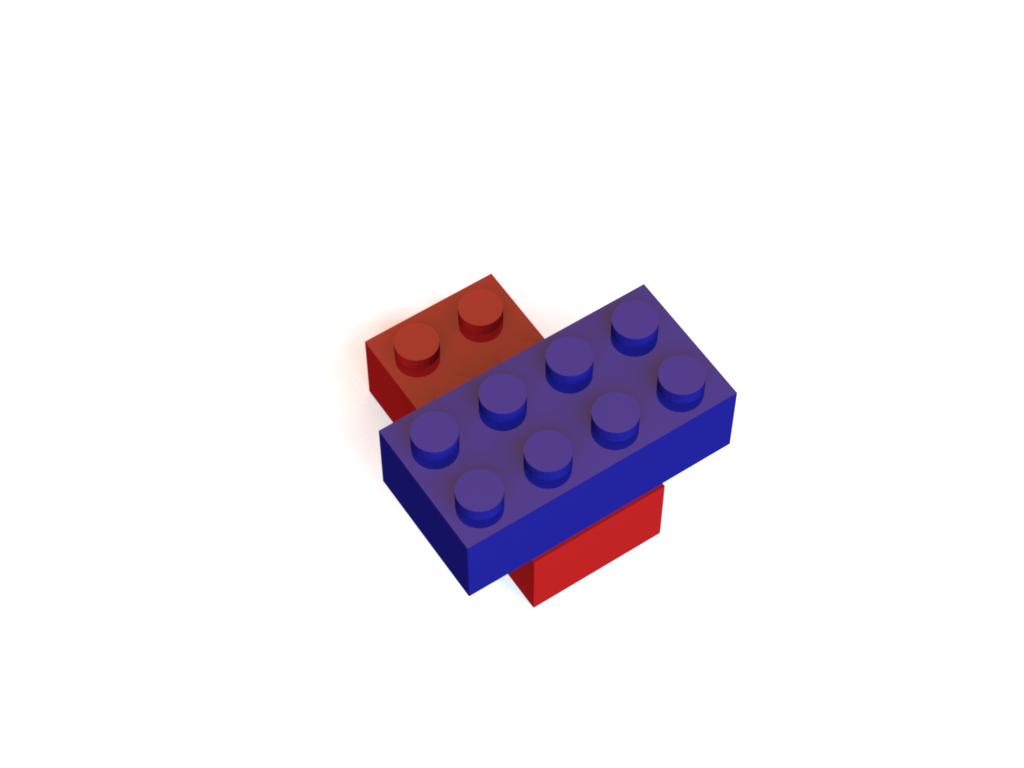}
}
\subfigure{
    \includegraphics[width=0.14\textwidth]{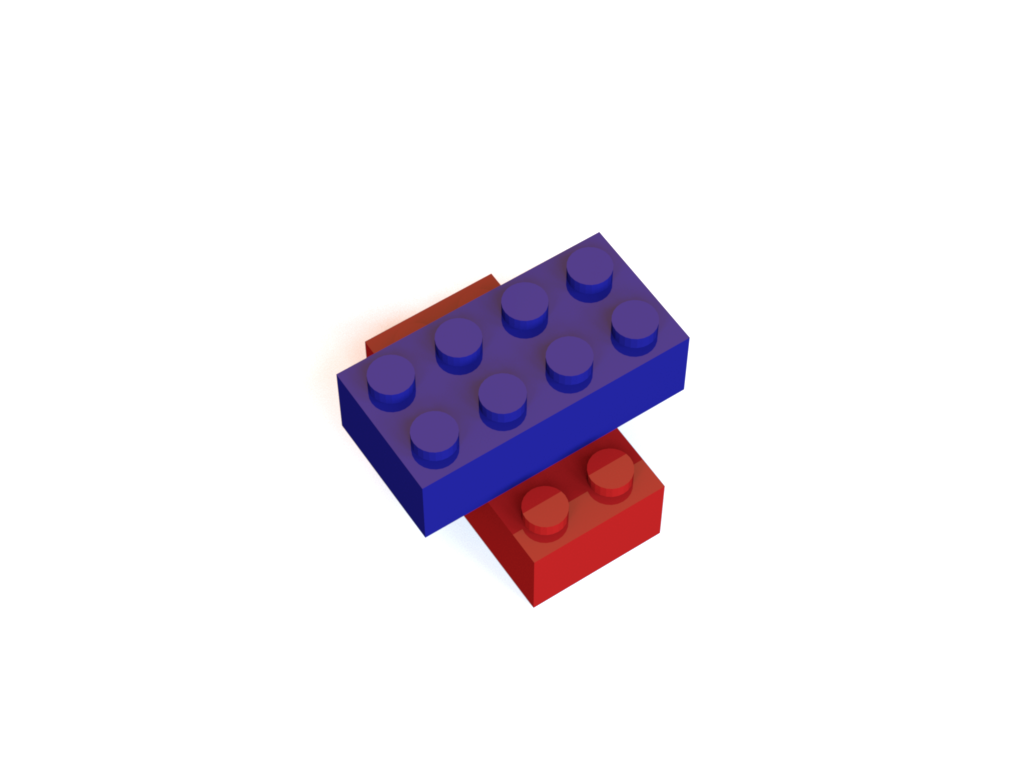}
}
\subfigure{
    \includegraphics[width=0.14\textwidth]{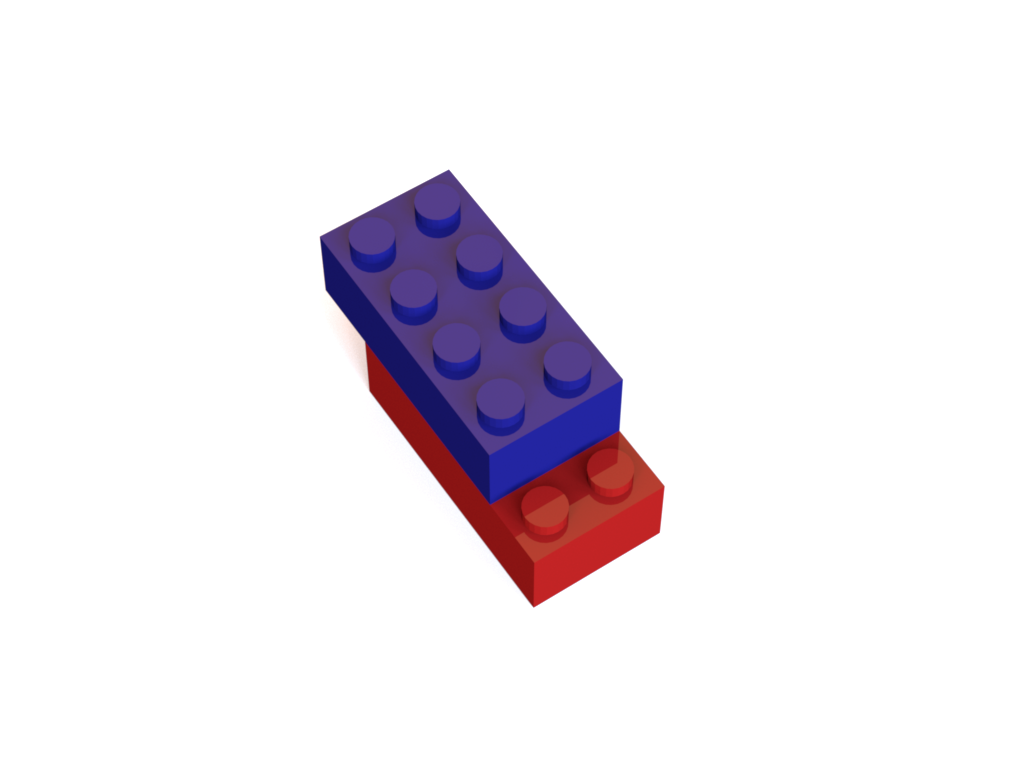}
}
\subfigure{
    \includegraphics[width=0.14\textwidth]{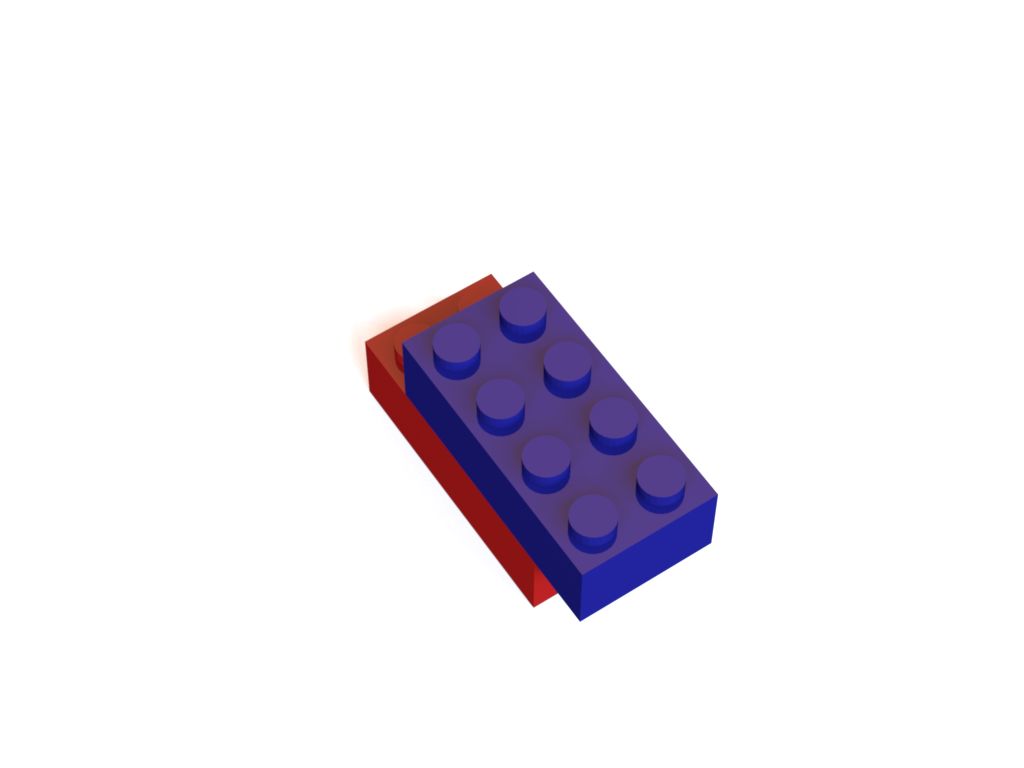}
}
\subfigure{
    \includegraphics[width=0.14\textwidth]{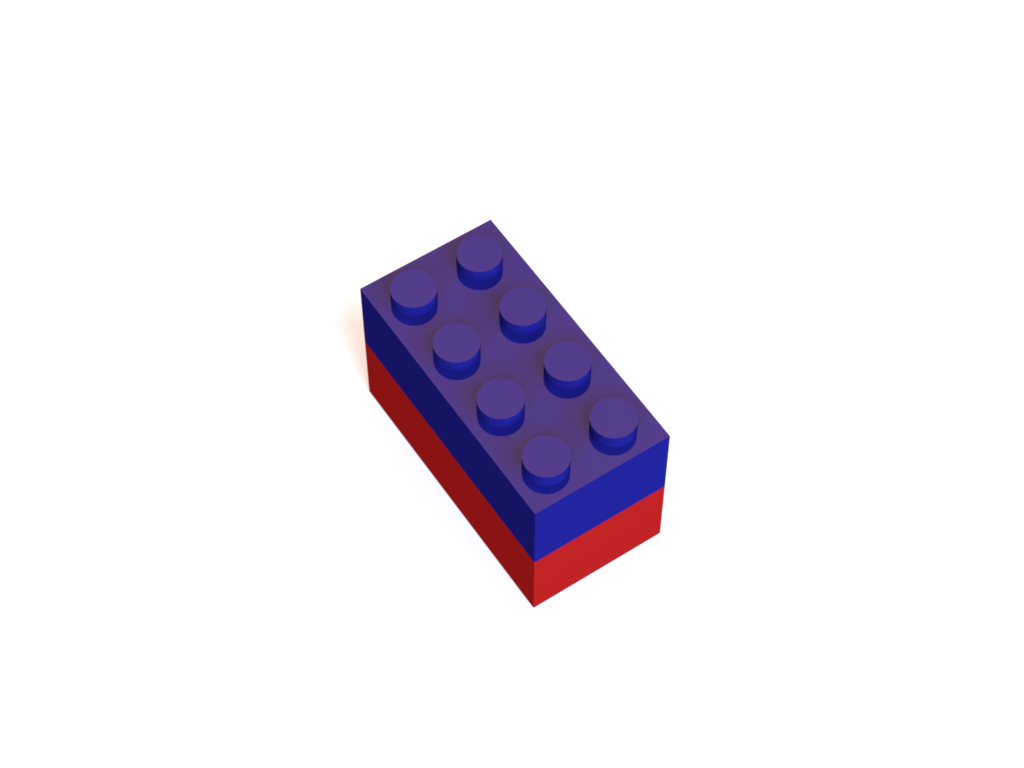}
}
\caption{Visualization of available offsets for randomly-assembled object construction.\label{fig:sup_artificial_offset}}
\end{figure}

\begin{table}[t]
    \centering
    \small
    \caption{Hyperparameters for MNIST construction.\label{tab:mnist_hyperparam}}
    \begin{tabular}{lr}
        \toprule
        \textbf{Hyperparameter} & \textbf{Value} \\
        \midrule
        Gradient clipping & 0.5 \\
        Entropy coefficient & 0.01 \\
        The number of timesteps & 512 \\
        Total timesteps & $3 \times 10^5$ \\
        The number of environments & 8 \\
        Learning rate & $1 \times 10^{-4}$ \\
        Gamma & 0.5 \\
        Lambda & 0.9 \\
        The number of epochs & 6 \\
        The number of mini-batches & 32 \\
        Value coefficient & 1 \\
        \bottomrule
    \end{tabular}
\end{table}

\begin{table}[t]
    \centering
    \small
    \caption{Hyperparameters for other benchmarks.\label{tab:other_hyperparam}}
    \begin{tabular}{lr}
        \toprule
        \textbf{Hyperparameter} & \textbf{Value} \\
        \midrule
        Gradient clipping & 0.5 \\
        Entropy coefficient & 0.01 \\
        The number of timesteps & 512 \\
        Total timesteps & $5 \times 10^5$ \\
        The number of environments & 8 \\
        Learning rate & $1 \times 10^{-4}$ \\
        Gamma & 0.75 \\
        Lambda & 0.9 \\
        The number of epochs & 6 \\
        The number of mini-batches & 32 \\
        Value coefficient & 1 \\
        \bottomrule
    \end{tabular}
\end{table}

Available offset types are illustrated in~\figref{fig:sup_artificial_offset}.
Unlike the experiments of MNIST construction, 
new brick (colored in red) can only be placed 
above the pivot brick (colored in dark blue).
The total number of bricks is chosen uniformly between 10 to 15.
In order to obtain target images, 
we first transform assembled bricks to voxels in closed grid of size $32 \times 32 \times 32$ 
and then crop images of size $14 \times 14$ from different viewpoints 
with the target residing close to the center of each image.

\subsection{ModelNet Construction}

In these experiments, we use the same subset of offset types that are available 
in randomly-assembled object construction 
whereas new brick now can be placed either above or below the pivot brick.
Thus, the total number of available offset types is 32, 
which is exactly twice of randomly-assembled object construction.
The process to acquire images as the desired target information 
is same as in randomly-assembled object construction.

\section{Details of Baseline Methods and Our Method}

In this section, we describe the details of 
baseline methods and our method $\ours$.

\input{tables/tab_baselines}

\subsection{Bayesian Optimization}

We conduct Bayesian optimization~\citep{BrochuE2010arxiv} 
on the tasks we solve, following 
the approach proposed by~\citet{KimJ2020ml4eng}.
Gaussian process regression with Mat\'ern 5/2 kernel and 
expected improvement strategy are 
used as a surrogate function and an acquisition function.
Unless otherwise specified, 
5 initial points and 10 timestep budget 
are given for a single construction step.

\subsection{Supervised Learning Model}

Supervised learning model is built upon the policy network of $\ours$ that is trained with the supervised learning approach instead of the reinforcement learning framework. In detail, sequence-level ground-truth of the pivot and the offset selection is used as cross entropy loss to train the network. Since value prediction of the current state is unnecessary, the value network is dropped. Due to the requirement of the pivot and the offset selection for each timestep as a label, this baseline method is only applicable in randomly-assembled object construction.

\subsection{MLP-based Model}

MLP-based model uses same pipeline as of $\ours$ but with MLPs instead of GNNs to compute features for the pivot and the offset selection. Thus, each brick feature is obtained without message passing between its neighbors. Value or estimated return of the current state, however, is computed similarly by using global average pooling over final brick features.

\subsection{Brick-by-Brick}

Implementation details of our method $\ours$ can be found in~\tabref{tab:sup_arch}.
The number of hidden units in both multi-layer perceptrons and convolutional neural networks is 64 if experiments are the MNIST construction task, 
or 192 otherwise. In both randomly-assembled object construction and ModelNet construction experiments, 
the dimension of target feature computed by $\cnn_{\tar}$ is then 192 by concatenating separately computed features of three images.
The output dimension of $\mlp_{\piv}$ is fixed to 
$N_{\textrm{max}}$ which is 70 in ModelNet construction and 45 in the other experiments.
Typically, the number of maximum bricks or the budget for target objects is predefined 
to values below $N_{\textrm{max}}$.
This can be replaced to a recurrent neural network 
such as Pointer networks~\citep{VinyalsO2015neurips} if no mask information is given.

\section{Additional Experimental Results}

\begin{figure}[t]
\centering
\subfigure[Class 1]{
    \includegraphics[width=0.3\columnwidth]{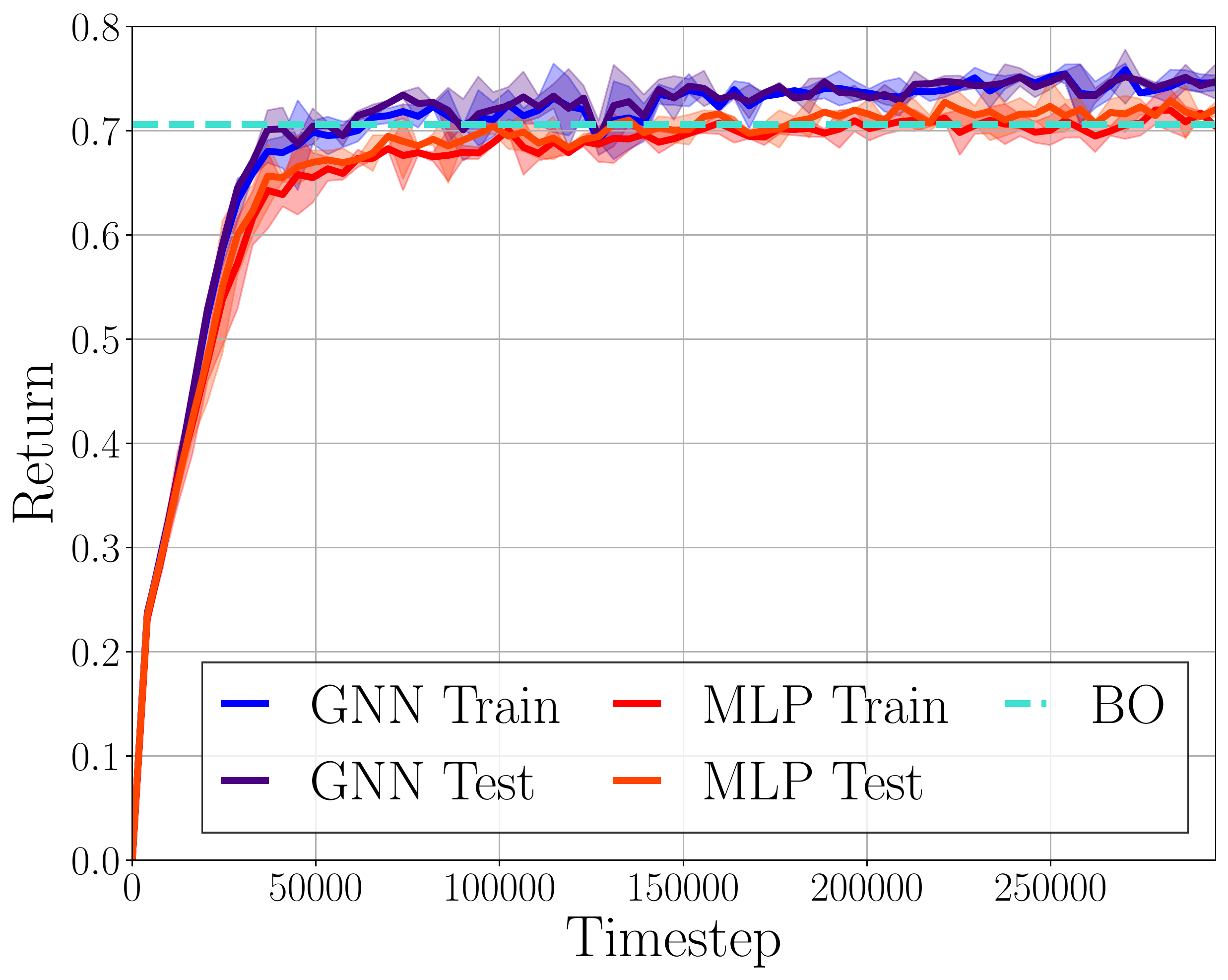}
    \label{fig:mnist_class_1}
}
\subfigure[Class 2]{
    \includegraphics[width=0.3\columnwidth]{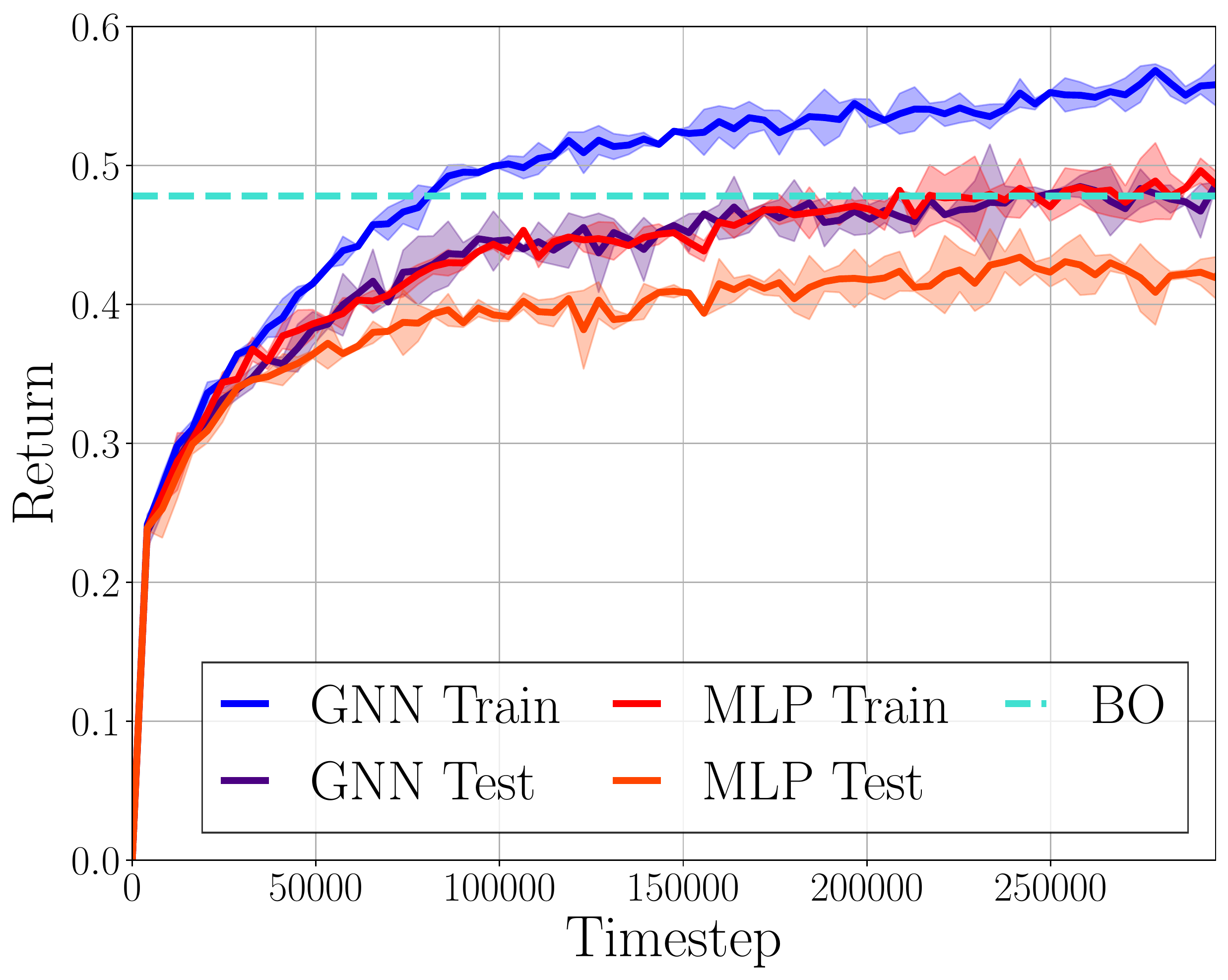}
    \label{fig:mnist_class_2}
}
\subfigure[Class 3]{
    \includegraphics[width=0.3\columnwidth]{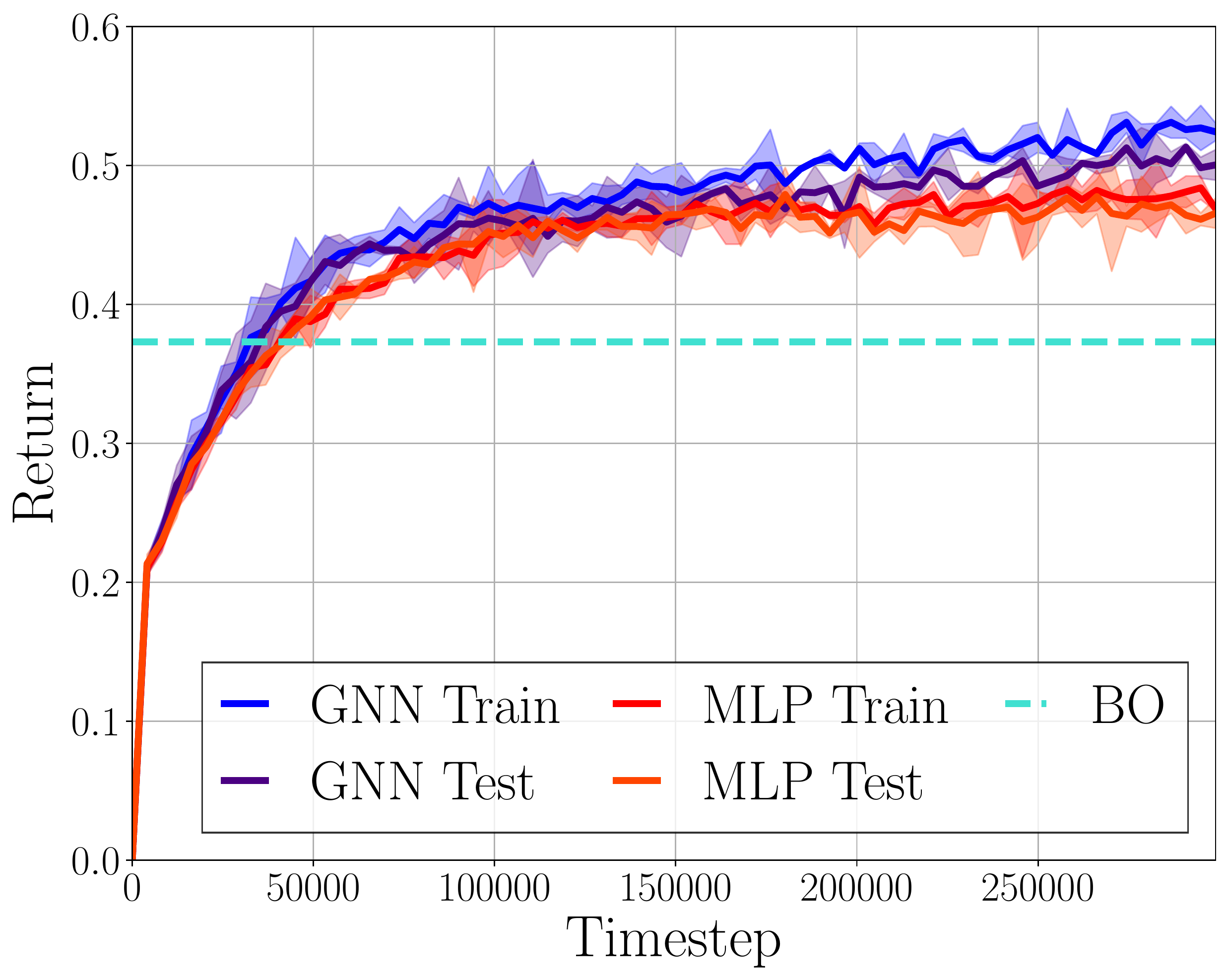}
    \label{fig:mnist_class_3}
}
\subfigure[Class 4]{
    \includegraphics[width=0.3\columnwidth]{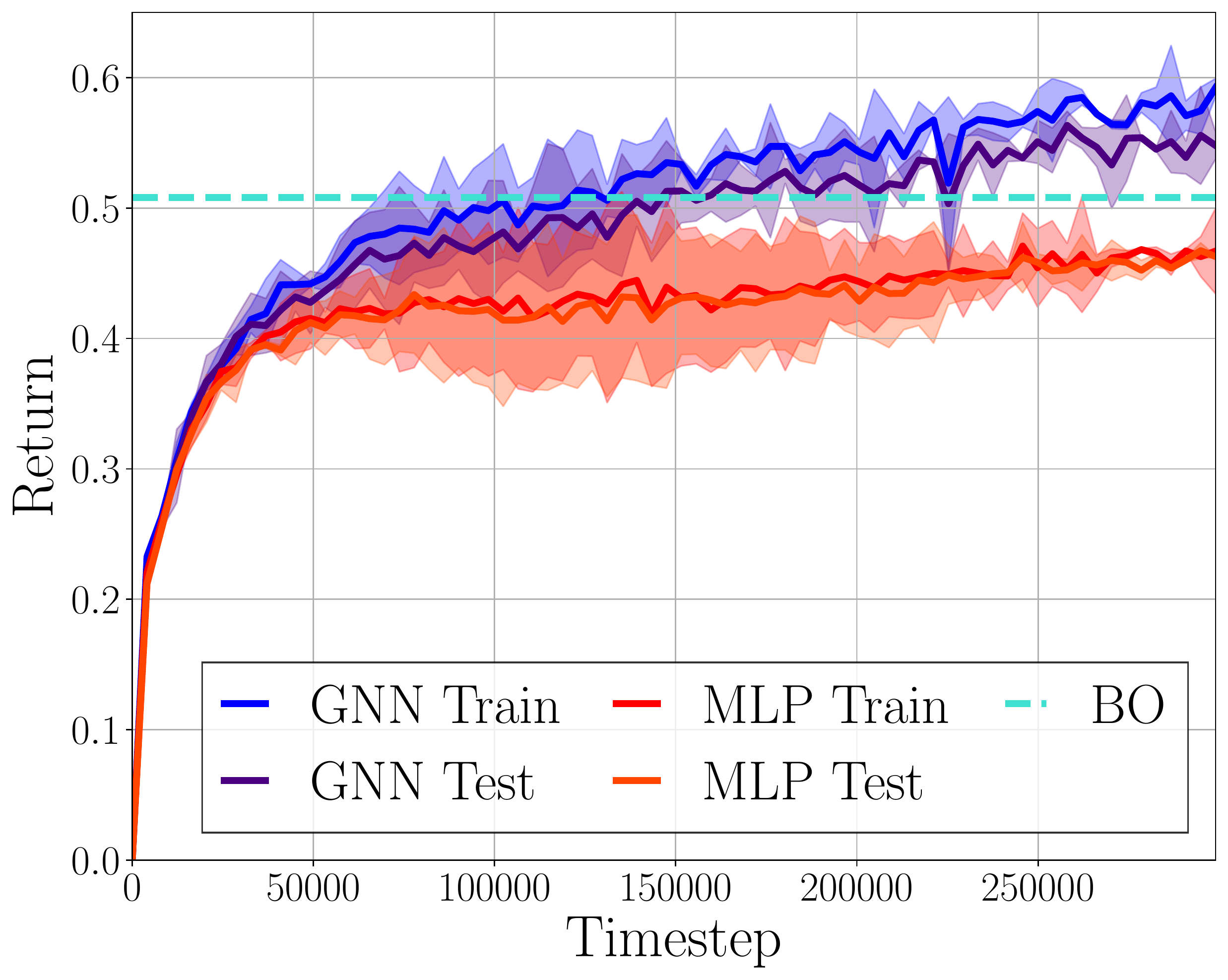}
    \label{fig:mnist_class_4}
}
\subfigure[Class 5]{
    \includegraphics[width=0.3\columnwidth]{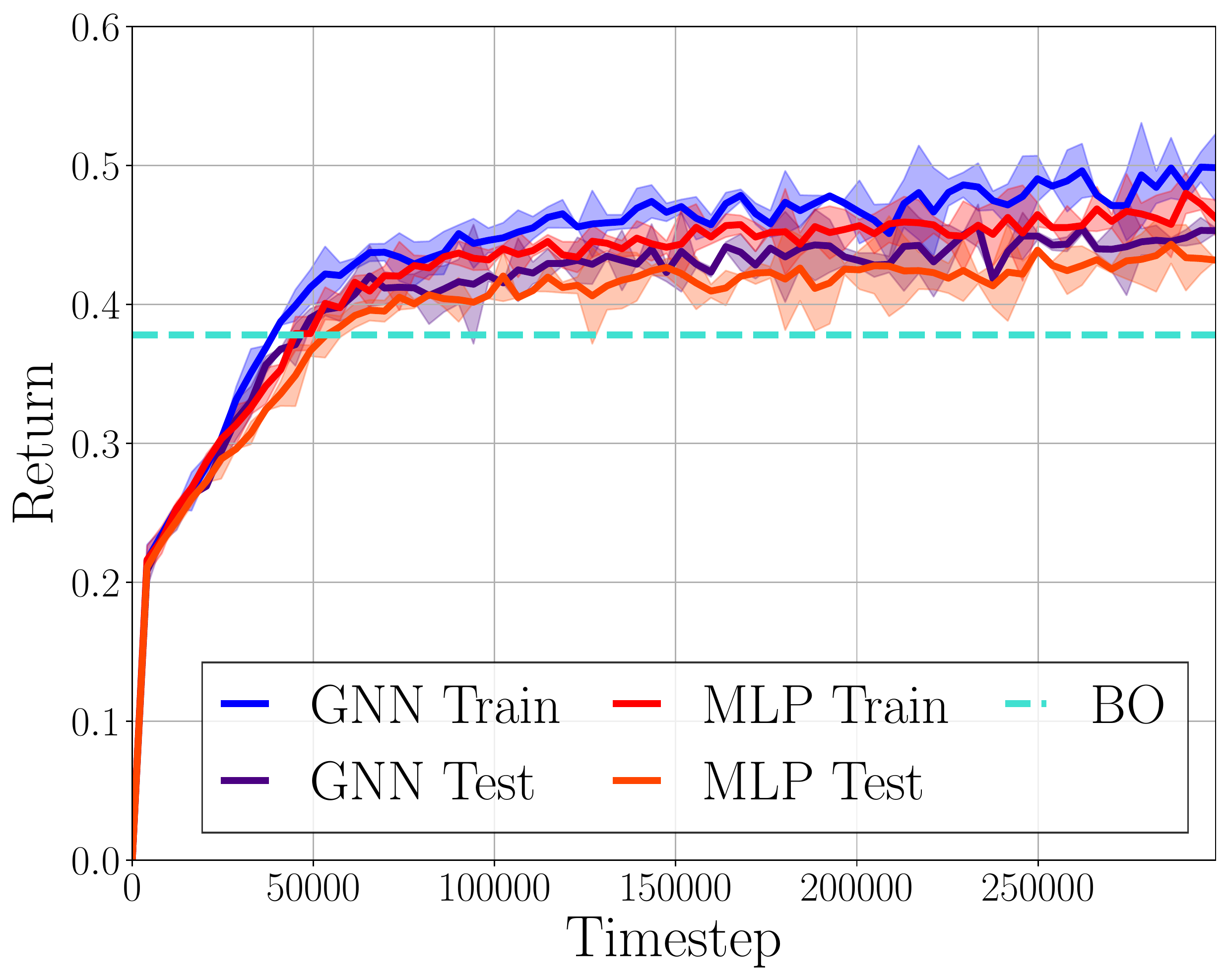}
    \label{fig:mnist_class_5}
}
\subfigure[Class 6]{
    \includegraphics[width=0.3\columnwidth]{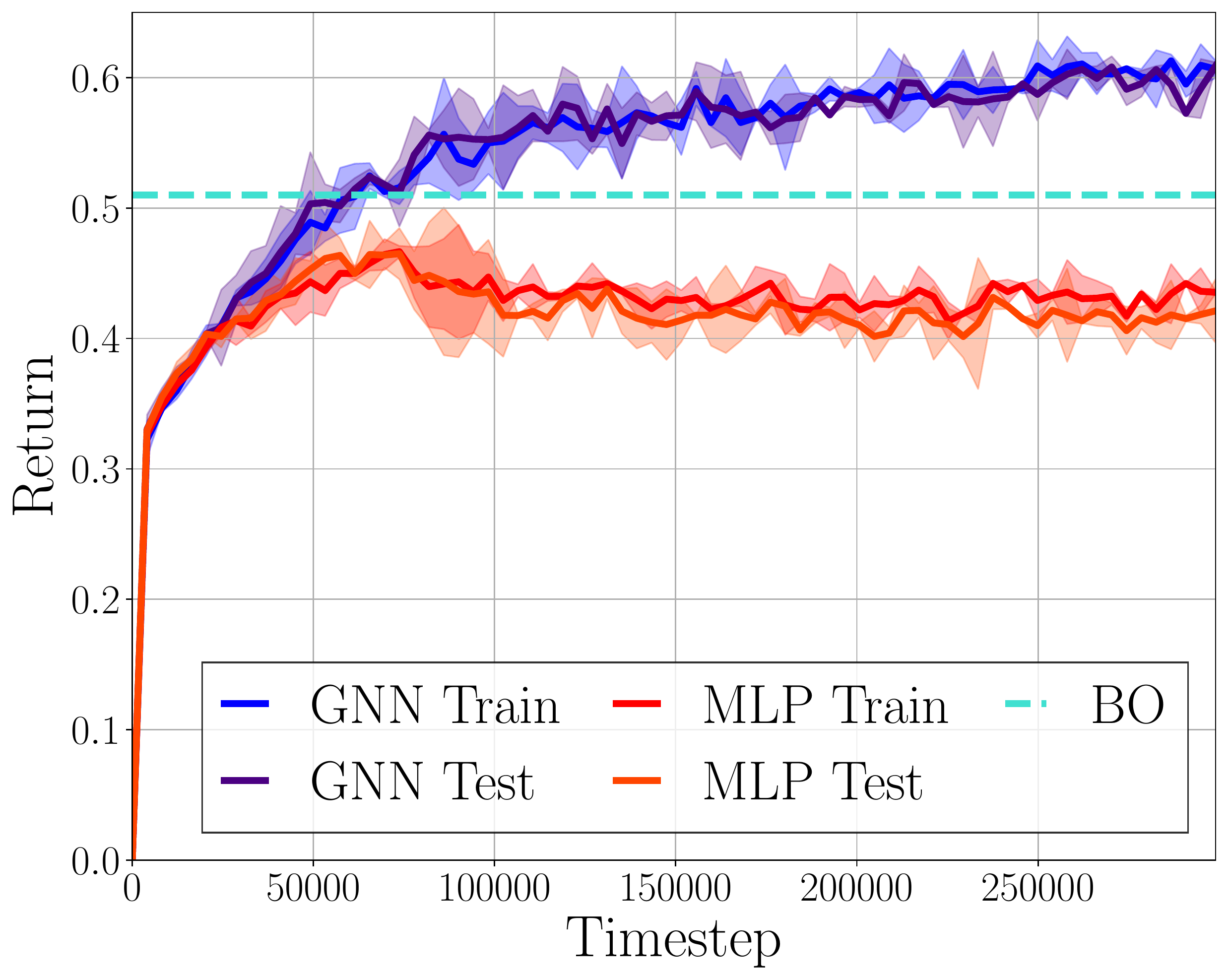}
    \label{fig:mnist_class_6}
}
\subfigure[Class 7]{
    \includegraphics[width=0.3\columnwidth]{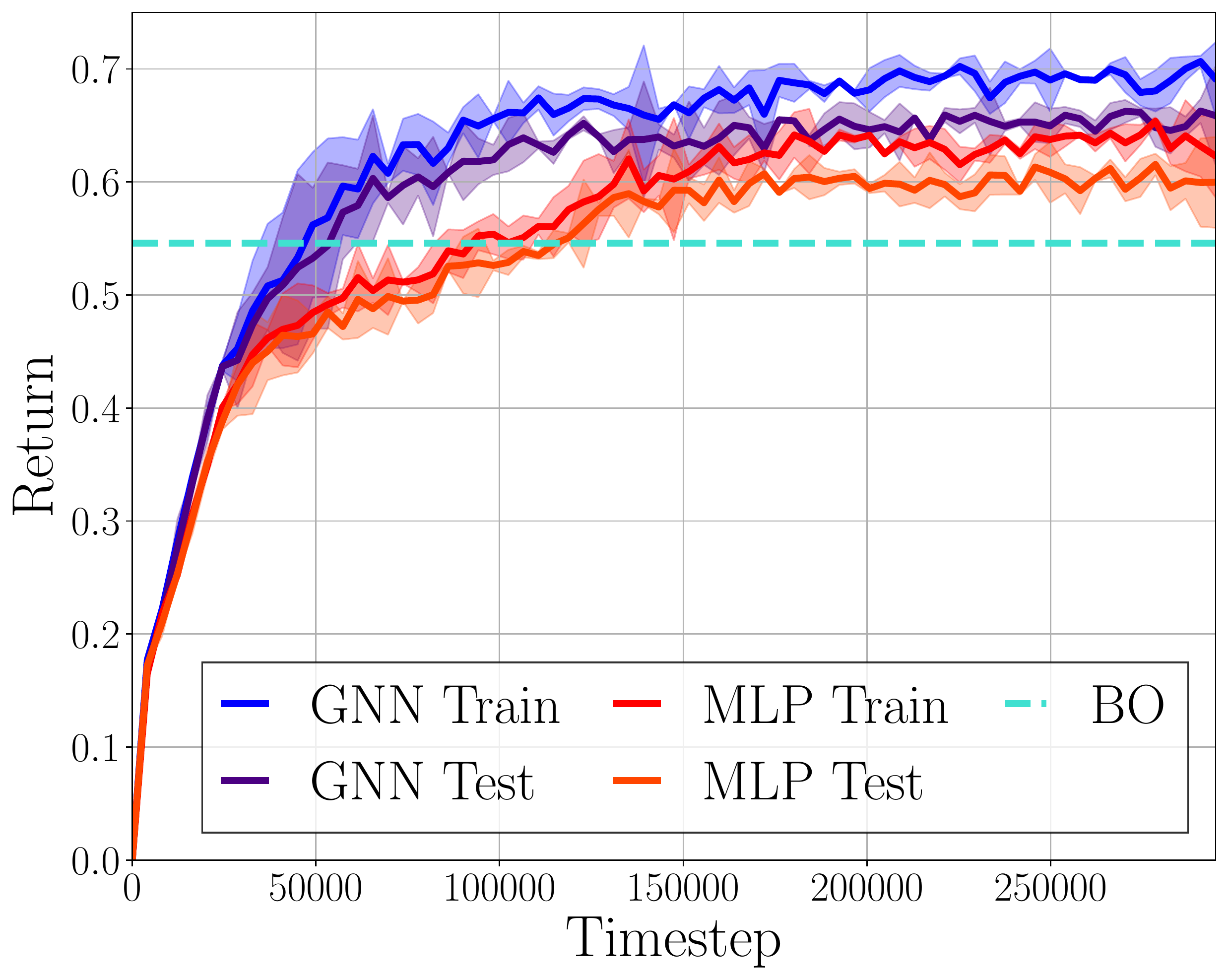}
    \label{fig:mnist_class_7}
}
\subfigure[Class 8]{
    \includegraphics[width=0.3\columnwidth]{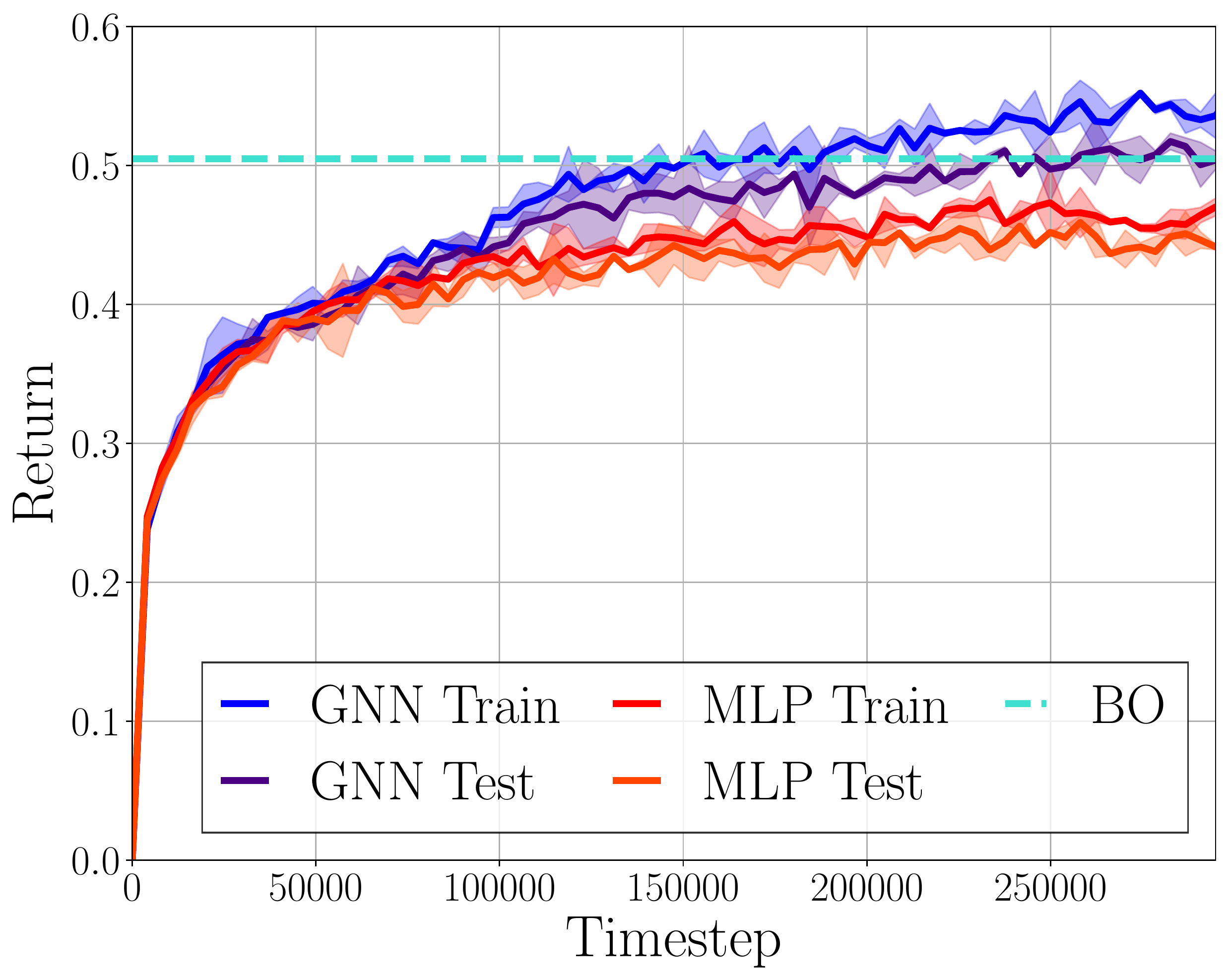}
    \label{fig:mnist_class_8}
}
\subfigure[Class 9]{
    \includegraphics[width=0.3\columnwidth]{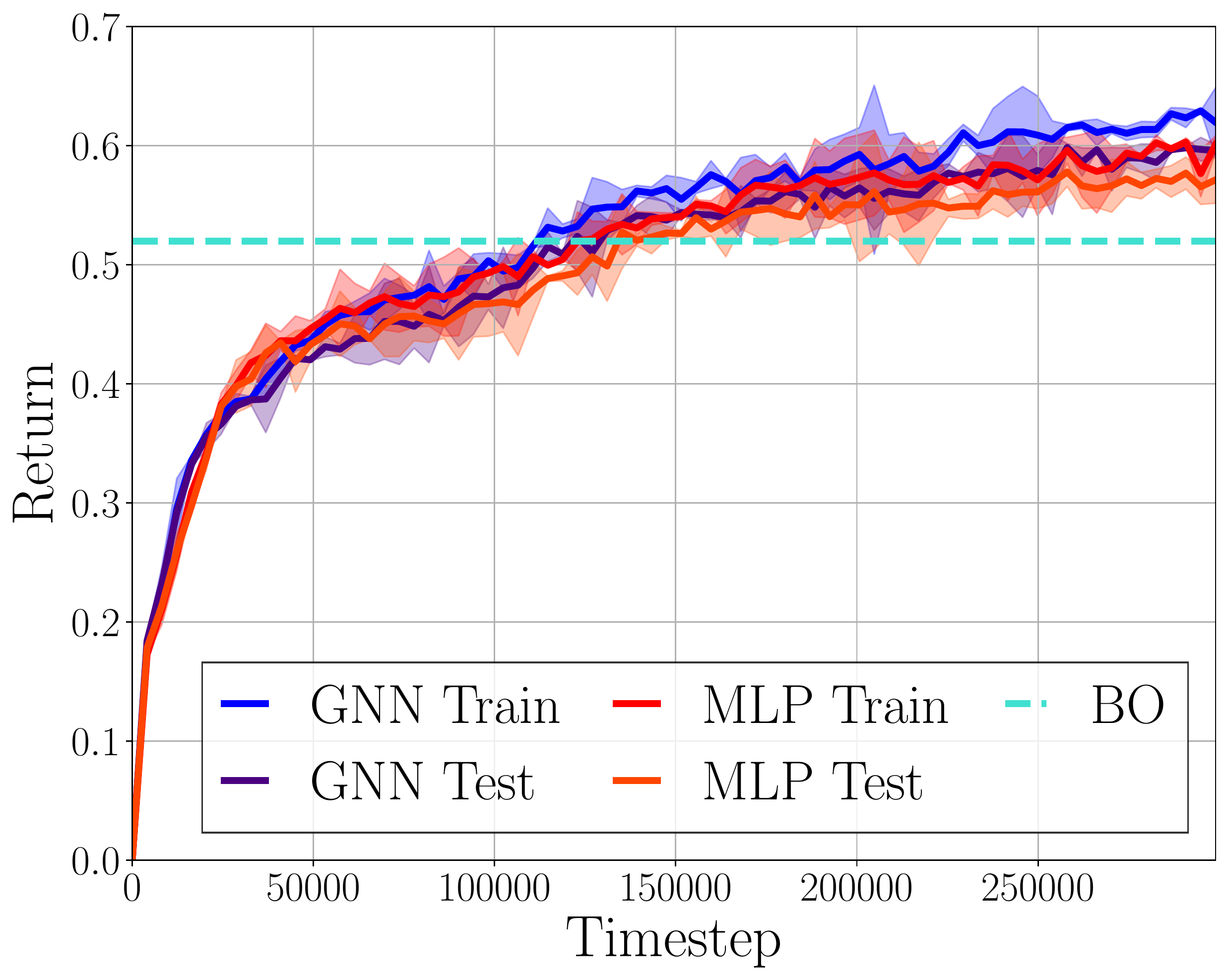}
    \label{fig:mnist_class_9}
}
\caption{Episode return curve for classes 1 to 9 in MNIST construction. Results are averaged over 3 random seeds.}
\label{fig:mnist_graph}
\end{figure}

\begin{figure}[t]
\centering
\subfigure[Monitor]{
    \includegraphics[width=0.47\columnwidth]{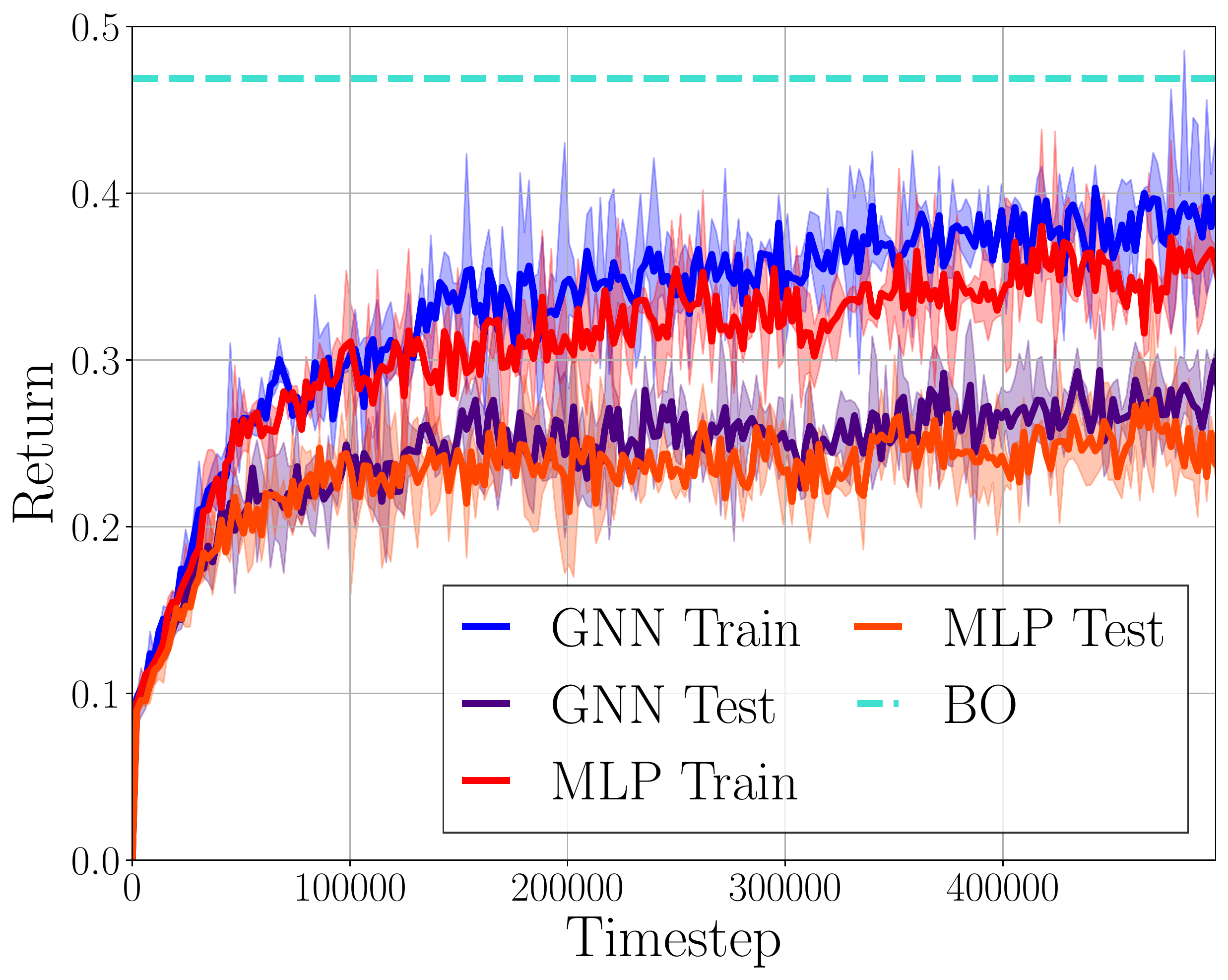}
    \label{fig:modelnet_monitor}
}
\subfigure[Table]{
    \includegraphics[width=0.47\columnwidth]{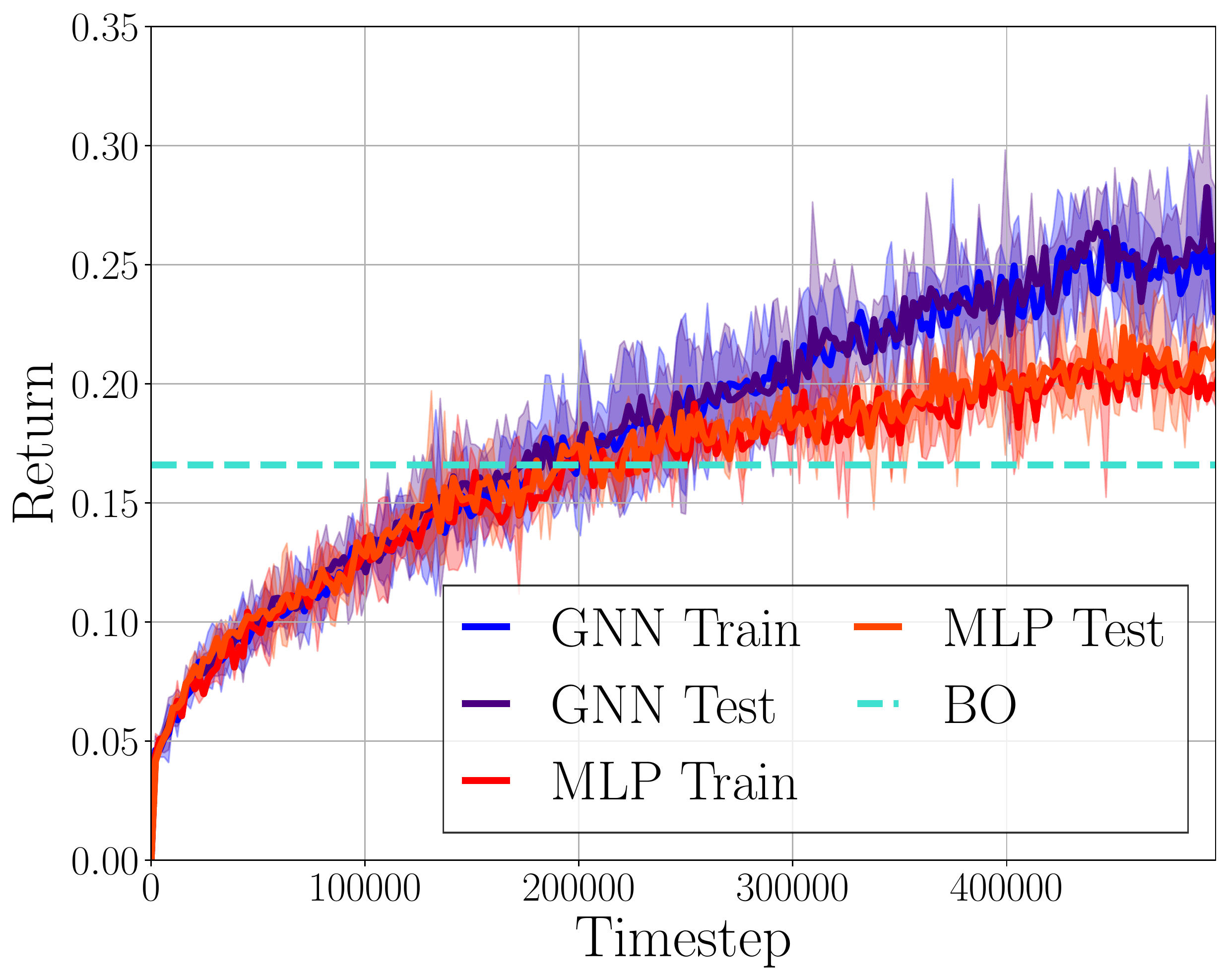}
    \label{fig:modelnet_table}
}
\caption{Episode return curve for monitor and table categories in ModelNet construction. Results are averaged over 3 random seeds.}
\label{fig:modelnet_graph}
\end{figure}

\begin{figure}[t]
    \centering
    \includegraphics[width=0.47\columnwidth]{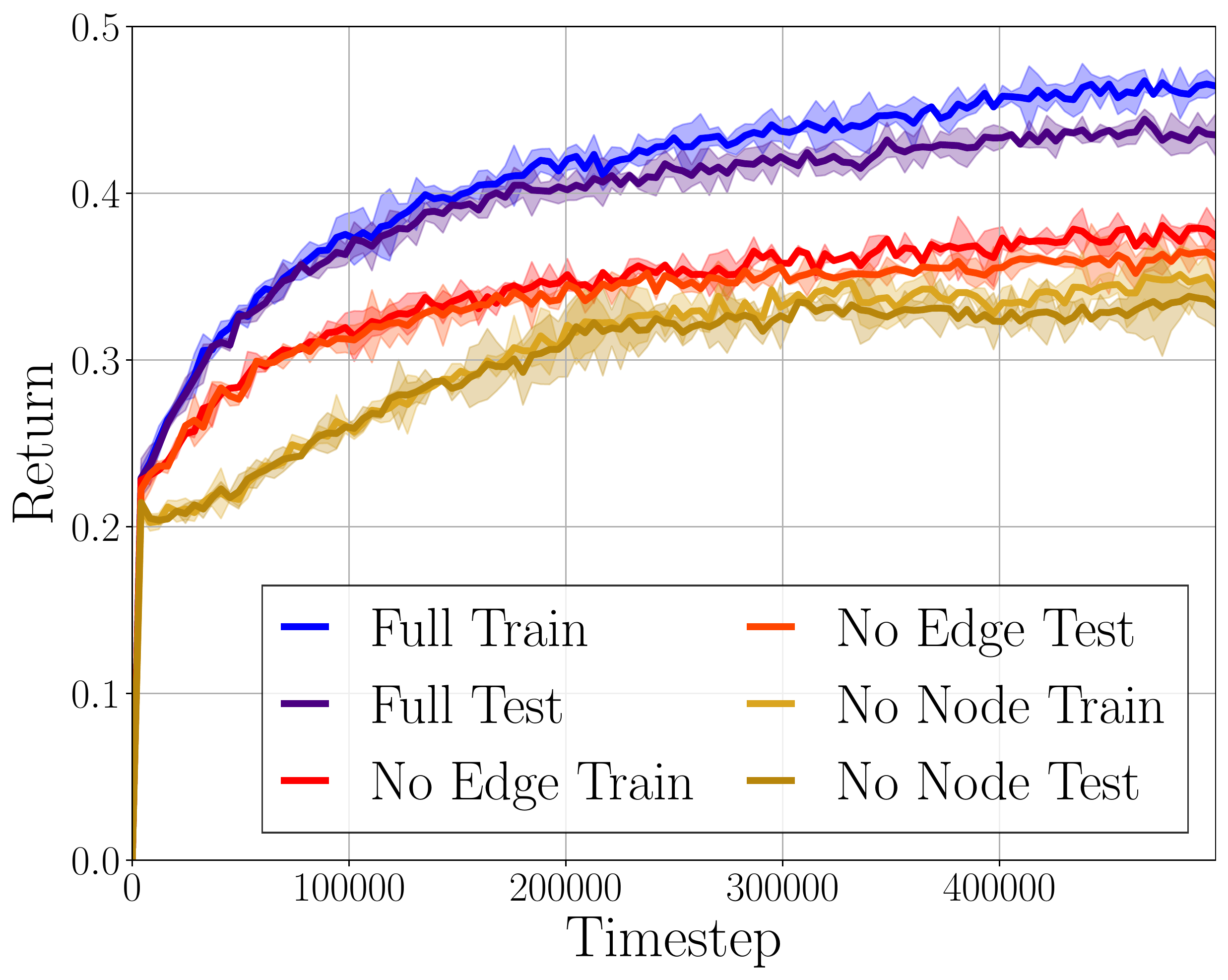}
    \caption{Episode return curve of different GNNs for randomly-assembled object construction. Results are averaged over 3 random seeds.}
    \label{fig:gnn_ablation}
\end{figure}

\begin{table}[ht]
    \centering
    \small
    \caption{Architecture of $\ours$. An asterisk $^*$ implies that its dimension can be changed according to a target benchmark.\label{tab:sup_arch}}
    \begin{tabular}{cccc}
        \toprule
        \multirow{2}{*}{\textbf{Network}} & \multirow{2}{*}{\textbf{Hidden Layer}} & \multirow{2}{*}{\textbf{Activation}} & \textbf{Output} \\
        &&& \textbf{Dimension} \\
        \midrule
        $\mlp_{v}$ & FC & ReLU & 64$^*$ \\
        & FC & ReLU & 64$^*$ \\
        & FC & Linear & 64$^*$ \\ 
        \midrule
        $\mlp_{e}$ & FC & ReLU & 64$^*$ \\
        & FC & ReLU & 64$^*$ \\
        & FC & Linear & 64$^*$ \\ 
        \midrule
        $\cnn_{\tar}$ & Conv2D, 32 channels, $3 \times 3$ filter, stride 1, same padding & Linear & $14 \times 14 \times 128$ \\
        & Maxpool 2D, pool size 3, strides 2, same padding & ReLU & $7 \times 7 \times 128$ \\
        & Conv2D, 32 channels, $3 \times 3$ filter, stride 1, same padding & ReLU & $7 \times 7 \times 128$ \\
        & Conv2D, 32 channels, $3 \times 3$ filter, stride 1, same padding & ReLU & $7 \times 7 \times 128$ \\
        & Conv2D, 32 channels, $3 \times 3$ filter, stride 1, same padding & ReLU & $7 \times 7 \times 128$ \\
        & Conv2D, 32 channels, $3 \times 3$ filter, stride 1, same padding & Linear & $7 \times 7 \times 128$ \\
        & Conv2D, 64 channels, $3 \times 3$ filter, stride 1, same padding & Linear & $7 \times 7 \times 64$ \\
        & Maxpool 2D, pool size 3, strides 2, same padding & ReLU & $4 \times 4 \times 64$ \\
        & Conv2D, 64 channels, $3 \times 3$ filter, stride 1, same padding & ReLU & $4 \times 4 \times 64$ \\
        & Conv2D, 64 channels, $3 \times 3$ filter, stride 1, same padding & ReLU & $4 \times 4 \times 64$ \\
        & Conv2D, 64 channels, $3 \times 3$ filter, stride 1, same padding & ReLU & $4 \times 4 \times 64$ \\
        & Conv2D, 64 channels, $3 \times 3$ filter, stride 1, same padding & ReLU & $4 \times 4 \times 64$ \\
        & Flatten & - & 1024 \\
        & FC & Linear & 64$^*$ \\
        \midrule
        $\mlp_{v}^{(\ell)}$ & FC & ReLU & 64$^*$ \\
        \midrule
        $\mlp_{v}^{(\ell)}$ & FC & ReLU & 64$^*$ \\
        \midrule
        $\mlp_{\piv}$ & FC & Softmax & $N_{\textrm{max}}$ \\
        \midrule
        $\mlp_{\off}$ & FC & Softmax & $N_{\off}$ \\
        \midrule
        $\mlp_{\textrm{val}}$ & FC & Linear & 1 \\
        \bottomrule
    \end{tabular}
\end{table}

All experiments are carried out on a Ubuntu 16.04 workstation, consisting of Intel(R) Core(TM) i7-6850K CPU and two NVIDIA Titan X Pascal GPUs.

\begin{figure}[ht]
\centering
\subfigure{
    \includegraphics[width=0.12\textwidth]{MNIST_all/class_0/mnist_0_0_img.png}
}
\subfigure{
    \includegraphics[width=0.16\textwidth]{MNIST_all/class_0/mnist_0_0_coords.png}
}
\subfigure{
    \includegraphics[width=0.12\textwidth]{MNIST_all/class_0/mnist_0_1_img.png}
}
\subfigure{
    \includegraphics[width=0.16\textwidth]{MNIST_all/class_0/mnist_0_1_coords.png}
}
\subfigure{
    \includegraphics[width=0.12\textwidth]{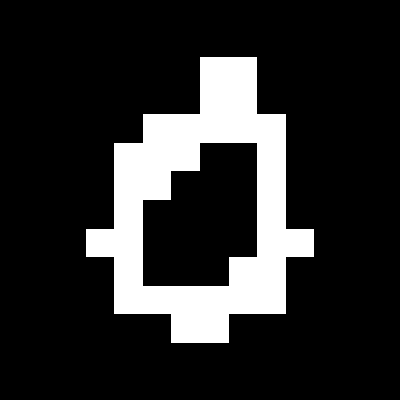}
}
\subfigure{
    \includegraphics[width=0.16\textwidth]{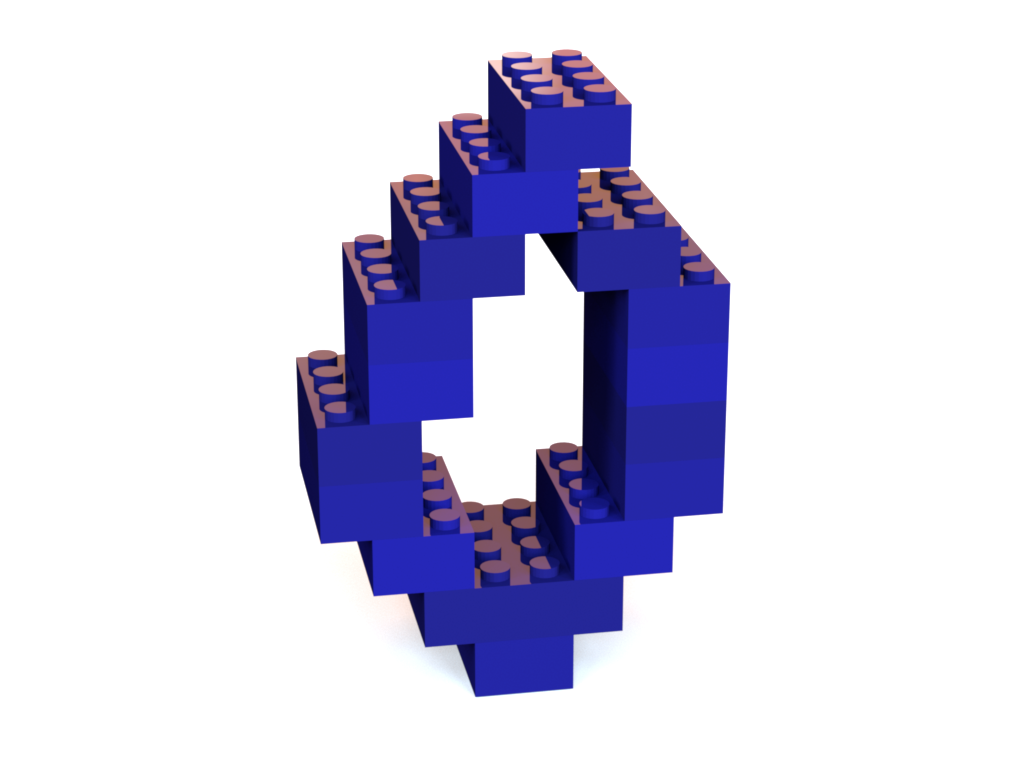}
}

\subfigure{
    \includegraphics[width=0.12\textwidth]{MNIST_all/class_1/mnist_1_0_img.png}
}
\subfigure{
    \includegraphics[width=0.16\textwidth]{MNIST_all/class_1/mnist_1_0_coords.png}
}
\subfigure{
    \includegraphics[width=0.12\textwidth]{MNIST_all/class_1/mnist_1_1_img.png}
}
\subfigure{
    \includegraphics[width=0.16\textwidth]{MNIST_all/class_1/mnist_1_1_coords.png}
}
\subfigure{
    \includegraphics[width=0.12\textwidth]{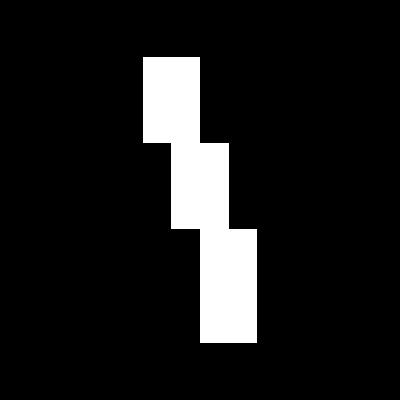}
}
\subfigure{
    \includegraphics[width=0.16\textwidth]{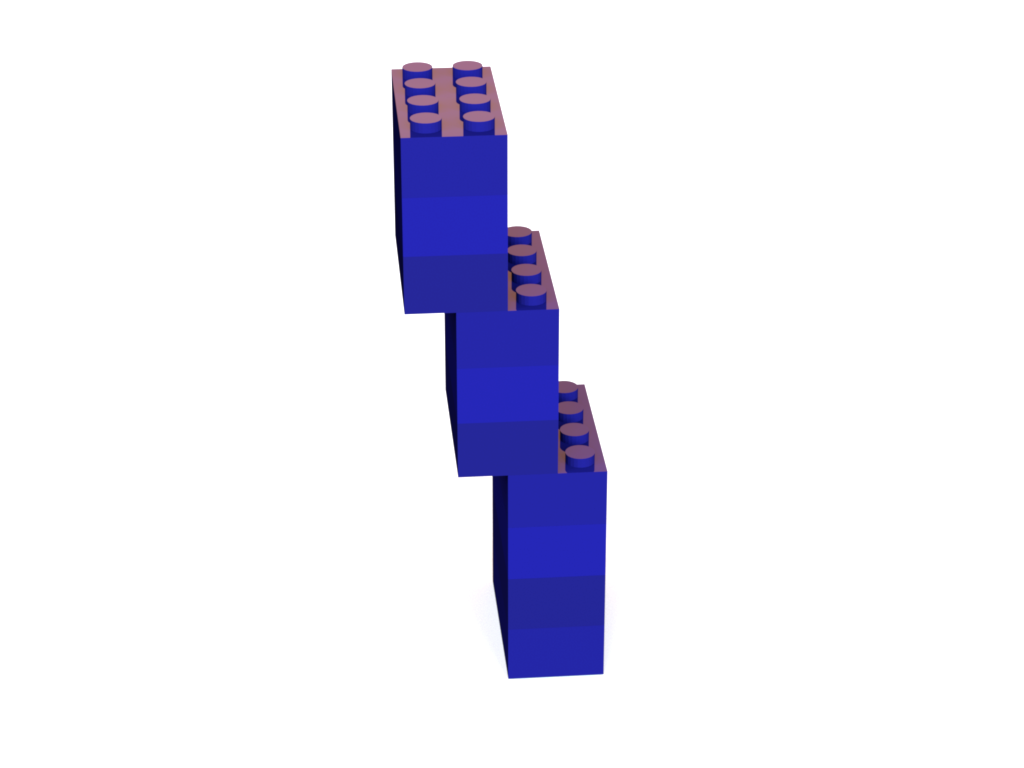}
}

\subfigure{
    \includegraphics[width=0.12\textwidth]{MNIST_all/class_2/mnist_2_0_img.png}
}
\subfigure{
    \includegraphics[width=0.16\textwidth]{MNIST_all/class_2/mnist_2_0_coords.png}
}
\subfigure{
    \includegraphics[width=0.12\textwidth]{MNIST_all/class_2/mnist_2_1_img.png}
}
\subfigure{
    \includegraphics[width=0.16\textwidth]{MNIST_all/class_2/mnist_2_1_coords.png}
}
\subfigure{
    \includegraphics[width=0.12\textwidth]{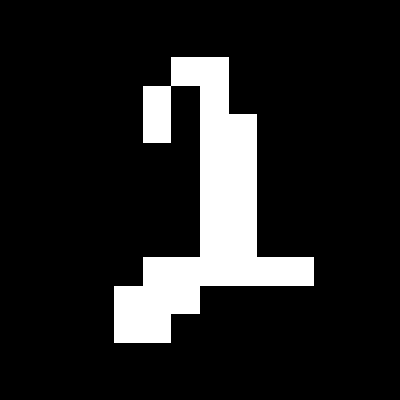}
}
\subfigure{
    \includegraphics[width=0.16\textwidth]{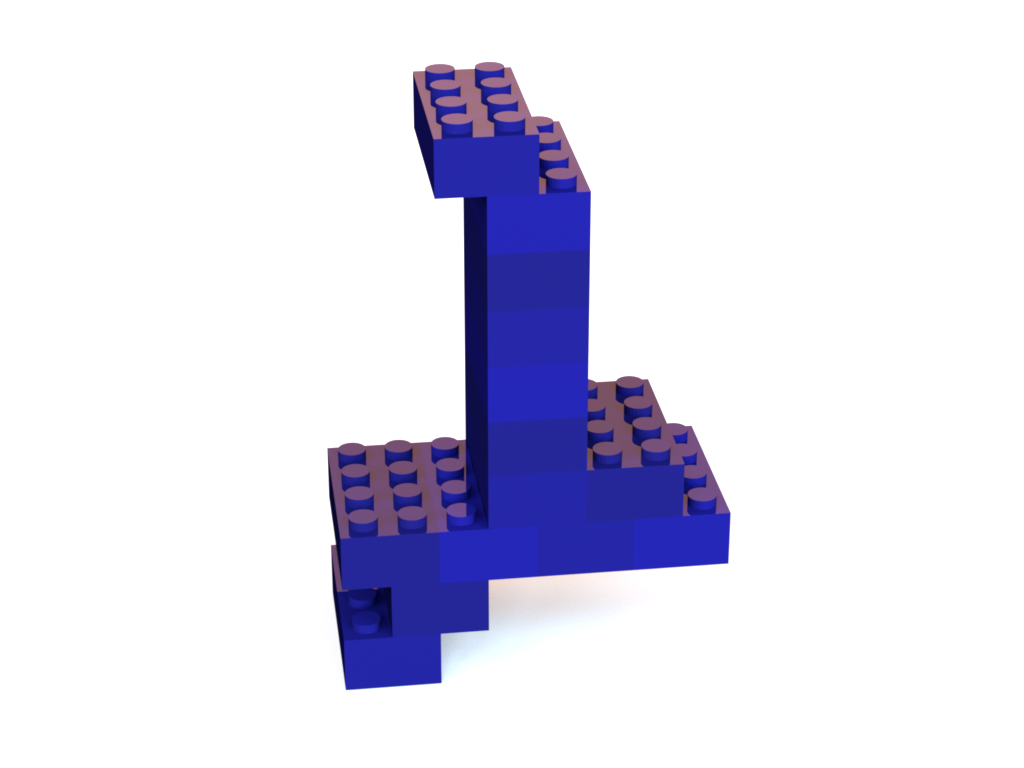}
}

\subfigure{
    \includegraphics[width=0.12\textwidth]{MNIST_all/class_3/mnist_3_0_img.png}
}
\subfigure{
    \includegraphics[width=0.16\textwidth]{MNIST_all/class_3/mnist_3_0_coords.png}
}
\subfigure{
    \includegraphics[width=0.12\textwidth]{MNIST_all/class_3/mnist_3_1_img.png}
}
\subfigure{
    \includegraphics[width=0.16\textwidth]{MNIST_all/class_3/mnist_3_1_coords.png}
}
\subfigure{
    \includegraphics[width=0.12\textwidth]{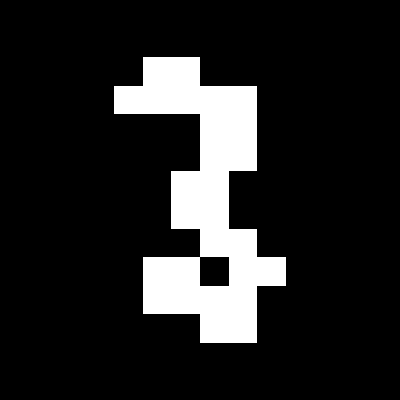}
}
\subfigure{
    \includegraphics[width=0.16\textwidth]{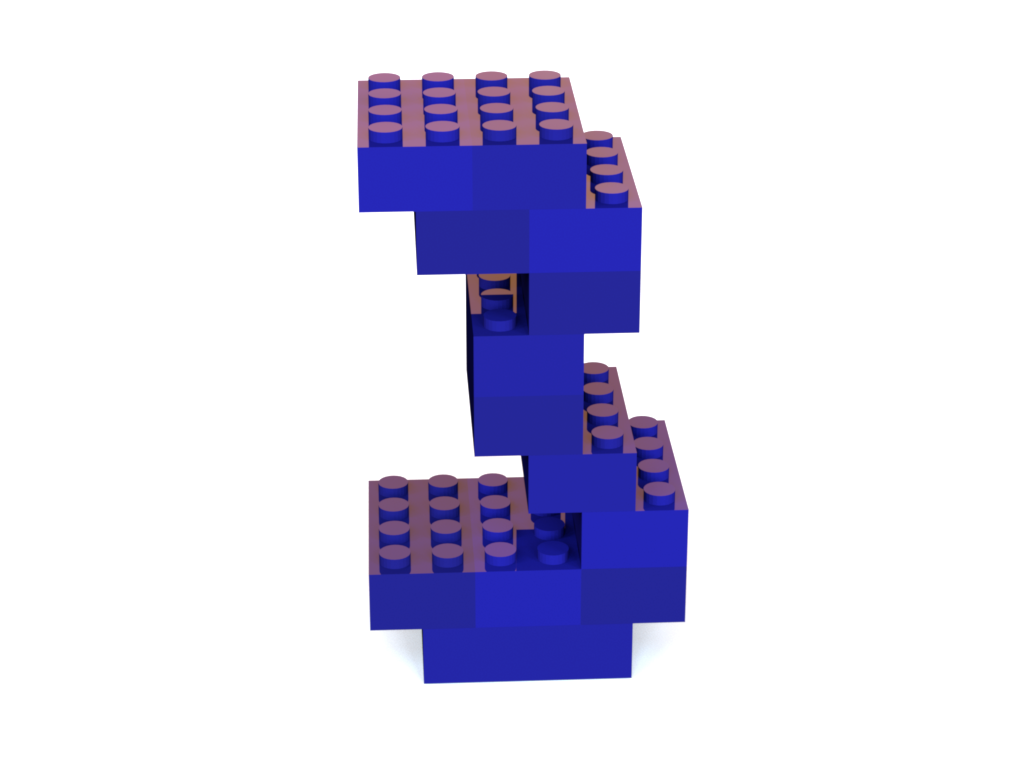}
}

\subfigure{
    \includegraphics[width=0.12\textwidth]{MNIST_all/class_4/mnist_4_0_img.png}
}
\subfigure{
    \includegraphics[width=0.16\textwidth]{MNIST_all/class_4/mnist_4_0_coords.png}
}
\subfigure{
    \includegraphics[width=0.12\textwidth]{MNIST_all/class_4/mnist_4_1_img.png}
}
\subfigure{
    \includegraphics[width=0.16\textwidth]{MNIST_all/class_4/mnist_4_1_coords.png}
}
\subfigure{
    \includegraphics[width=0.12\textwidth]{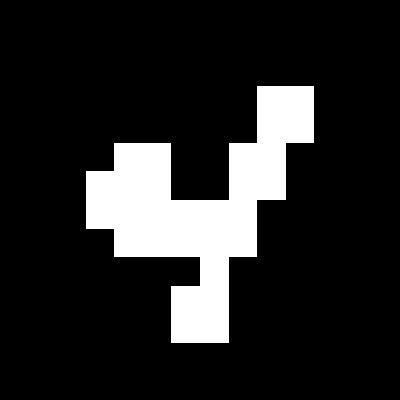}
}
\subfigure{
    \includegraphics[width=0.16\textwidth]{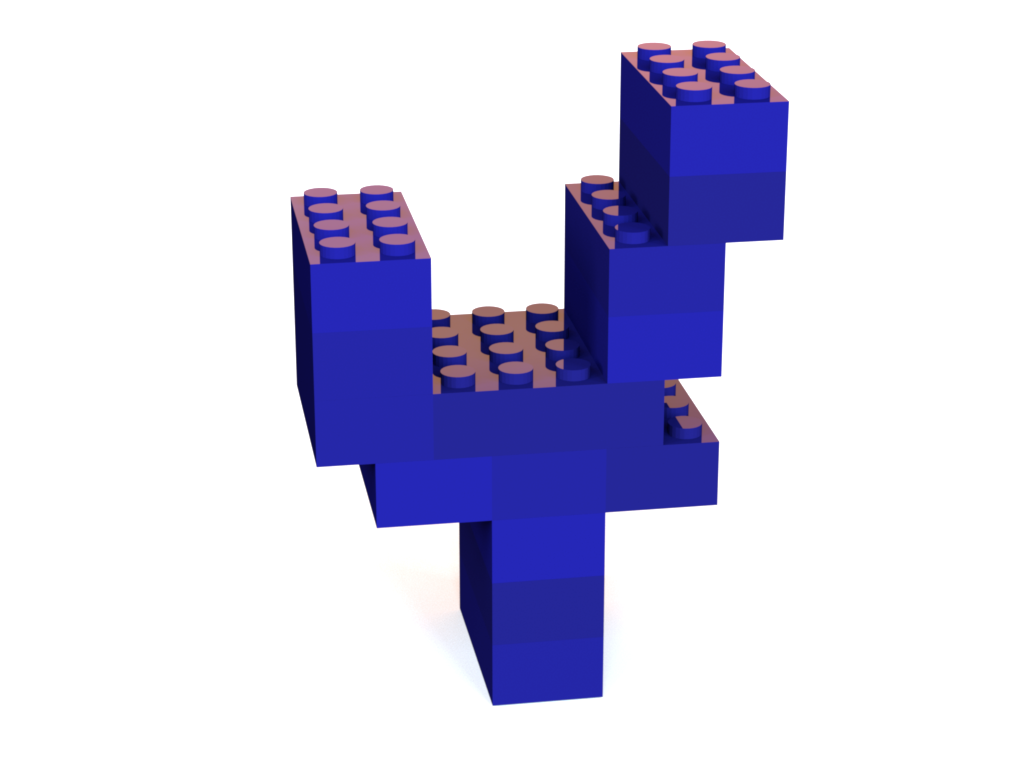}
}

\subfigure{
    \includegraphics[width=0.12\textwidth]{MNIST_all/class_5/mnist_5_0_img.png}
}
\subfigure{
    \includegraphics[width=0.16\textwidth]{MNIST_all/class_5/mnist_5_0_coords.png}
}
\subfigure{
    \includegraphics[width=0.12\textwidth]{MNIST_all/class_5/mnist_5_1_img.png}
}
\subfigure{
    \includegraphics[width=0.16\textwidth]{MNIST_all/class_5/mnist_5_1_coords.png}
}
\subfigure{
    \includegraphics[width=0.12\textwidth]{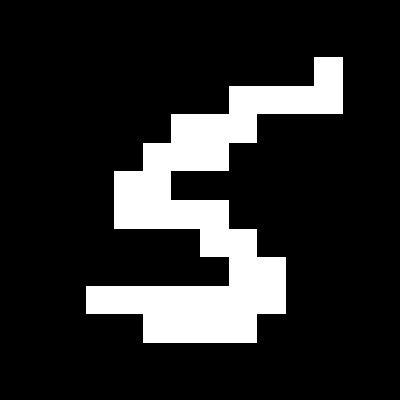}
}
\subfigure{
    \includegraphics[width=0.16\textwidth]{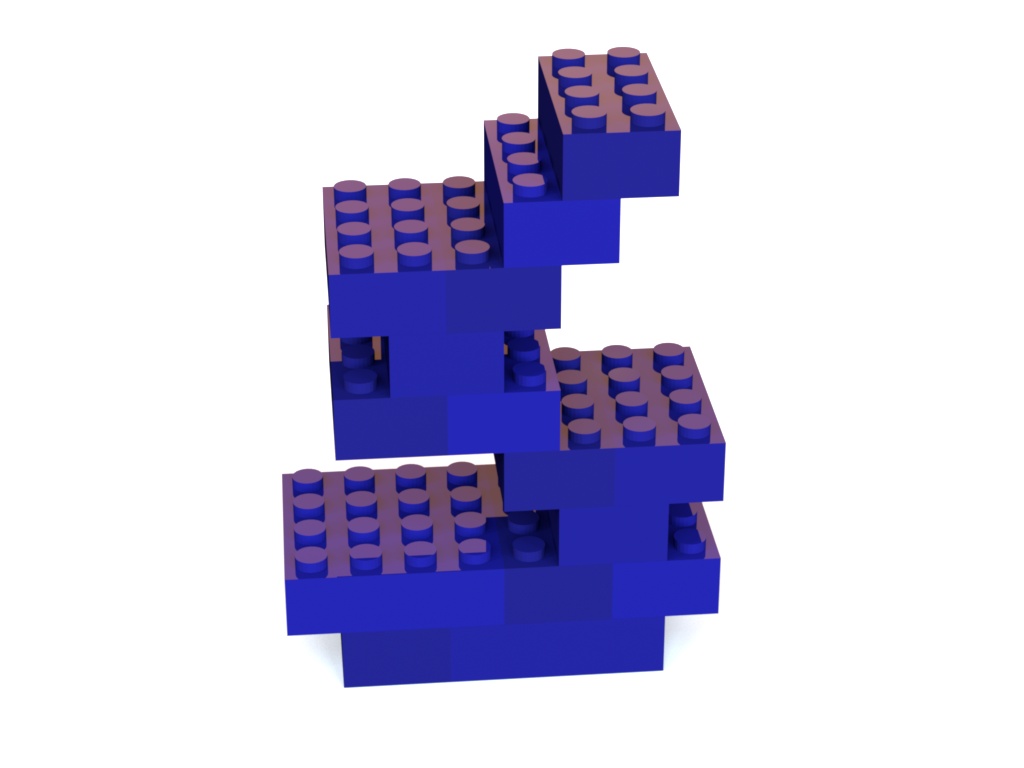}
}

\subfigure{
    \includegraphics[width=0.12\textwidth]{MNIST_all/class_6/mnist_6_0_img.png}
}
\subfigure{
    \includegraphics[width=0.16\textwidth]{MNIST_all/class_6/mnist_6_0_coords.png}
}
\subfigure{
    \includegraphics[width=0.12\textwidth]{MNIST_all/class_6/mnist_6_1_img.png}
}
\subfigure{
    \includegraphics[width=0.16\textwidth]{MNIST_all/class_6/mnist_6_1_coords.png}
}
\subfigure{
    \includegraphics[width=0.12\textwidth]{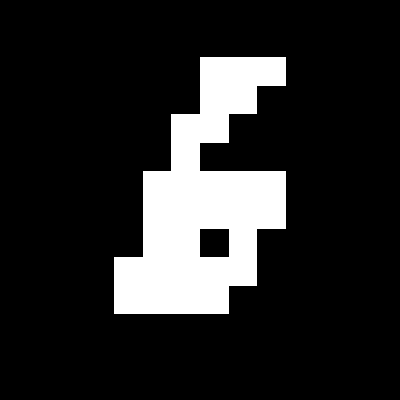}
}
\subfigure{
    \includegraphics[width=0.16\textwidth]{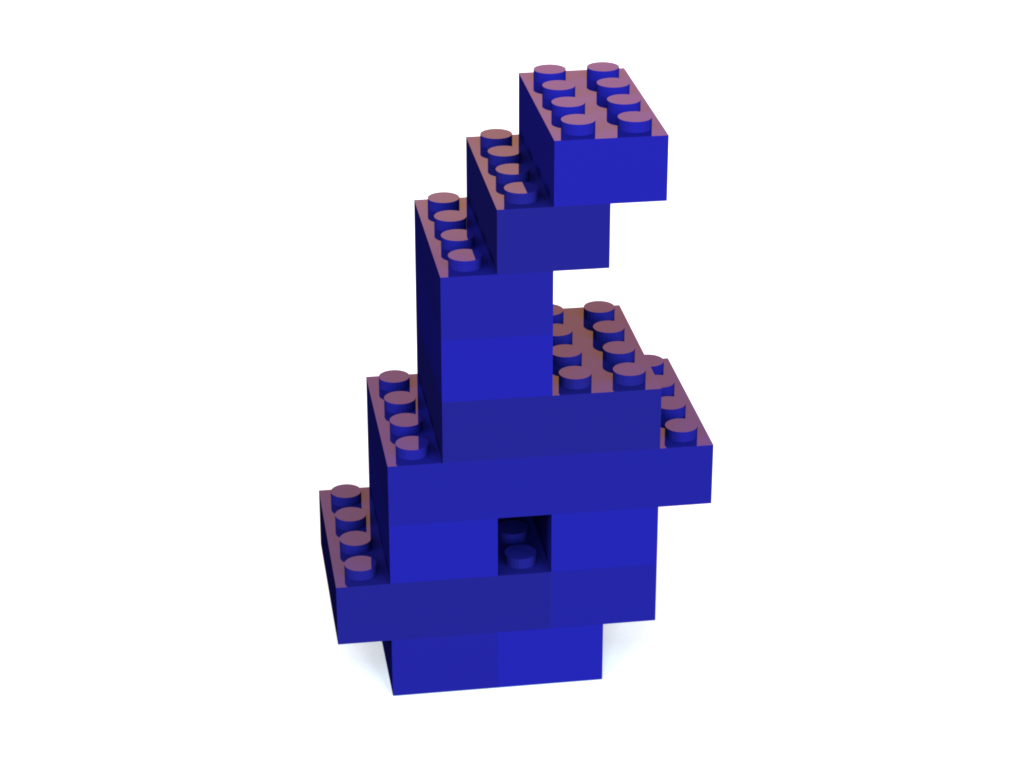}
}

\subfigure{
    \includegraphics[width=0.12\textwidth]{MNIST_all/class_7/mnist_7_0_img.png}
}
\subfigure{
    \includegraphics[width=0.16\textwidth]{MNIST_all/class_7/mnist_7_0_coords.png}
}
\subfigure{
    \includegraphics[width=0.12\textwidth]{MNIST_all/class_7/mnist_7_1_img.png}
}
\subfigure{
    \includegraphics[width=0.16\textwidth]{MNIST_all/class_7/mnist_7_1_coords.png}
}
\subfigure{
    \includegraphics[width=0.12\textwidth]{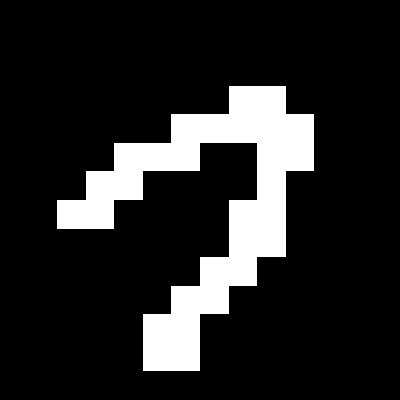}
}
\subfigure{
    \includegraphics[width=0.16\textwidth]{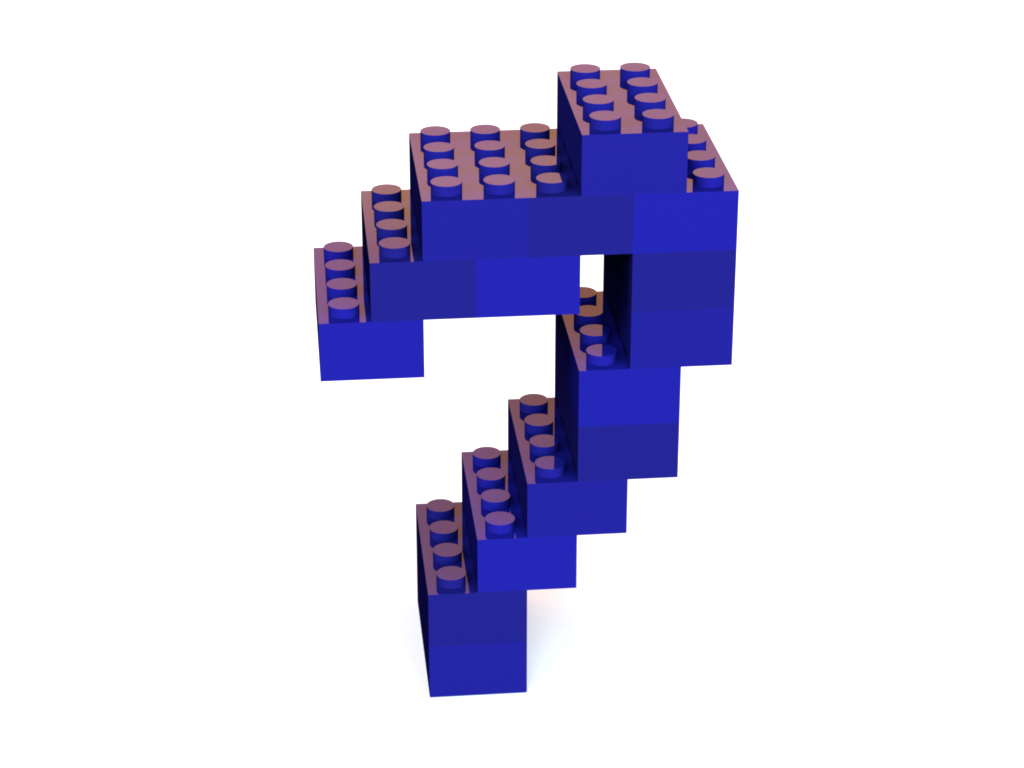}
}

\subfigure{
    \includegraphics[width=0.12\textwidth]{MNIST_all/class_8/mnist_8_0_img.png}
}
\subfigure{
    \includegraphics[width=0.16\textwidth]{MNIST_all/class_8/mnist_8_0_coords.png}
}
\subfigure{
    \includegraphics[width=0.12\textwidth]{MNIST_all/class_8/mnist_8_1_img.png}
}
\subfigure{
    \includegraphics[width=0.16\textwidth]{MNIST_all/class_8/mnist_8_1_coords.png}
}
\subfigure{
    \includegraphics[width=0.12\textwidth]{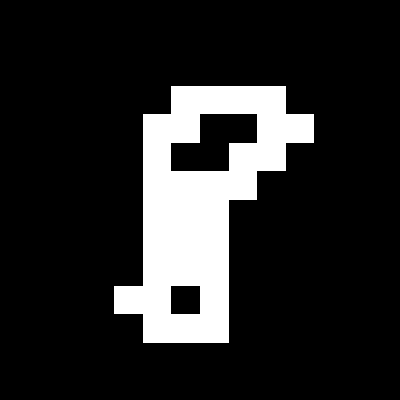}
}
\subfigure{
    \includegraphics[width=0.16\textwidth]{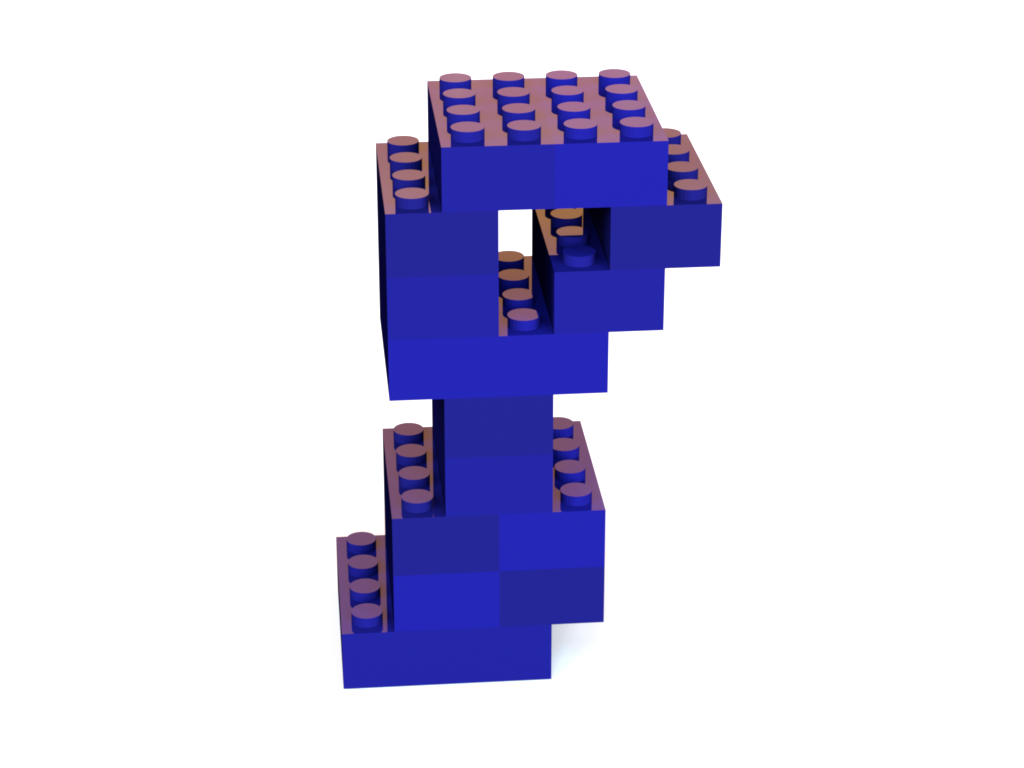}
}

\subfigure{
    \includegraphics[width=0.12\textwidth]{MNIST_all/class_9/mnist_9_0_img.png}
}
\subfigure{
    \includegraphics[width=0.16\textwidth]{MNIST_all/class_9/mnist_9_0_coords.png}
}
\subfigure{
    \includegraphics[width=0.12\textwidth]{MNIST_all/class_9/mnist_9_1_img.png}
}
\subfigure{
    \includegraphics[width=0.16\textwidth]{MNIST_all/class_9/mnist_9_1_coords.png}
}
\subfigure{
    \includegraphics[width=0.12\textwidth]{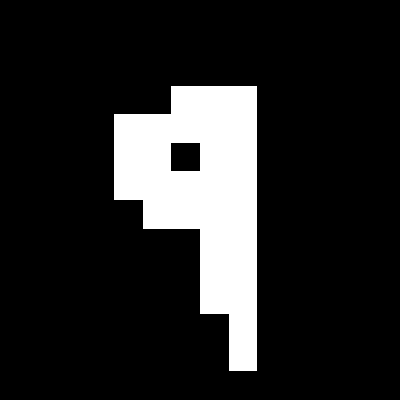}
}
\subfigure{
    \includegraphics[width=0.16\textwidth]{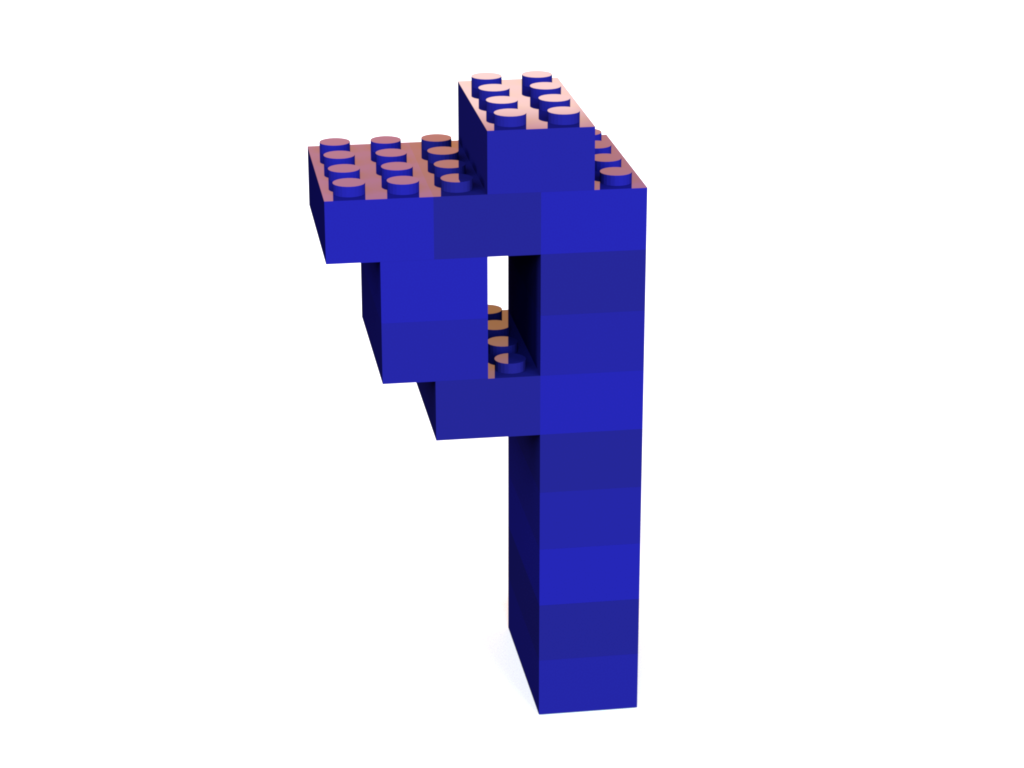}
}
\caption{Qualitative results for unseen images of all classes in MNIST construction task. The first two columns are already shown in the main article.}
\label{fig:mnist_all_qual}
\end{figure}

Average episode return for other classes of MNIST are shown in~\figref{fig:mnist_graph}. Similarly, the return curve for monitor and table classes of ModelNet are provided in~\figref{fig:modelnet_graph}.
Note that the baseline performance is measured separately for each class of MNIST and ModelNet.
We observe that $\ours$ generally outperforms the baselines 
in not only training episodes but also test episodes 
where unseen images are given.
Additional qualitative results on each digit class are presented 
in~\figref{fig:mnist_all_qual}.
Note that three images of a constructed object in both 
randomly-assembled object construction and ModelNet construction 
are extracted from same viewpoints of the target.

\section{Comparison on Graph Neural Networks}

We test $\ours$ on randomly-assembled object construction and compare to graph neural networks without node or edge features.
Specifically, no edge model only utilizes the node features 
that contain positional and directional information of each brick 
whereas no node model only uses displacement information of edge features.
The result is presented in~\figref{fig:gnn_ablation}.
Similar to the validity prediction network experiments, 
$\ours$ that exploits both node and edge features reports the best performance compared to the others.

\section{Limitations and Societal Impacts}

Our work can generate a sequence of bricks to construct
a target object of which the partial information is only available.
However, the partial information does not always guarantee that
our model constructs a 3D object accurately
because the incomplete information
cannot express the object we would like to assemble.
For example, the cases that belong to table category are difficult to assemble,
in particular with only three views of 3D object
the legs of table are not distinguishable whether a true object has two legs
or four legs. This ambiguity leads us not to successfully
construct a target object. To solve this problem,
we can provide more information than three images
from different viewpoints, but it causes an additional cost
for obtaining the information.
We need to balance a trade-off between elaborate information and
additional cost.

If our work successfully assembles any 3D object in a combinatorial manner where partial information
is given, it is capable of constructing
dangerous and illegal products with basic unit primitives.
For example, when 3D printing has been widely adopted, some people start
to produce a dangerous and illegal object such as gun, rifle, and knife
without difficulty.
Similar to this, our approach can be also employed in such tasks.
Additionally, due to the characteristics of combinatorial, addable,
removable components, a copyright of creation is able to be easily infringed.
Since our method can generate a unique sequence or assembly instruction
of object diversely, the vast number of slightly different objects can be created.

%% file: tables/tab_action_validation.tex
\begin{table}[ht]
    \centering
    \small
    \caption{Comparisons of action validation approaches.
    \label{tab:action_validation}}
    \setlength{\tabcolsep}{4pt}
    \begin{tabular}{lccc}
        \toprule
        \textbf{Method} & \textbf{Separate module} & \textbf{No access to action validity in test phase} & \textbf{Reusability} \\
        \midrule
        Direct forwarding & & \checkmark & \\
        Sampling \& checking & \checkmark & & \\
        Ours (Jointly) & \checkmark & \checkmark & \\
        Ours (Pretrained) & \checkmark & \checkmark & \checkmark \\
        \bottomrule
    \end{tabular}
\end{table}

%% file: tables/tab_avn.tex
\begin{table}[ht]
    \centering
    \small
    \caption{Results on predicting invalid actions by an action validity prediction network. Thresholds for deciding either valid or invalid actions are set to 0.5.\label{tab:avn}}
    \setlength{\tabcolsep}{4pt}
    \begin{tabular}{cccccccccc}
        \toprule
        & \multicolumn{4}{c}{\textbf{Pivot}} && \multicolumn{4}{c}{\textbf{Offset}}\\
        \cmidrule{2-5}\cmidrule{7-10}
        & \multicolumn{2}{c}{Training} & \multicolumn{2}{c}{Test} && \multicolumn{2}{c}{Training} & \multicolumn{2}{c}{Test}\\
        & Precision & Recall & Precision & Recall && Precision & Recall & Precision & Recall\\
        \midrule
        MLP & 0.9618 & \textbf{1.0000} & 0.9557 & \textbf{1.0000} && 0.5614 & 0.1410 & 0.5130 & 0.1398 \\
        No Node & 0.9874 & 0.9895 & 0.9804 & 0.9869 && 0.8261 & 0.7518 & 0.7931 & 0.7344 \\
        No Edge & 0.9947 & 0.9986 & 0.9850 & 0.9948 && 0.9199 & \textbf{0.9736} & 0.8897 & \textbf{0.9672} \\
        Ours (Jointly) & 0.9881 & 0.9988 & 0.9809 & 0.9982 && 0.9001 & 0.9505 & 0.8674 & 0.9467 \\
        Ours (Pretrained) & \textbf{0.9976} & 0.9987 & \textbf{0.9909} & 0.9944 && \textbf{0.9408} & 0.9709 & \textbf{0.9125} & 0.9661 \\
        \bottomrule
    \end{tabular}
\end{table}

%% file: tables/tab_baselines.tex
\begin{table}[t]
    \centering
    \small
    \caption{Description of baselines and our method.
    \label{tab:baselines}}
    \setlength{\tabcolsep}{4pt}
    \begin{tabular}{lcc}
        \toprule
        \textbf{Method} & \textbf{Figures} & \textbf{Description} \\
        \midrule
        Baseline \#1 - BO & \figref{fig:episode_return}, \figref{fig:mnist_graph}, \figref{fig:modelnet_graph} & Bayesian optimization~\citep{KimJ2020ml4eng} \\
        Baseline \#2 - SL & \figref{fig:3dobject_easy} & Supervised learning~\citep{ThompsonR2020neuripsw} \\
        Baseline \#3 - MLP & \figref{fig:episode_return}, \figref{fig:mnist_graph}, \figref{fig:modelnet_graph} & Our MLP-based method \\
        Ours - GNN & \figref{fig:episode_return}, \figref{fig:mnist_graph}, \figref{fig:modelnet_graph} & Brick-by-Brick \\
        \bottomrule
    \end{tabular}
\end{table}